%% file: EAP.tex
\documentclass[10pt,journal,compsoc]{IEEEtran}
\usepackage{graphicx}
\usepackage{csquotes}
\usepackage{amsfonts}
\usepackage{amsmath}
\usepackage{bbm}
\usepackage{algpseudocode,algorithm,algorithmicx}
\usepackage{booktabs}
\usepackage{tikz,pgfplots}
\usetikzlibrary{shapes,arrows}
\usetikzlibrary{matrix,positioning}
\usepackage{subfigure}
\usepackage{ulem}
\usepackage[center]{caption}
\usepackage[percent]{overpic}
\usepackage[export]{adjustbox}

\usepackage{pgfplots}
\usetikzlibrary{spy}

\newcommand{\Avec}{\mathbb{A}}
\newcommand{\Bvec}{\mathbb{B}}
\newcommand{\mc}[1]{\mathcal{#1}}
\newcommand{\Svec}{\mathbb{S}}
\DeclareMathOperator*{\argmax}{argmax}
\newcommand{\reals}{\mathbb{R}}

\ifCLASSOPTIONcompsoc
  \usepackage[nocompress]{cite}
\else
  \usepackage{cite}
\fi

\begin{document}
\title{Extended Affinity Propagation: \\Global Discovery and Local Insights}
\author{Rana Ali Amjad,~\IEEEmembership{Student Member,~IEEE,} %
	Rayyan A. Khan
	and Martin Kleinsteuber
	\IEEEcompsocitemizethanks{\IEEEcompsocthanksitem Rana Ali Amjad is with the Institute for Communications Engineering, Technical University of Munich, Germany. Email: ranaali.amjad@tum.de}
	\thanks{Rayyan A. Khan contributed equally as first author.}}

\IEEEtitleabstractindextext{%
	\begin{abstract}
		We propose a new clustering algorithm, Extended Affinity Propagation, based on pairwise similarities. Extended Affinity Propagation is developed by modifying Affinity Propagation such that the desirable features of Affinity Propagation, e.g., exemplars, reasonable computational complexity and no need to specify number of clusters, are preserved while the shortcomings, e.g., the lack of global structure discovery,  that limit the applicability of Affinity Propagation are overcome. Extended Affinity Propagation succeeds not only  in achieving this goal but can also provide various additional insights into the internal structure of the individual clusters, e.g., refined confidence values, relative cluster densities and local cluster strength in different regions of a cluster, which are valuable for an analyst. We briefly discuss how these insights can help in easily tuning the hyperparameters. We also illustrate these desirable features and the performance of Extended Affinity Propagation on various synthetic and real world datasets.
	\end{abstract}
			
	}

\maketitle

\IEEEdisplaynontitleabstractindextext
\IEEEpeerreviewmaketitle
\IEEEraisesectionheading{\section{Introduction}\label{sec:introduction}}
\IEEEPARstart{C}{luster} analysis is one of the most important tools in data mining with widespread applications ranging from bioinformatics \cite{eisengenome} to social sciences \cite{girvansocial}.

% TODO: ref

We focus on clustering based on pairwise similarities between data points. This is equivalent to graph clustering, where the aim is to cluster nodes based on pairwise interactions between the nodes,  defined via the edge weights. Numerous well known clustering algorithms have been proposed over the years to tackle this problem. Each of these algorithms has different strengths and weaknesses.  On one end of the spectrum we have Markov Clustering (MCL) \cite{mcl} that is often successful in recognizing the global structure but fails to provide a local perspective of the dataset and also suffers from high computational complexity. On the other end of the spectrum we have Affinity Propagation (AP) \cite{affinitypropagation} that has been proposed as an alternative approach to Partitioning Around Medoids (PAM) for K-medoids clustering. AP provides exemplars for clusters which reveal meaningful information about the typical characteristics of the data points in a cluster. However it is restricted to discovering only globular clusters, which severely limits its applicability. Table~\ref{tab:APqualcomp} provides a qualitative overview of the strengths and weaknesses of various well known clustering algorithms. 

In this work we modify AP to develop a new clustering algorithm, Extended Affinity Propagation (EAP), such that EAP is successful in both global structure discovery and in providing local insights into the dataset. We choose AP as our starting point due the following two reasons:
\begin{itemize}
	\item AP already possesses many desirable properties, for example, no need for a (hard) specification of the number of clusters, no initialization issues and low complexity. 
	\item The mathematical framework of factor graphs and message passing, on which AP  is based, is very flexible and provides a natural way to incorporate new information and requirements into the clustering algorithm.
\end{itemize}

We start by reviewing the related work (Sec.~\ref{sec:APrelated}) and AP (Sec.~\ref{sec:APAP}). In Sec.~\ref{sec:APEAP} we modify AP to develop EAP. Sec.~\ref{sec:APeapprop} explores some of the desirable features that EAP possesses. In Sec.~\ref{sec:APtuning} we discuss how to use the available local information to easily tune the hyperparameters of EAP. In Sec.~\ref{sec:APsynthetic} and Sec.~\ref{sec:APrealworld} we illustrate the desirable characteristics of EAP on synthetic and real world experiments, respectively. 

We do not address the important issue of how to compute the pairwise similarities in a specific scenario as it is highly application specific and should be best proposed by domain experts. Our goal in this work is to find application agnostic clustering algorithm which is able to discover the hidden structure in the pairwise similarities matrix, regardless of how the pairwise similarities matrix is computed. 
\begin{table*}[thb]
	\centering   
	\begin{tabular}{ | p{2cm} || p{2cm} | p{2cm} || p{2cm} | p{2cm} || p{2cm} | p{2cm} |}
		\toprule
		Algorithm & \multicolumn{2}{c||}{Optimization Aspects} & \multicolumn{2}{c||}{Global Discovery} & \multicolumn{2}{c|}{Local Insights} \\ \hline
		                              & Requires No of Clusters & Complexity                      & Global Structure Discovery & Outlier Detection & Exemplars & Additional Local Information \\ \hline 
		Desired                       & No                      & Low                             & Yes                        & Yes               & Yes       & Yes                          \\ 
		\midrule
		PAM \cite{pam}                & Yes                     & $\mc{O}(n^2)$                   & No                         & No                & Yes       & No                           \\ \hline
		AP \cite{affinitypropagation} & No                      & $\mc{O}(n^2)$                   & No                         & Yes               & Yes       & No                           \\ \hline
		MCL \cite{mcl}                & No                      & $\mc{O}(n^3)$                   & Yes                        & Yes               & No        & No                           \\ \hline
		DBSCAN \cite{dbscan}          & No                      & $\mc{O}(n^2)$                   & $\sim$                     & $\sim$            & No        & No                           \\ \hline
		Spectral \cite{spectral}      & Yes                     & $\mc{O}(n^3)$                   & $\sim$                     & No                & No        & No                           \\ \hline
		Heirarchical                  & No                      & $\scriptstyle\mc{O}(n^2\log n)$ & $\sim$                     & $\sim$            & No        & $\sim$                       \\ \hline 
		\textbf{EAP}                  & No                      & $\mc{O}(n^2)$                   & Yes                        & Yes               & Yes       & Yes                          \\ \hline
	\end{tabular}
	\caption{ Qualitative comparison of famous pairwise similarity based  clustering algorithms w.r.t. various desired characteristics. $n$ denotes the cardinality of the dataset and $\sim$ implies that the algorithm is not too good but also not bad for the desired trait.}
	\label{tab:APqualcomp}
\end{table*}
\section{Related Work}\label{sec:APrelated}
Since it was first published in \cite{affinitypropagation}, AP has been modified in different ways. Hierarchical Affinity Propagation, introduced in \cite{hap}, proposes a layered structure where the exemplars of previous optimization layer are considered as the data points for the next layer. Hierarchical Affinity Propagation tries to cluster the data hierarchically without making hard decisions at each hierarchical layer. Although the local exemplars obtained so are more meaningful than AP, the clusters obtained by Hierarchical Affinity Propagation are still globular at each layer and there is limited information about the local structure of the clusters beyond local exemplars. Multi-Exemplar Affinity Propagation \cite{meap} is another approach, closely related to HAP with two layers, where the authors propose the use of exemplars and super-exemplars. Exemplars can select super-exemplars as representatives but super-exemplars are forced to select themselves. Multi-Exemplar Affinity Propagation has similar drawbacks as Hierarchical Affinity Propagation.

Unlike other variants, Soft-Constraint Affinity Propagation, introduced in \cite{scap},  allows exemplars to select other exemplars as their representative by relaxing the consistency constraint. As a result Soft-Constraint Affinity Propagation can discover a wider variety of cluster shapes. On the other hand since Soft-Constraint Affinity Propagation tries to identify global structure based on the pairwise similarities between exemplars, this corresponds to expanding a cluster by establishing direct links between the local exemplars. Hence it often leads to sub-optimal clustering. Furthermore, the only local information available is the local exemplars and the direct connections between them. 

A more natural approach to identify arbitrarily shaped clusters is to combine subclusters corresponding to each local exemplar by exploring the shared connections between subcluster boundaries. This is the approach we will take in this work. It has not only better and more robust clustering results but also provides us with additional local insights into the discovered clusters.
\section{Affinity Propagation}\label{sec:APAP}
In Affinity Propagation each cluster is represented using an exemplar. The exemplar itself is a data point belonging to the respective cluster. Given the dataset $\mathcal{X} = \{x_,x_2,\cdots,x_n\}$ and the pairwise similarity matrix $\Svec$, the goal in AP is to find exemplars and cluster assignments such that the sum of pairwise similarities between the data points and the exemplars associated with the respective clusters assigned to the data points is maximized:
\begin{equation}
	\argmax_{c_1\dots,c_n} \sum\limits_{i} s_{i,c_i}    \quad s.t. \quad c_{c_j} = c_j \ \forall j \label{eq:APobjsum}
\end{equation}
where $s_{ij} = \Svec(i,j) \in \reals$ and $c_i$ refers to the exemplar selected by the data point $x_i$, which also determines the cluster for $x_i$ since each cluster is associated with a single exemplar. $s_{jj}$ represents the preference of data point $x_j$ to become an exemplar. In case of no additional a priori information $s_{jj}$ is assigned the same value $p$ for all data points where $p$ is a hyperparameter of the algorithm, representing the initial desire of each point to become an exemplar. Hence in AP, initially each data point is treated as a potential exemplar and, therefore, initialization issues are circumvented. The constraint $c_{c_j} = c_j$, known as the consistency constraint, forces that if a data point $x_i$ is chosen as exemplar by some other point(s), it must choose itself as it's exemplar too, hence promoting compact clusters. The self preference $p$ in conjunction with the consistency constraint, which motivates compact clusters, acts as a soft initial guidance for AP to determine the number of clusters needed for the dataset.

\begin{figure}[t]
	\centering
	\resizebox{0.8\linewidth}{!}{%
		\input{./figures/AP_fac_graph}
	}
	\caption{ Factor graph of \eqref{eq:obj_final} \cite{dueckthesis}.}
	\label{fig:binary_factor_graph}
\end{figure}
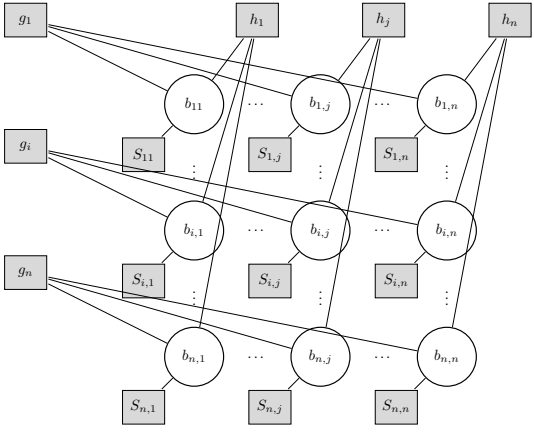

\eqref{eq:APobjsum} is an NP-Hard combinatorial optimization problem \cite{inmar_thesis}. AP solves it sub-optimally in $\mc{O}(n^2)$ computations using max-sum algorithm. For this purpose we present the reformulated version of AP, described in \cite{dueckthesis}, involving binary optimization variables: Let us define a matrix $\Bvec \in \{0,1\}^{n \times n}$ of binary variables $b_{ij} = \Bvec(i,j)$, where $b_{ij} = 1 \iff c_i = j$. $\Bvec (i,:)$ can be considered as a one hot encoding of $c_i$. Since every data point can choose only one exemplar, the objective function in \eqref{eq:APobjsum} can be written as:

\begin{align}
	\argmax_{\Bvec}  \sum \limits_{i,j} s_{ij}b_{ij}  \ \ \quad s.t. \quad \ \begin{cases}
	\sum \limits_{j} b_{ij} = 1 \&\forall i      &   \\
	b_{jj} = \max \limits_{i}b_{ij} \&\forall j. &   
	\end{cases} \label{obj-3}
\end{align}
We can reformulate \eqref{obj-3} as the following unconstrained optimization problem:
\begin{equation}
	\argmax_{\Bvec} \Big(
	\sum \limits_{i,j} s_{ij}b_{ij} + 
	\sum \limits_{i}{g}_i(\Bvec(i,:)) +
	\sum \limits_{j}{h}_j(\Bvec(:,j))
	\Big)
	\label{eq:obj_final}
\end{equation}
where
\begin{align}
	g_i(\Bvec(i,:)) = & \begin{cases}                               
	0                 & if \enspace \sum \limits_{j} b_{ij} = 1     \\
	-\infty           & otherwise                                   
	\end{cases} \label{eq:I_constraint} \\ 
	h_j(\Bvec(:,j)) = & \begin{cases}                               
	0                 & if \enspace b_{jj} = \max \limits_{i}b_{ij} \\
	-\infty           & otherwise.                                  
	\end{cases} \label{eq:E_constraint}
\end{align}
AP solves \eqref{eq:obj_final} by mapping it to the factor graph shown in Fig.~\ref{fig:binary_factor_graph}. The factor nodes $S_{ij}$ correspond to the function $s_{ij}b_{ij}$. AP involves two steps: message passing and decision mechanism.
\subsection{Message Passing}\label{subsec:APmsgpassing}
\begin{figure}[t]
	\centering
	\input{./figures/AP_messages}
	\caption{Message passing between variable nodes and factor nodes in AP  \cite{dueckthesis}.}
	\label{fig:ap-messages}
\end{figure}
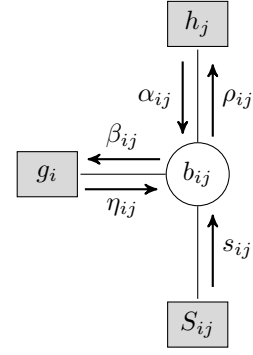

We can find a (possibly sub-optimal) solution of \eqref{eq:obj_final} by applying max-sum algorithm to the factor graph in Fig.~\ref{fig:binary_factor_graph}.  The outgoing messages from factor nodes are denoted by $\nu_{.}$ and the outgoing messages from variable nodes are denoted by $\mu_{.}$. The messages exchanged between the variable node $b_{ij}$ and the factor nodes $g_i$ and $h_j$ are shown in Fig.~\ref{fig:ap-messages}, where we have used the following shorthand notation for convenience:
\begin{align}
	\beta_{ij}  & = \mu_{b_{ij} \rightarrow g_i} (1) -  \mu_{b_{ij} \rightarrow g_i}(0) \label{eq:betadiffdef} \\
	\eta_{ij}   & =  \nu_{g_i \rightarrow b_{ij}}(1) - \nu_{g_i \rightarrow b_{ij}}(0)  \label{eq:etadiffdef}  \\
	\rho_{ij}   & = \mu_{b_{ij} \rightarrow h_j} (1) -  \mu_{b_{ij} \rightarrow h_j}(0)                        \\
	\alpha_{ij} & =  \nu_{h_j \rightarrow b_{ij}}(1) - \nu_{h_j \rightarrow b_{ij}}(0).                        
\end{align}
For solving \eqref{eq:obj_final}, it sufficient to exchange these scalar value between the factor and variable nodes, constituting the difference of the message values for $b_{ij}=1$ and $b_{ij}=0$, as we are only interested in determining the maximizer. The final equations used for computing these difference messages are:
\begin{align}
	\beta_{ij} =\,                                                            & s_{ij} + \alpha_{ij}\label{eq:APmsgbeta}                \\
	\rho_{ij} =\,                                                             & s_{ij} + \eta_{ij} \label{eq:APmsgrho}                  \\
	\eta_{ij} =\,                                                             & -\max \limits_{k \neq j} \beta_{ik} \label{eq:APmsgeta} \\
	\alpha_{ij} =\,                                                           & \begin{cases}                                           
	\sum \limits_{ k \neq j} max(0, \rho_{kj})                                & i=j                                                     \\
	\min[0, \rho_{jj} + \sum \limits_{k \not \in \{i,j\}} \max(0, \rho_{kj})] & i \neq j                                                
	\end{cases} \label{eq:APmsgalpha}
\end{align}
Since $g_i(\cdot)$ and $h_j(\cdot)$ represent constraints in the original optimization problem, we also refer to the corresponding factor nodes sometimes as constraint nodes.  
The detailed derivations  and intuitive meanings of \eqref{eq:APmsgalpha} and \eqref{eq:APmsgeta} are given in \cite{inmar_thesis}. 
\subsection{Decision Phase}\label{subsec:APAPdecision}
During the iterations, we can approximate the difference of the cost function \eqref{eq:obj_final} for the two values of $b_{ij}$ via accumulated belief $a_{ij}$ given by the sum of all incoming messages to $b_{ij}$, i.e., $a_{ij} = s_{ij} + \eta_{ij} + \alpha_{ij}$. This sum corresponds to an approximate evaluation of \eqref{eq:obj_final} for  the two values of $b_{ij}$ and then computing the difference. Current cost maximizing value of $b_{ij}$ is decided by thresholding $a_{ij}$ at 0, i.e., $b_{ij} = 1$ if $a_{ij} > 0$ and $b_{ij}=0$ otherwise. Let $\Avec$ be the matrix of accumulated beliefs $a_{ij}$. Convergence is achieved in AP if the values of diagonal elements of $\Avec$ do not change over a specified number of iterations. Once message passing phase converges, we choose the set of exemplars as follows: $\mc{E} = \{k \ | \  a_{kk} > 0\}$. Each non exemplar point is then assigned to the exemplar most similar to it \cite{inmar_thesis}
\begin{align}
	c_i = \argmax \limits_{k \in \mc{E}}  s_{ik} \qquad  \forall i \not \in \mc{E}. \label{eq:APAPdec} 
\end{align}
\section{Extended Affinity Propagation (EAP)}\label{sec:APEAP}
AP possesses various desired traits, for example, it does not need  to initialize cluster assignments, hence making it impervious to initialization issues. It only requires a soft estimate of the desired number of clusters fed as an input in the form of self preference $p$. The computational complexity of AP is $\mc{O}(n^2)$. Furthermore it provides some local information about each cluster in the form of an exemplar. Unfortunately AP also suffers from the following issues: it can only discover globular clusters which seriously limits it's applicability. Besides the only local information it provides about each of these globular clusters is an exemplar. 

In this section we will modify AP to develop a new algorithm which inherits the positive aspects of AP while alleviating its' shortcomings. To achieve this we need to modify the cost function (consequently also the message passing phase) and the decision mechanism for AP. The basic principle driving our approach is to allow multiple well separated local exemplars in a cluster and to permit a data point to connect to multiple local exemplars if it is close enough to all of them. These data points that form \enquote{boundary} connections between local exemplars, by connecting to multiple local exemplars, enable the decision mechanism, which looks for connected components in a graph,  to discover a wide variety of global patterns in $\Svec$. Moreover, the local exemplars together with the boundary connections between them provide meaningful insights into the internal structure of a cluster as we will see in the Secs. \ref{sec:APeapprop}, \ref{sec:APsynthetic} and \ref{sec:APrealworld}.
\subsection{Modified Cost Function and Message Passing}\label{subsec:APEAPmsg}
\subsubsection{Improved Global Discovery}
The $g_i(.)$ constraint in AP aim to enforce that $x_i$ chooses only one exemplar. Combined with this, the decision process described in Sec.~\ref{subsec:APAPdecision} forces globular clusters. For better global structure discovery EAP allows each data point to connect to possibly multiple local exemplars and hence form boundary connections between local exemplars. This is achieved by modifying $g_i(\cdot)$ as follows:
\begin{equation}
	\overline{g}_i(\Bvec(i,:)) = \begin{cases} 
	u(\Bvec(i,:)) & \text{if} \enspace \sum \limits_{j} b_{ij} \geq 1 \\
	-\infty & \text{otherwise}
	\end{cases}\label{eq:g_func}
\end{equation}
where 
\begin{equation}
	u(\Bvec(i,:)) := \sum \limits_{j} b_{ij}q\label{eq:g_func2}
\end{equation}
where $q$  is a new hyperparameter denoting the penalty incurred when a data point connects to an exemplar. This leads to the following unconstrained optimization
\begin{align}
	\argmax_{\Bvec} & \sum \limits_{i,j}b_{ij}\Big(s_{ij} + q \Big) + \sum \limits_{j}{h}_j(\Bvec(:,j)) \nonumber \\ &+ \sum \limits_{i} \log \mathbbm{1}\Big(\sum \limits_j b_{ij} \geq 1 \Big).
	\label{eq:APEAPobjmain1}
\end{align}
where $\mathbbm{1}(\cdot)$ denotes the indicator function. Our choice of $u(.)$ is motivated by the following two reasons:
\begin{itemize}
	\item It can be seen from \eqref{eq:g_func} and \eqref{eq:APEAPobjmain1} that a data point $x_i$ is motivated to connect to a potential local exemplar $x_j$ only if their pairwise similarity is high.  $\overline{g}_i(.)$ penalizes the global cost function linearly with a penalty $q$ for each potential local exemplar that $x_i$ chooses to connect to. However $q$ does not affect the selection of the first exemplar as this is still a hard requirement of $\overline{g}_i(.)$, i.e., every data point is bound to select at least one local exemplar.
	\item The computations to determine all the outgoing messages from factor node $\overline{g}_i(\cdot)$ still have  $\mathcal{O}(n)$ overall computational complexity, same as the factor node $g_i(.)$ in AP. Hence the complexity of the new algorithm does not increase due to this change. Other choices of $u(\cdot)$ that are suitable in terms of the performance, which we investigated,  lead to higher computational complexity \cite{hopmap}. 
\end{itemize}
The outgoing messages from $\overline{g}_i(.)$ are
\begin{equation}
	\eta_{ij} = \max \left( -\max \limits_{m \neq j} \beta_{im} \ ,\  q\right). \label{eq:APEAPmsgi}
\end{equation}
The derivation of \eqref{eq:APEAPmsgi} is given in Appendix~\ref{app:EAPeta}. $\eta_{ij}$ represents the maximum penalty due to other potential local exemplars for data point $x_i$. To understand the impact of modifying $g_i(.)$, let's compare \eqref{eq:APEAPmsgi} to \eqref{eq:APmsgeta}. Unlike in AP, this penalty is now limited by $q$. Thus the negative effect of other potential local exemplars for $i$ on $b_{ij}$ being $1$ is limited. This allows a point to connect to more than one potential local exemplar if it is close enough to them. We can recover AP by setting $q = - \infty$ (and subtracting the additional $nq$ factor in \eqref{eq:APEAPobjmain1} to avoid the an ill-posed optimization problem).
\subsubsection{Improved Local Insights}
\begin{figure}[t]
	\centering 
	\subfigure[Without neighbourhood constraint $r_j$]{\includegraphics[width=0.22\textwidth]{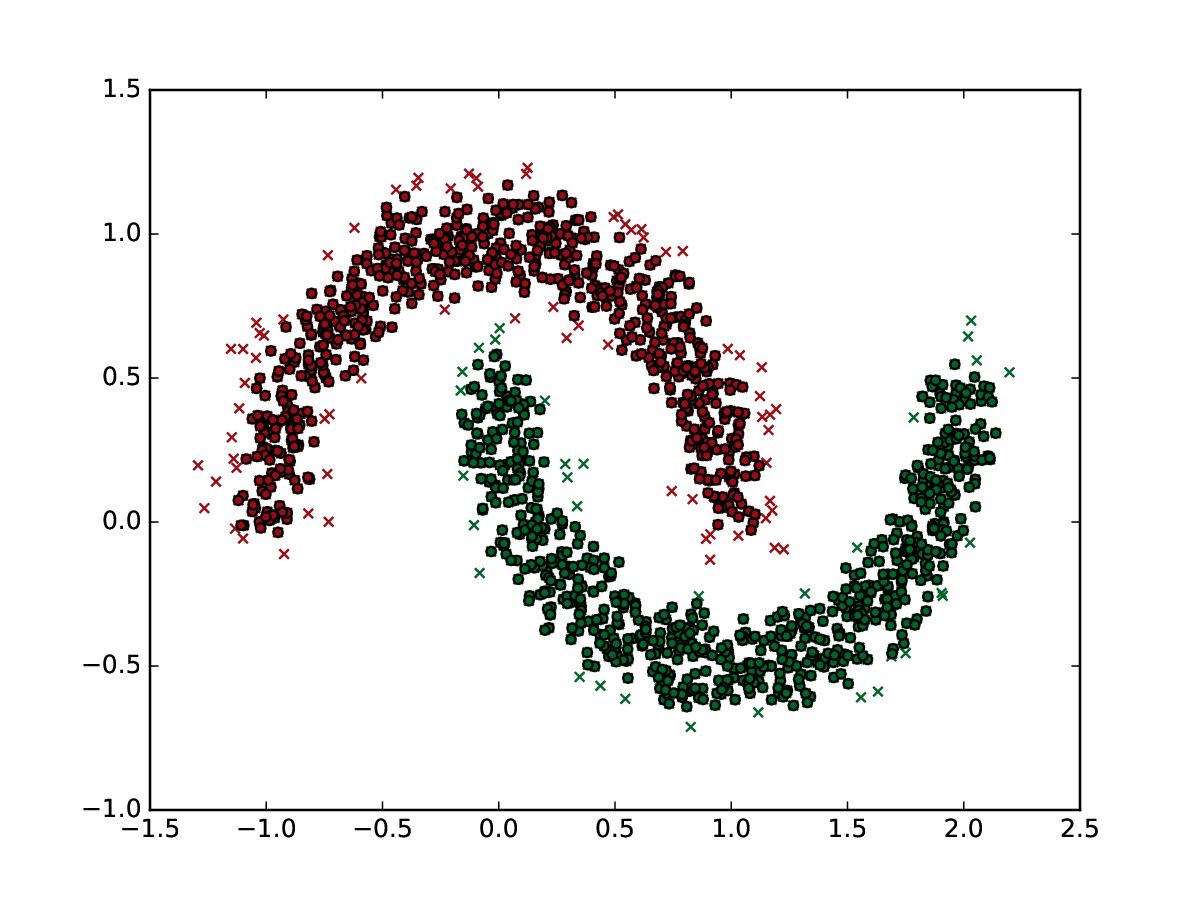}\label{fig:APEAPwithoutneigh}} \hfill
	\subfigure[With neighbourhood constraints $r_j$]{\includegraphics[width=0.22\textwidth]{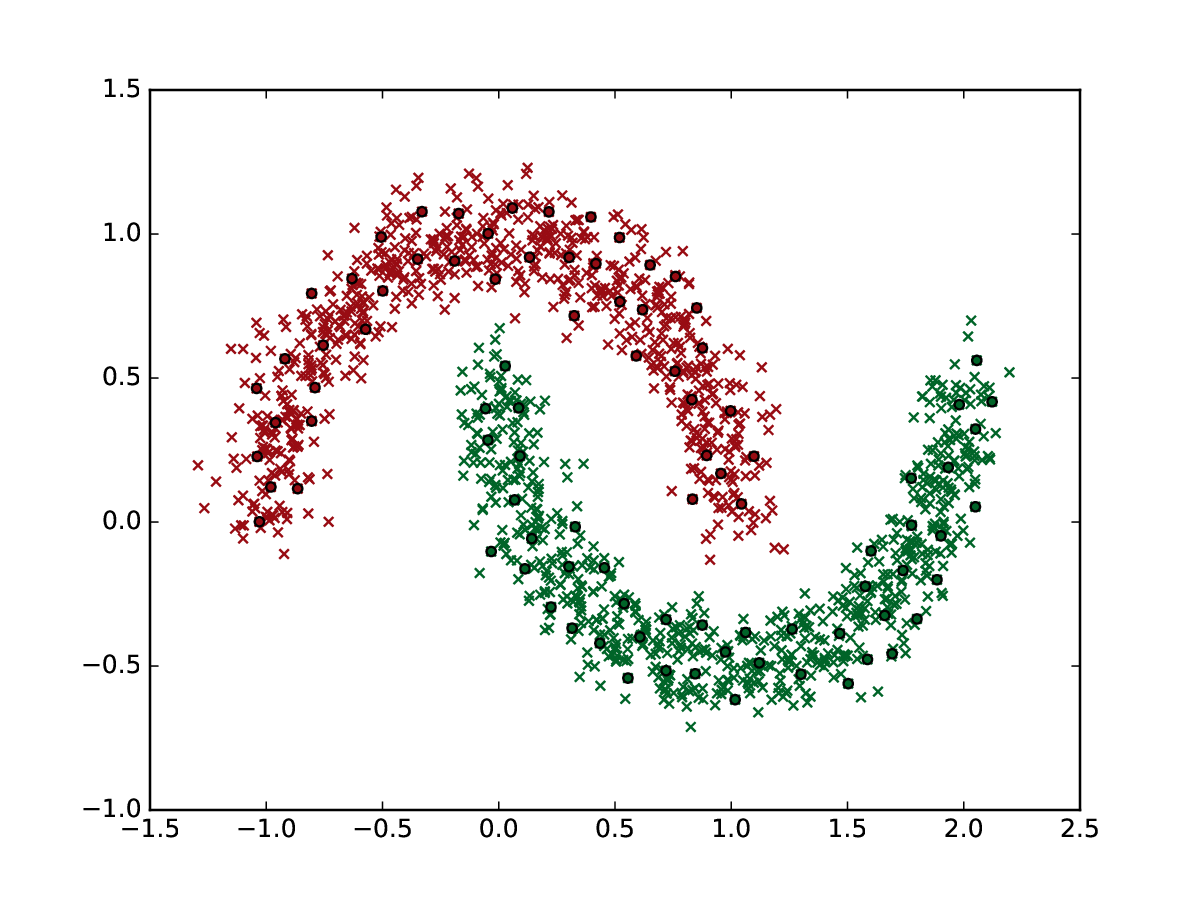}\label{fig:APEAPwithneigh}}
	\caption{The figures show the impact of neighbourhood constraints $r_j(\cdot)$ on the discovered local exemplars. Data points belonging to same discovered cluster are marked using the same color and local exemplars are marked using circles. It is also clear from the figure that the global structure discovery has not been impacted by the neighbourhood constraints. The two figures have been generated for the same $p=0.6$ and $q=-0.97$, whereas the figure on the right has $\Delta = 0.99$.}
	\label{fig:EAPlocalviewneighimpact} 
\end{figure}

The new relaxed constraint $\overline{g}_i(\cdot)$ along with the new decision mechanism that we will explain in Sec.~\ref{subsec:APEAPdecision} solves the problem of global structure discovery by allowing the subclusters to be merged together using the \enquote{boundary} connections. However, the relaxed constraint $\overline{g}_i(\cdot)$ obscures the local information. To understand why, consider a dataset that lies in a metric space and the pairwise similarities depend inversely on the pairwise distances. In AP, if a point $x_j$ is selected as an exemplar, the points close to $x_j$ also have a high motivation of becoming an exemplar. However $g_i(\cdot)$ forces only one exemplar to appear in each cluster and suppresses others. When $g_i$ is relaxed to $\overline{g}_i$, many local exemplars appear very close by in dense regions of the dataset. Hence it becomes difficult to point out good representatives of data points. In other words the local information about the cluster structure is obscured. This phenomenon of close by local exemplars is shown in Fig.~\ref{fig:APEAPwithoutneigh}. In order to resolve this, we introduce a new set of constraints on the diagonal elements of $\Bvec$, i.e., the elements signifying potential local exemplars.  For every data point $x_j$, let $\mc{N}_j = \{j\} \bigcup\{k|s_{jk} > \Delta \}$, i.e., the $\Delta$-neighborhood around $x_j$. Well separated exemplars can  be obtained by enforcing a maximum of one exemplar in each neighbourhood  $\mc{N}_j$ by introducing the following constraint in \eqref{eq:APEAPobjmain1} as an additional added term for each $x_j$
\begin{equation}
	r_j(\mc{N}_j) = \begin{cases}
	0 & if \enspace \sum \limits_{k \in \mc{N}_j} b_{kk} \leq 1 \\
	-\infty & otherwise.
	\end{cases}\label{eq:F_j-constraint}
\end{equation}

\begin{figure}[t]
	\centering
	\input{./figures/EAP_diagonalnode}
	\caption{Messages passed between a diagonal variable node $b_{ii}$ and it's adjacent constraint nodes in EAP.}
	\label{fig:EAP_diagonalnode}
\end{figure}
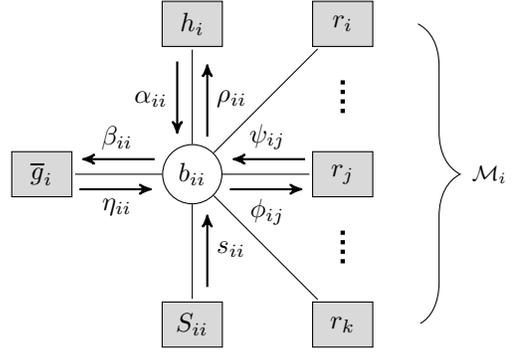

Since a point $x_j$ can belong to $\Delta$-neighbourhood of multiple points, each $b_{jj}$ can have multiple adjacent $r_k(\cdot)$ constraints. $\mc{M}_j = \{k|j \in \mc{N}_k\}$ denotes the set of all points whose $\Delta$-neighbourhoods contain $x_j$. Fig.~\ref{fig:EAP_diagonalnode} shows the  messages exchanged between the diagonal variable nodes and the factor nodes for EAP, where we defined the following shorthand notation:
\begin{align}
	\phi_{ij} & = \mu_{b_{ii} \rightarrow r_j} (1) -  \mu_{b_{ii} \rightarrow r_j}(0) \\
	\psi_{ij} & =  \nu_{r_j \rightarrow b_{ii}}(1) - \nu_{r_j \rightarrow b_{ii}}(0). 
\end{align}  
The outgoing messages from $r_j$ are:
\begin{equation}
	\psi_{ij} = - \max(0, \max \limits_{\substack{l \in \mc{N}_j \\ l \neq i}} \phi_{lj}). \label{eq:eap_psi_main}
\end{equation}
The derivation of \eqref{eq:eap_psi_main} is given in Appendix~\ref{app:EAPpsi}. The overall complexity of computing all the outgoing messages $\psi_{ij}$ from $r_j$ is $\mathcal{O}(n)$. \eqref{eq:eap_psi_main} provides an intuitive insight into the effect of constraint $r_j(\cdot)$. If we have more than one strong contenders for being a local exemplar in a neighborhood $\mc{N}_j$, they will all try to suppress one another. In such a scenario most of the potential local exemplars in the neighbourhood will get suppressed, leading to well separated local exemplars in the neighbourhood. Fig.~\ref{fig:APEAPwithneigh} shows impact of introducing the $r_j(\cdot)$ constraints when compared to Fig.~\ref{fig:APEAPwithoutneigh}. We can also observe that, within a reasonable range, the introduction of $r_j(\cdot)$ has negligible (if any) impact on the cluster assignments. 

The messages exchanged between the variable and the factor nodes for the new factor graph corresponding to EAP can be summarized as follows:
\begin{align}
	\beta_{ij}                                                                & = s_{ij} + \alpha_{ij} + \mathbbm{1}(i=j)\sum \limits_{k \in \mc{M}_i} \psi_{ik}\label{eq:eap_beta_1}         \\
	\eta_{ij}                                                                 & = \max \left( - \max \limits_{k \neq j}  \beta_{ik}  \ , q \right) \label{eq:eap_eta_1}                       \\
	\phi_{ij}                                                                 & = s_{ii} + \alpha_{ii} + \eta_{ii} + \sum \limits_{k \in \mc{M}_i\backslash j} \psi_{ij} \label{eq:eap_phi_1} \\
	\psi_{ij}                                                                 & = - \max(0, \max \limits_{\substack{l \in \mc{N}_j                                                            \\ l \neq i}} \phi_{lj}) \label{eq:eap_psi_1} \\
	\rho_{ij}                                                                 & = s_{ij} + \eta_{ij} + \mathbbm{1}(i=j)\sum \limits_{k \in \mc{M}_i} \psi_{ik}.\label{eq:eap_rho_1}           \\
	\alpha_{ij}                                                               & =\,  \begin{cases}                                                                                            
	\sum \limits_{ k \neq j} max(0, \rho_{kj})                                & i=j                                                                                                           \\
	\min[0, \rho_{jj} + \sum \limits_{k \not \in \{i,j\}} \max(0, \rho_{kj})] & i \neq j.                                                                                                     
	\end{cases} \label{eq:eap_alpha_1}
\end{align}

The message $\alpha_{ij}$ is the same as in AP as we have not modified the consistency constraint. So it can be interpreted in a similar way as for AP \cite{affinitypropagation}. 
\subsection{Decision Mechanism}\label{subsec:APEAPdecision}
Even with the modified cost function and messages from Sec.~\ref{subsec:APEAPmsg}, the decision mechanism specified in Sec.~\ref{subsec:APAPdecision} will lead to globular clusters. The is because \eqref{eq:APAPdec} does not utilize the boundary connections, where some data points want to connect to more than one local exemplar. Rather it connects each data point only to the closest local exemplar and considers the globular subcluster associated with each local exemplar as a separate cluster. Basically it ignores the information provided by the off diagonal elements of $\Avec$.  Hence we need to modify the decision mechanism to utilize the information present in the new $\Avec$ in a better manner. 
Like AP, after every iteration we check for the sum of all incoming messages to $b_{ii}$, i.e., $a_{ii} = s_{ii}+\eta_{ii}+\alpha_{ii}+\sum \limits_{k:i \in \mc{M}_i} \psi_{ik}$, for convergence. 
The decision phase that follows convergence of these messages is described in Alg.~ \ref{algo:eap-dec}.  Clusters are discovered by extracting connected components of the graph for adjacency matrix defined by $\Bvec_s$, the symmetrized version of $\Bvec$ \cite{connectedComponents}. $\mathcal{E}$, $\Bvec$ and $\mc{L}_i$ now contain local information about the clusters. $\ell(\cdot)$ defines the cluster assignment
\begin{algorithm}[t]
	\caption{EAP Decision Mechanism}\label{algo:eap-dec}
	\begin{algorithmic}[1]
		\Function {DecisionPhase }{$\Avec$}
		\State $\Bvec \gets \mathbbm{1}(\Avec > 0)$ \Comment{Element-wise thresholding of $\Avec$}
		\State $\mathcal{E} \gets \{k \ | \ b_{kk} = 1\}$ \Comment{Set of local exemplars}
		% \State $\bm{\mathcal{L}} \gets [ \ ]$ \Comment{List of labels}
		\For {$j \not \in \mathcal{E}$} \Comment{Ensure local consistency constraint}
		\State $\Bvec(:,j) \gets 0$
		\EndFor
		\For{$i\gets 1$ to $N$}
		\State $\mathcal{L}_i \gets \{k | k \in \mathcal{E}$ and $\ h_{ik} = 1 \}$ \Comment{Local exemplars connected to $x_i$}
		\If {$\mathcal{L}_i == \emptyset$} 
		\Comment{Isolated points}
		\State $\mathcal{L}_i \gets \{i\}$
		\State $b_{ii} \gets 1$
		\EndIf
		\EndFor
		\State $\Bvec_s \gets \Bvec \wedge \Bvec^T$ \Comment{Symmetrize $\Bvec$}
		\State $\ell(\cdot)$ $\gets$ ConnectedComponents($\Bvec_s$) \Comment{Discover connected components \cite{connectedComponents}}
		\State \textbf{return} $\mathcal{E}$, $\Bvec$, $\mathcal{L}$, $\ell$
		\EndFunction
	\end{algorithmic}
\end{algorithm}

The overall computational complexity of EAP is  $\mathcal{O}(n^2)$. In practice EAP often requires less message passing iterations to converge due to the relaxed $\overline{g}_i(\cdot)$ constraints.
\section{Local Insights Gained via EAP}\label{sec:APeapprop}
Besides being able to discover clusters with different structures, EAP has various additional features. We will briefly discuss some of them in this section and we refer the interested reader to Appendix \ref{app:addfeatures} for further details. 
\subsection{Local Exemplars}\label{subsec:localexemp}
\begin{figure}
	\centering
	\subfigure[Confidence via nearest neighbours]{
		\begin{tikzpicture}[spy using outlines={rectangle,magnification=3,size=3cm, connect spies}]
			\node[anchor=south west] at (0,0) {\includegraphics[width=0.5\textwidth, height=0.4\textwidth]{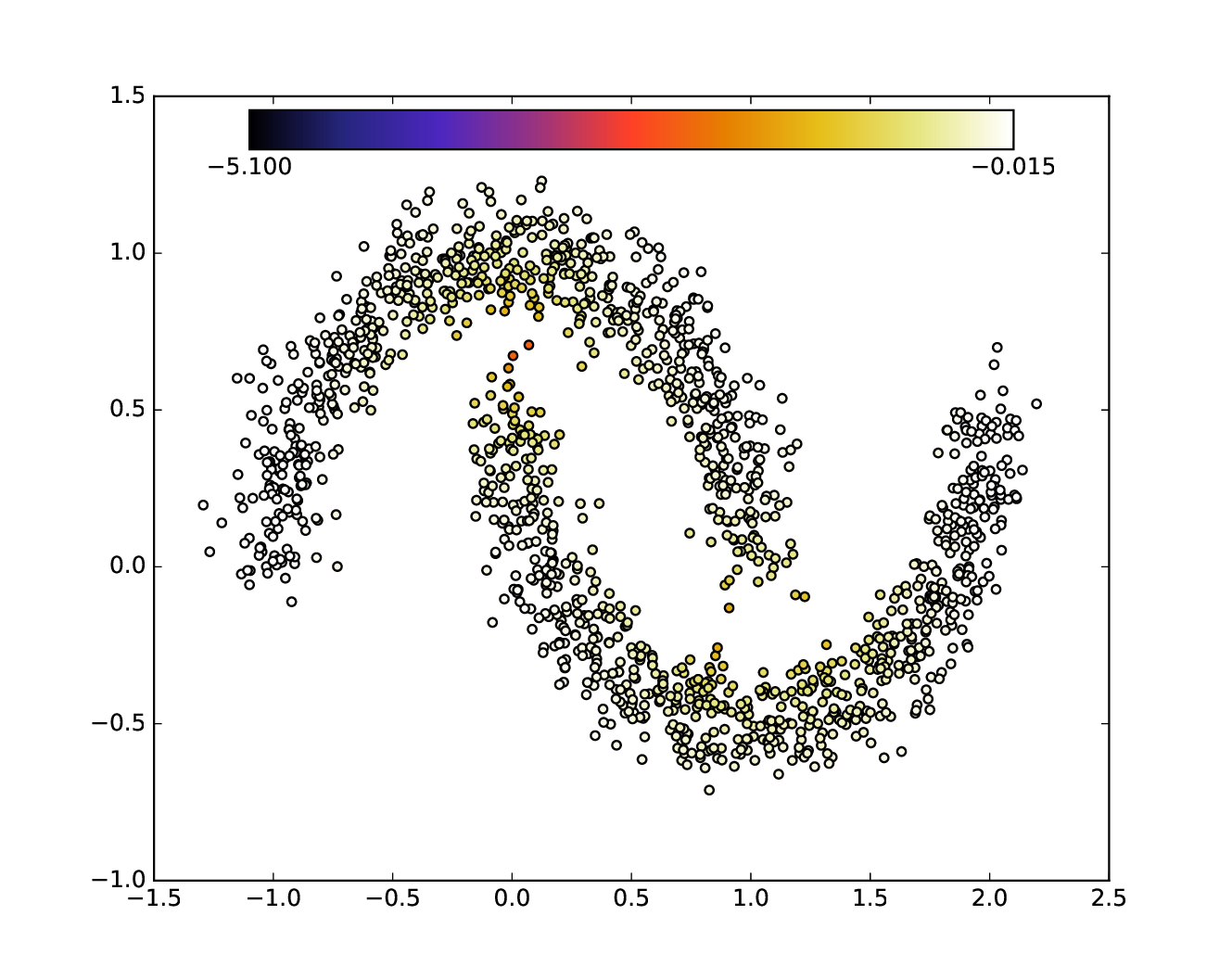}};
			\spy[red, every spy on node/.append style={thick}] on (4,4.8) in node [left] at (4.3, -1.5);
			\spy[blue, every spy on node/.append style={thick}] on (5.8,2.7) in node [left] at (8.3, -1.5);
		\end{tikzpicture}
		\label{fig:APEAPconfidencemcltype}
	}
		   
	\subfigure[Confidence via nearest local exemplars]{
		\begin{tikzpicture}[spy using outlines={rectangle,magnification=3,size=3cm, connect spies}]
			\node[anchor=south west] at (0,0) {\includegraphics[width=0.5\textwidth, height=0.4\textwidth]{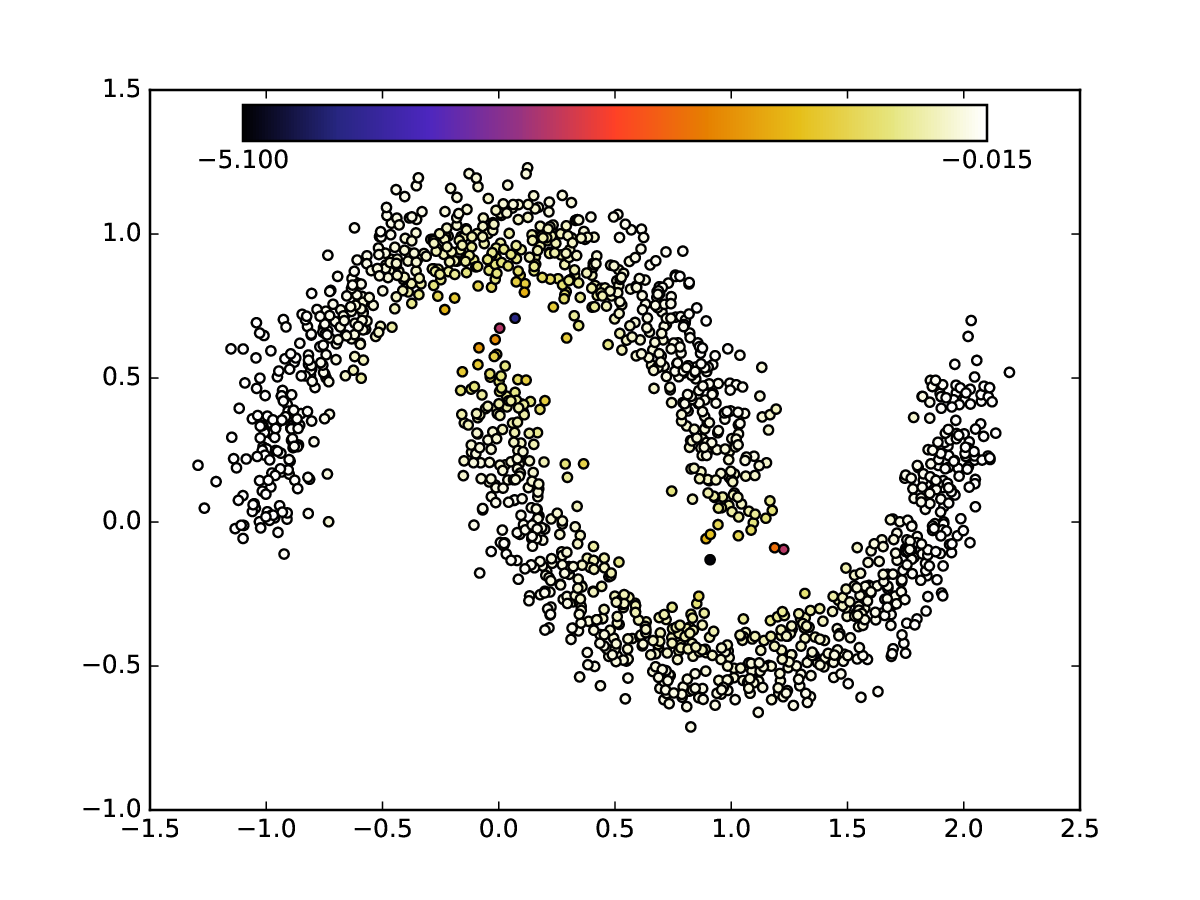}};
			\spy[red, every spy on node/.append style={thick}] on (4,4.8) in node [left] at (4.3, -1.5);
			\spy[blue, every spy on node/.append style={thick}] on (5.8,2.7) in node [left] at (8.3, -1.5);
		\end{tikzpicture}
		\label{fig:APEAPconfidenceeap}
	}
		
	\caption{The figures depict the confidence values for data points in the half-moons dataset, obtained by (a) using \eqref{eq:APmcltypeconfidence}, (b) using  \eqref{eq:APEAPconfidence}. The clustering results used in either case are the same as in Fig.~\ref{fig:APEAPwithneigh}, only the confidence metric used is different. Note that we have used the same color gradient scale in both figures for comparison.}
	\label{fig:APEAPconfidenceplot} 
\end{figure}

The local exemplars discovered by AP and EAP often represent the typical characteristics of a subset of the dataset. For EAP, however, their benefit is not only limited to this. They can be used to cluster new data points more efficiently ($\mc{O}(|\mc{E}|)$ instead of $\mc{O}(n)$, cf. Appendix~\ref{app:newpoints}) and more robustly while still maintaining the global structure discovered earlier. The local exemplars can also be used to assign insightful confidence values for the clustering decisions taken by EAP. Consider that we want to assign a value to $x_i$, indicating the confidence of EAP about the cluster assigned to $x_i$. To illustrate the advantage of having local exemplars for this purpose, consider the two very similar ways in \eqref{eq:APmcltypeconfidence} and \eqref{eq:APEAPconfidence} to compute confidence values, with the only difference that  \eqref{eq:APmcltypeconfidence} uses nearest neighbours whereas \eqref{eq:APEAPconfidence} exploits local exemplars. 
\begin{align}
	\overline{w}(x_i) & = \log \left(\max \limits_{\substack{x_k \in \mathcal{X} \\ \ell(x_k)=\ell(x_i)}} s_{ik} - \max \limits_{\substack{x_j \in \mathcal{X} \\ \ell(x_j)\neq\ell(x_i)}} s_{ij}\right) \label{eq:APmcltypeconfidence} \\
	w(x_i)            & = \log \left(\max \limits_{\substack{k \in \mathcal{E}   \\ \ell(x_k)=\ell(x_i)}} s_{ik} - \max \limits_{\substack{j \in \mathcal{E} \\ \ell(x_j)\neq\ell(x_i)}} s_{ij}\right). \label{eq:APEAPconfidence}
\end{align}
The confidence values obtained via \eqref{eq:APmcltypeconfidence} and \eqref{eq:APEAPconfidence}, for the same clustering result, are shown in Fig.~\ref{fig:APEAPconfidencemcltype} and Fig.~\ref{fig:APEAPconfidenceeap}, respectively. We use darker colors to indicate lower confidence in this work. The highlighted regions of both figures are zoomed in to demonstrate the confidence difference in the blurred boundary regions. We can see that Fig.~\ref{fig:APEAPconfidenceeap} more clearly highlights the data points that are close to the boundary of another cluster when compared to Fig.~\ref{fig:APEAPconfidencemcltype}.  Furthermore, \eqref{eq:APEAPconfidence} requires less computations.

\subsection{Local Exemplars Connected to a Data Point}\label{subsec:APdensehist}
\begin{figure}[t]
	\centering 
	\subfigure[Equal density clusters]{\includegraphics[width=0.23\textwidth]{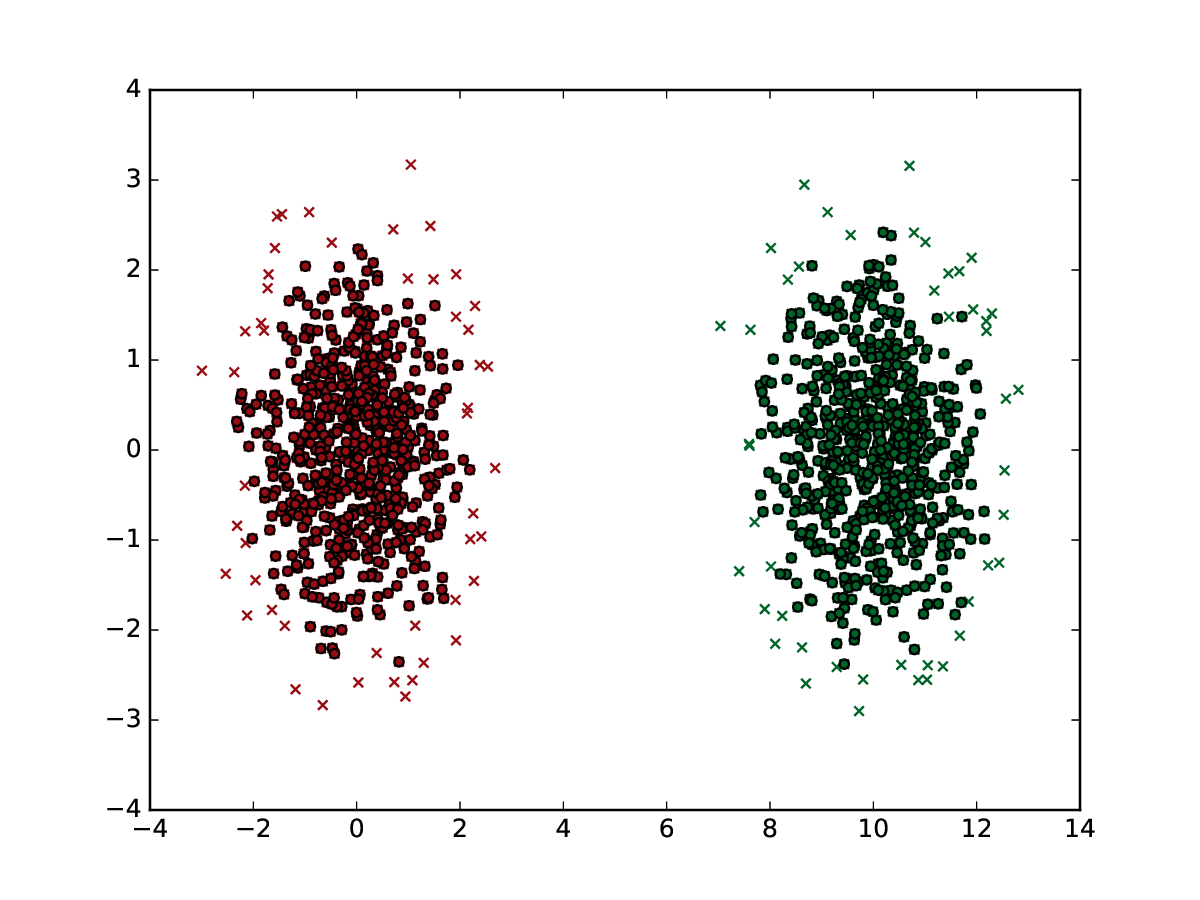}\label{fig:APeqblobexp}} \hfill
	\subfigure[LEC histogram for equal density clusters]{\includegraphics[width=0.23\textwidth]{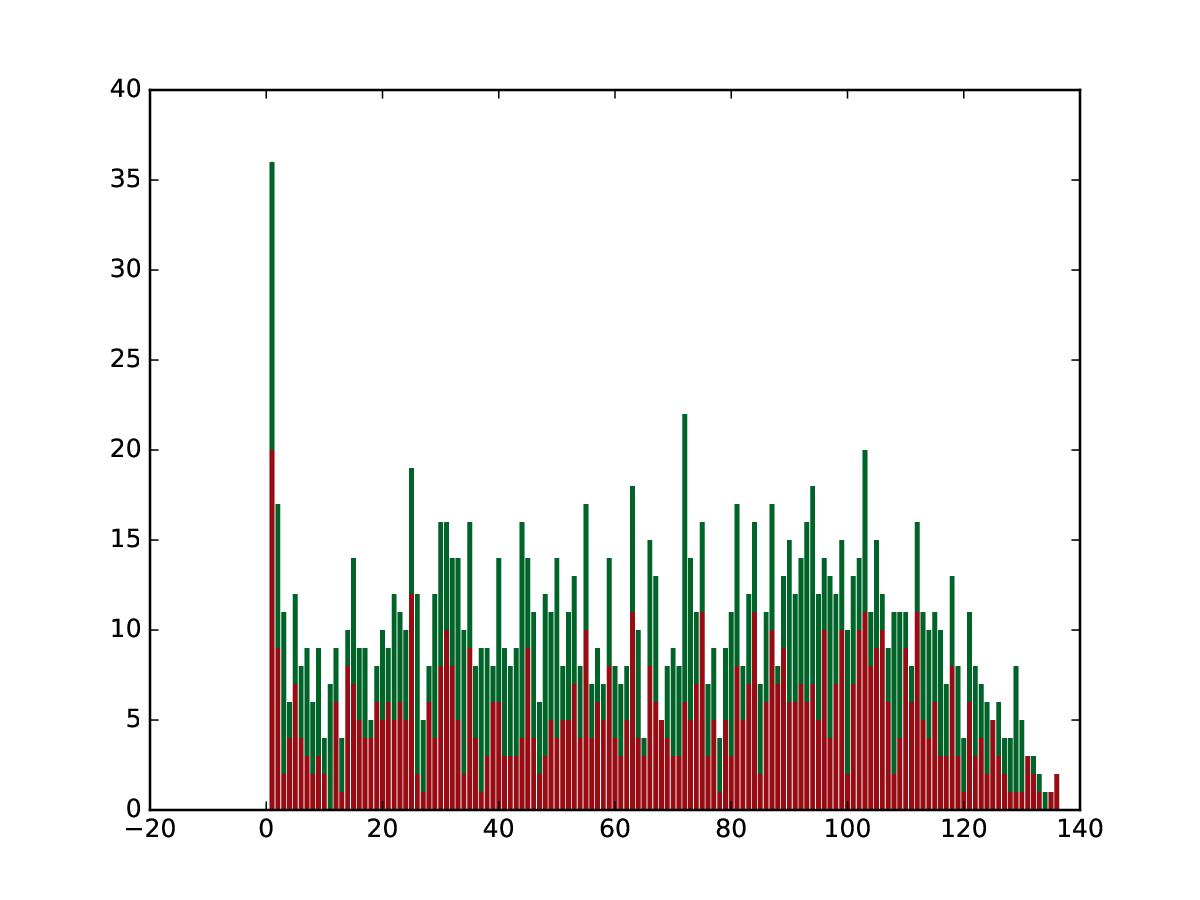}\label{fig:APeqblobhist}}
	\subfigure[Unequal density clusters]{\includegraphics[width=0.23\textwidth]{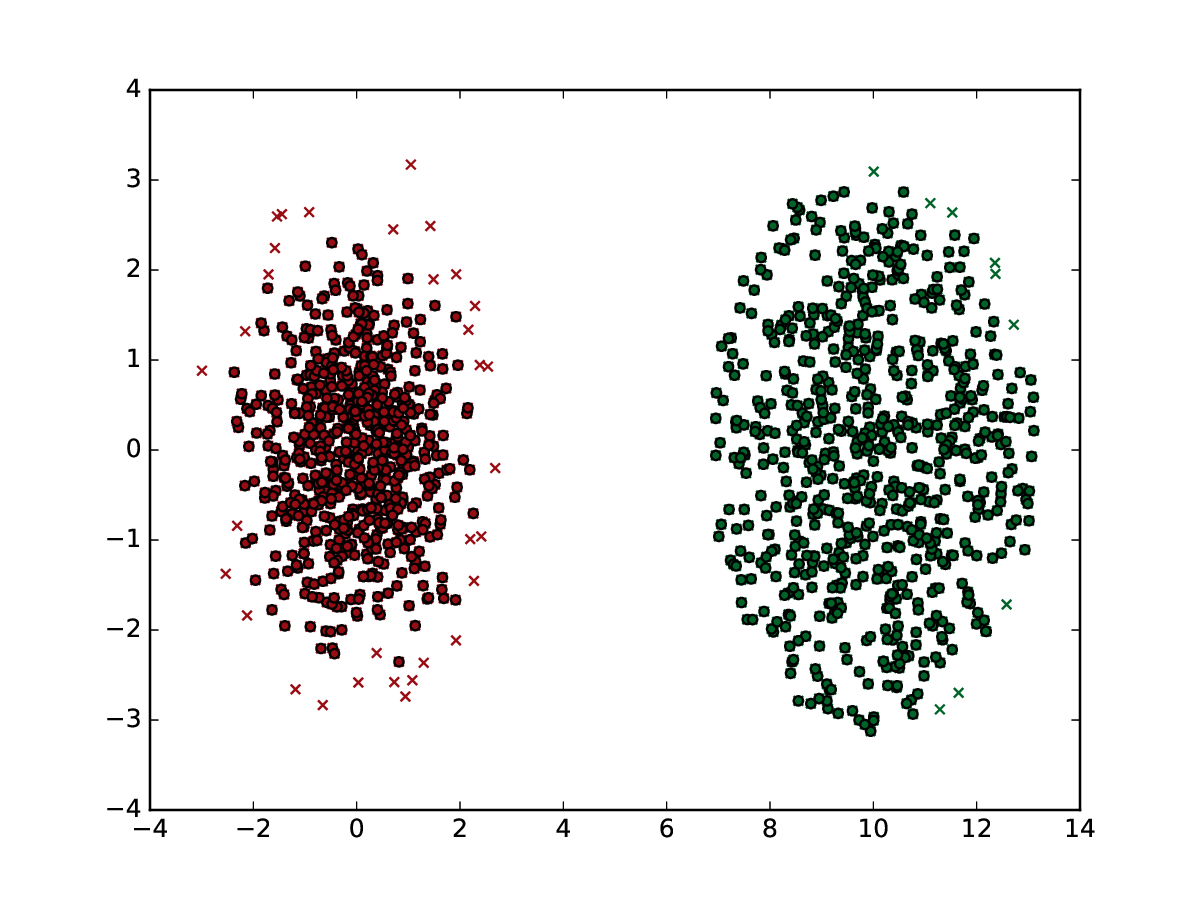}\label{fig:APblobunexp}}\hfill
	\subfigure[LEC histogram for unequal density clusters]{\includegraphics[width=0.23\textwidth]{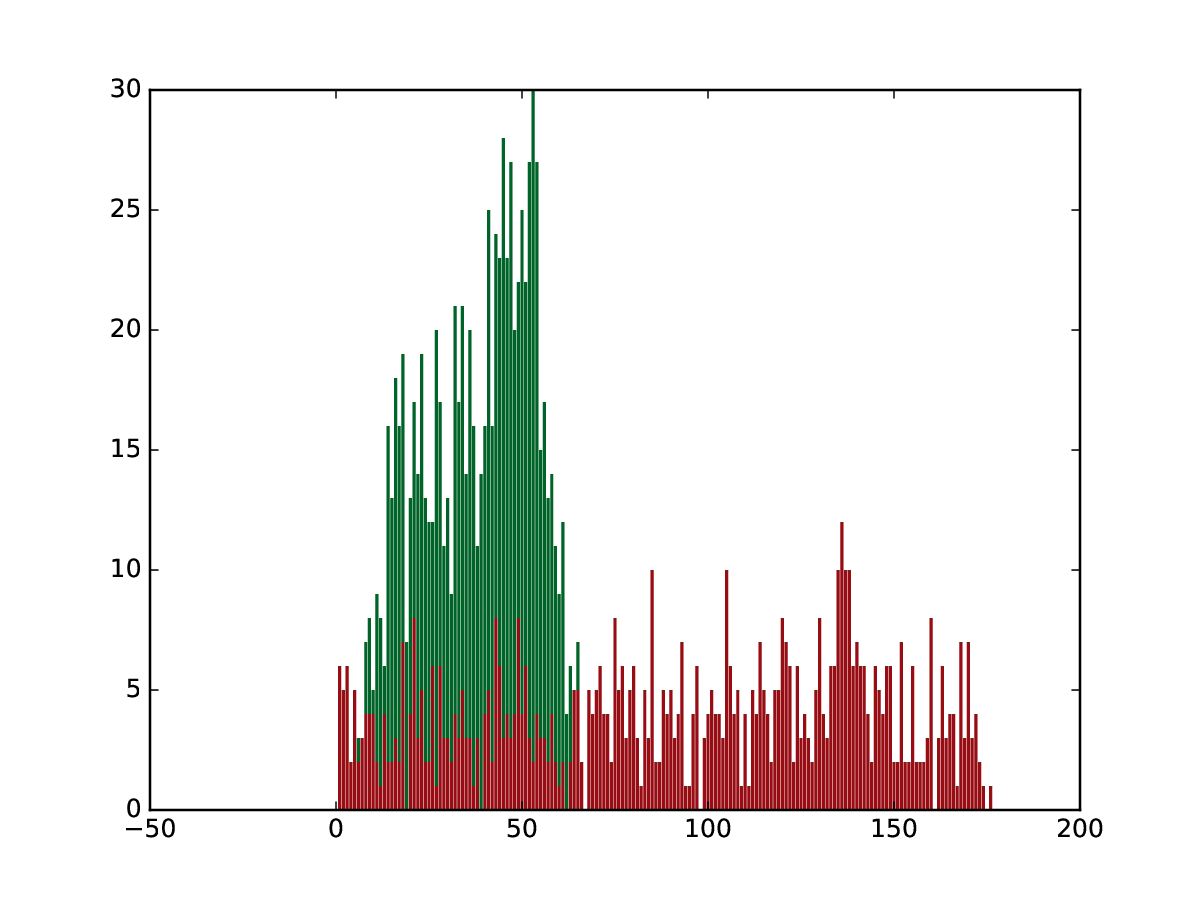}\label{fig:APblobhist}} 
	\caption{The figure demonstrates how LEC histogram can be used to identify different relative densities of the clusters when the $r_j$ constraints are inactive. These results were obtained for $p=0.6$, $q=0.95$ and $\Delta >1$.}
	\label{fig:EAPreldense} 
\end{figure}

When a data point is connected to more than one well separated local exemplars, this signifies that the data point shares a mix of properties of these local exemplars. The local exemplars connected to $x_i$ are given in $\mc{L}_i$ which can be computed in $\mc{O}(|\mc{E}|)$. Hence we can find the number of local exemplars connected to each data point in $\mathcal{O}(n|\mc{E}|)$. This information can be visualized using a histogram where the x-axis represents the number of local exemplars and the height of each bar represents the number of data points connected to these many local exemplars. We call this the LEC histogram. LEC histogram can be used for various purposes. It can allow for efficient hyperparameter tuning which we will discuss in Sec.~\ref{sec:APtuning}. It can also be used to recognize the relative densities of different clusters in a dataset. For this purpose we choose $\Delta >1$, i.e., $r_j$ constraints are rendered inactive. As mentioned in Sec.~\ref{sec:APEAP} this often has negligible impact on the global cluster discovery. Consider the datasets shown in Fig.~\ref{fig:APeqblobexp} and Fig.~\ref{fig:APblobunexp}, each with two clusters of equal and unequal densities, respectively. By looking at the LEC histograms we observe the following: the bars for the two clusters in Fig.~\ref{fig:APeqblobhist} are co-located, indicating that the two clusters in Fig.~\ref{fig:APeqblobexp} have similar density profiles. On the other hand, the bars for red cluster in Fig.~\ref{fig:APblobhist} are located further to the right of the bars for the green cluster, indicating that for red cluster the data points are connected to more local exemplars, which in turn implies a higher density of the red cluster in Fig.~\ref{fig:APblobunexp}.

\subsection{Inter Exemplar Connection Strength and Pruning}
\begin{figure}[t]
	\centering 
	% \subfigure[Clustering results with connections of each data point to local exemplars highlighted. ]{\includegraphics[width=0.23\textwidth]{./figures/global_and_local_view/halfmoon_weaklink-eps-converted-to}\label{fig:APweaklink}} \hfill
	\subfigure[IES histogram]{\includegraphics[width=0.23\textwidth]{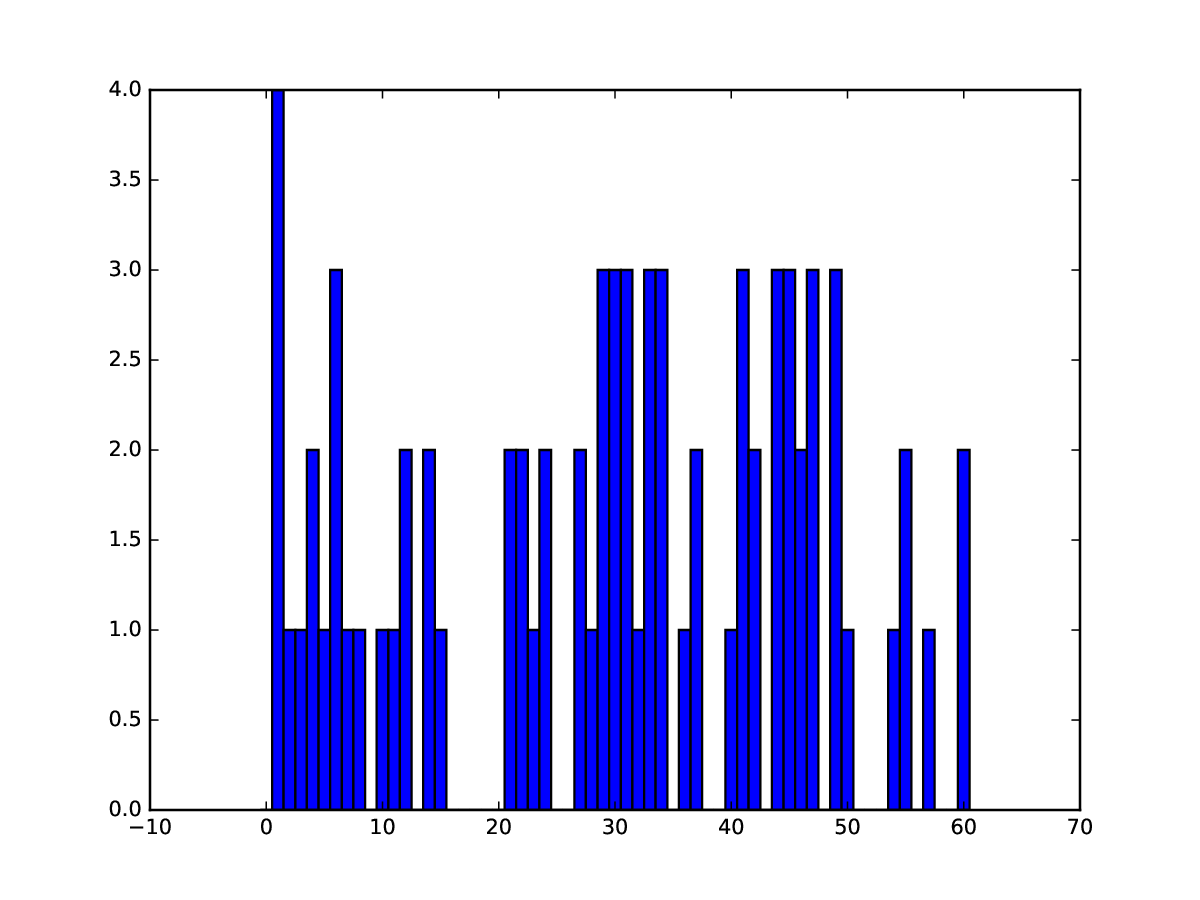}\label{fig:APweaklinkhist}}
	\caption{The figures depict the clustering results for half-moons dataset with a weakly connected region in one cluster and the corresponding IES histogram.}
	\label{fig:EAPweakregion} 
\end{figure}

Another approach to analyze individual clusters is to count the number of data points that form boundary connections between two local exemplars in the same cluster. This count can be considered as an indicator of local cluster strength and for neighbouring local exemplars this can indicate the confidence of EAP to merge the associated regions into one cluster. This information can then be used to highlight potential inconsistencies where EAP is not too certain about it's decision to merge the regions into one cluster. These few cases can then be evaluated by an analyst. We can again form a histogram from this information where, for each exemplar we can check it's connection strength to the closest $m$ local exemplars in the same cluster. Such a histogram is shown in Fig.~\ref{fig:APweaklinkhist} for the clustering results of  modified half-moons dataset in Fig.~\ref{fig:APweaklink}. The half-moons dataset has been modified in Fig.~\ref{fig:APweaklink} to contain a weakly connected region. The x-axis of the histogram represents the connection strength (i.e., number of shared data points between two local exemplars) and the height of the bar indicates the number of local exemplar pairs with the given connection strength. We call this the IES histogram. If there is a relatively low fraction of exemplar pairs in the left-most bars, these left-most bars indicate potential inconsistencies which may need to be examined by an analyst to decide if the exemplar pairs should be connected or the links between them should be pruned. For example, in Fig.~\ref{fig:APweaklinkhist}, there are only 4 pairs that have a single boundary connection between them. By inspecting these cases individually, we realize that 3 of them correspond to connections between relatively distant local exemplars and pruning the boundary connections between them does not have an impact on the clusters. The $4$th boundary connection provides the only link between the two subregions of the red cluster with a sparse region between them. An analyst can decide whether this data point and the corresponding local exemplars pair provide enough merit to connect otherwise two separate subsets of the dataset into one cluster. The computational complexity of constructing IES histogram is $\mathcal{O}(nm|\mc{E}|)$. We can also automate this process of pruning. The connections between all the pairs of local exemplars on the L.H.S. of a threshold, denoted by $n_t$ in the histogram can be pruned before finding the connected components in Alg.~\ref{algo:eap-dec}. $n_t$ can either be chosen by an analyst after having a look at the IES histogram or can be determined automatically, for example, by assigning $n_t$ a value such that, for example, $99\%$, of the local exemplar pairs in the IES histogram have higher number of boundary connections connecting them. For our synthetic and real world experiments in Sec.~\ref{sec:APrealworld} and Sec.~\ref{sec:APsynthetic} we use $n_t=3$ independent of the dataset and other settings. 
\section{Hyperparameter Tuning}\label{sec:APtuning}
EAP has following three hyperparameters:
\begin{itemize}
	\item 	The self preference $p$ serves a similar purpose as in AP, i.e., it indicates the interest of a data point to become a local exemplar, forming a spherical subcluster around it. A higher value of $p$ motivates more points to become local exemplars whereas a low value works to suppress potential local exemplars. For EAP, $p$ takes normally a higher value, i.e., often between $55$ and $65$  percentile of the pairwise similarity values in $\Svec$, than AP where $p$ is normally equal to the median of $\Svec$ \cite{affinitypropagation}. This is because we want to motivate enough local exemplars to be distributed all over the dataset, forming subclusters which are later merged using Alg.~\ref{algo:eap-dec} based on the boundary connections between the local exemplars.
	\item The linkage penalty $q$ defines the maximum penalty (per extra local exemplar) that a data point has to pay for connecting to more than one local exemplar. A higher penalty implies fewer boundary connections being formed between local exemplars. As we want a data point to connect to multiple local exemplars only if it is close enough to all of them, subsequently helping in global structure discovery, so we use a quite high penalty, e.g., in range of negative of  $96$ to $98$ percentile of $\Svec$. 
	\item  The separation radius $\Delta$ defines a neighbourhood around each local exemplar such that there should be no other local exemplar appearing in this neighbourhood. The lower the value of $\Delta$, the bigger are the neighbourhoods $\mc{N}_i$. Since the aim of introducing $r_j$ is just to counteract the effect of loosened $g_i$ constraint, $\Delta$ is normally chosen to be a high value, e.g., in range from $98.5$ to $99.5$ percentile of $\Svec$, hence suppressing other potential local exemplars in only a close by neighbourhood.
\end{itemize}
The range in which these hyperparameters assume values are highly dependent on the pairwise similarity being used and other factors,e.g., the size of the dataset. The typical ranges mentioned earlier are, for example, suitable for euclidean norm based  pairwise similarities or  for probabilistic notion of an edge weight between two nodes in a graph. In this work we specify the values of the hyperparameters in terms of percentiles of $\Svec$. We also use the notation $\Delta>1$ to indicate that $\mc{N}_j= \{j\}$, for all $j$, hence $r_j$ constraints are rendered inactive in this case. 

Despite the relatively high number of hyperparameters we can perform efficient successive tuning of the hyperparameters easily to obtain desired results in a few steps. This is due to the following reasons:
\begin{itemize}
	\item It is easy and intuitive to understand the impact of changing any of the three hyperparameters on the clustering outcome. 
	\item The abundance of insightful information, beyond just cluster assignments, provided by EAP at each step helps the analyst in deciding how to adapt the hyperparameters to improve the results.
\end{itemize}
Usually, we successively tune $p$, then $q$ and finally $\Delta$. Some ways to utilize local information to guide the subsequent steps in hyperparameter tuning are as follows:
\begin{itemize}
	\item \textbf{Too few local exemplars:} Assuming that $\Delta$ is in an acceptable range, this implies that we have set $p$ too low and we should increase it such that we obtain enough local exemplars that are spread throughout the dataset. 
	\item \textbf{Boundary connections:} Once we have adequate number of local exemplars available, we focus on building boundary connections between the local exemplars for global structure discovery. For this we use LEC and IES histograms to tune $q$ as follows
	      \begin{enumerate}
	      	\item \textbf{LEC and IES shifted too much to the left:} This configuration of histograms implies that the linkage penalty is too high and hence not allowing sufficient boundary connections to form between the local exemplars. This will also manifest itself in terms of many more clusters than expected. In this scenario we should increase $q$ to decrease the linkage penalty so that more boundary connections can appear. 
	      	\item \textbf{LEC and IES shifted too much to the right:} This scenario is the opposite of the previous one. Due to the low linkage penalty data points also get connected to far off local exemplars (which possibly may belong to a different \enquote{ground truth} cluster). This can result in too few clusters than expected and violates the basic principle of introducing the penalty $q$ that we want to form only boundary connections between close enough local exemplars to facilitate global cluster discovery. Allowing points to connect to even far off local exemplars rather distorts the global cluster discovery by even merging well separated clusters. In this scenario we should increase the linkage penalty (decrease $q$). 
	      \end{enumerate}
	\item \textbf{Local information:} Finally we can tune $\Delta$ to obtain the desired type of local information. If we believe that the discovered local exemplars are too close, we can decrease $\Delta$, increasing the neighbourhood radius around each local exemplar. This will results in fewer and farther away local exemplars and a shift of both LEC and IES towards left but as long as $\Delta$ is not decreased too much the global clusters discovered remain unaffected, as was also illustrated in Fig.~\ref{fig:EAPlocalviewneighimpact}. The sole purpose of $\Delta$ should be to adapt the local information, hence normally, as long as one starts the tuning process with a reasonable value of $\Delta$ there is often no need to change it later in the successive tuning process.
	\item \textbf{Confidence values:} The confidence values discussed in Sec.~\ref{subsec:localexemp} can also be used to determine if the results obtained for a specific choice of hyperparameters are satisfactory. If there are too many data points with low enough confidence, this often implies that the algorithm has not been able to form boundary connections to merge subclusters where it should have. In this case we need to reduce the linkage penalty to merge these subclusters. On the other hand, when there is a relatively small number of points that exhibit low confidence, these points may be manually analyzed to see if they should be considered to form a bridge to connect two currently separate clusters to merge them.
\end{itemize}
In Appendix~\ref{sec:tuningexp} we will present a step by step example of doing hyperparameter tuning using these insights. We also briefly discuss about the sensitivity of the global structure discovery to parameter variations in Appendix~\ref{sec:tuningsensitive}. 

Finally we want to emphasize that all the results presented in this work, unless explicitly stated otherwise, are obtained via successive tuning of hyperparameters rather than any kind of parameter sweeps. Hence these results are a good representation of what one can expect in normal application of EAP and do not correspond to the ceiling performance on the respective datasets. Furthermore, we also notice that the final results shown in Sec.~\ref{sec:APsynthetic} and Sec.~\ref{sec:APrealworld} correspond to hyperparameter values that are very similar for different datasets although the datasets have very different characteristics. This also points to the ease of finding suitable hyperparameters for a wide variety of datasets with different cluster characteristics by exploring a very narrow range of hyperparameter values, as long as the datasets employ the same (or similar) pairwise similarity metric.
\begin{table*}[t]
	\centering
	\begin{tabular}{|c||c|p{1.6cm}|p{2.5cm}|c|c|c|c|c|} 
		\toprule
		\textbf{Dataset}     & \textbf{Points} & \textbf{GT clusters} & \textbf{Params}    & \textbf{Sn} & \textbf{PPV} & \textbf{Acc} & \textbf{NMI} & \textbf{ARI} \\
		\hline
		\hline
		\textbf{Aggregation} & 788             & 7                    & $q= -0.97$, $p=0.5$ & 0.995       & 0.995        & 0.995        & 0.985        & 0.990        \\
		\hline
		\textbf{Flame}       & 240             & 2                    & $q= -0.95$, $p=0.6$ & 0.983       & 0.991        & 0.987        & 0.9          & 0.955        \\
		\hline
		\textbf{R15}         & 600             & 15                   & $q= -0.97$, $p=0.6$ & 0.992       & 0.992        & 0.992        & 0.988        & 0.982        \\
		\hline
		\textbf{Circles}     & 1500            & 2                    & $q= -0.97$, $p=0.6$ & 1           & 1            & 1            & 1            & 1            \\
		\hline
		\textbf{Spiral}      & 312             & 3                    & $q= -0.96$, $p=0.6$ & 1           & 1            & 1            & 1            & 1            \\
		\hline
		% \textbf{Jain}           & 373   & 2     & 1.000 & 1.000 & 1.000\\
		%  \hline
	\end{tabular}
	\caption{Quantitative global performance evaluation of EAP on synthetic datasets along with the final hyperparameters obtained via successive tuning.}
	\label{tab:APEAPsynthetic}
\end{table*}
\section{Synthetic Experiments}\label{sec:APsynthetic}
We will use various synthetic datasets to analyze how well EAP performs in terms of cluster assignment and how it highlights other useful information about the datasets. These datasets lie in $\mathbb{R}^2$ and the clusters are distinguishable in a Euclidean sense. Therefore, we use the negative of Euclidean distance between the data points as pairwise similarity metric. Note that since the aim of EAP is to discover the structure in the provided $\Svec$, discussing synthetic datasets that lie in $\mathbb{R}^2$ and are all separable in a Euclidean sense does not limit the generality of our experiments. This is because the datasets lead to different structures in $\Svec$ due to the varied nature of the clusters. If the same $\Svec$ was the outcome of computations based on some other pairwise similarity metric for some other dataset (possibly not lying in a euclidean space), EAP will discover the same structure in the data. Datasets lying in $\mathbb{R}^2$ that are separable in a euclidean sense just allow us to illustrate our observations in a more lucid way without invoking special domain knowledge to interpret the results.   

Table~\ref{tab:APEAPsynthetic} shows the quantitative evaluation of cluster assignments obtained via EAP, when compared to the ground truth based on euclidean separation. We present the results for Sensitivity (Sn), Positive Predictive Value (PPV), Accuracy (Acc, geometric mean of Sn and PPV), Normalized Mutual Information (NMI) and Adjusted Rand Index (ARI). We emphasize again that these results are obtained using the successive tuning procedure described in Sec.~\ref{sec:APtuning} and Appendix~\ref{sec:tuningexp}, not by parameter sweeps. For synthetic experiments we do not present a comparison with other clustering algorithms due to the following reasons
\begin{itemize}
	\item For the global discovery, EAP obtains nearly perfect results on these synthetic datasets. Therefore, a comparison with other algorithms can only reveal their weaknesses in discovering certain global structures when compared to EAP. These weaknesses are well documented for the well known algorithms and we do not want to reiterate them in this work. Rather our focus is to  the highlight the findings of EAP. 
	\item The algorithms that provide local information about individual clusters, for example AP, mostly only provide local exemplars. These algorithms also suffer from a lack of global structure discovery, rendering them not too effective for clustering in general unless the dataset consists of only spherical clusters. Therefore comparing them to EAP in terms of local information when they cannot discover the global structure in the dataset does not lead to an interesting comparison. 
\end{itemize}
In the following, we will discuss various additional interesting aspects, beyond global structure,  discovered by EAP for each dataset individually. The confidence values shown in this section are computed using  \eqref{eq:APEAPconfidence}
\subsection{Aggregation \cite{aggregation}}
\begin{figure}[t]
	\centering 
	\subfigure[Clustering results with exemplars for $\Delta =0.99$]{\includegraphics[width=0.23\textwidth]{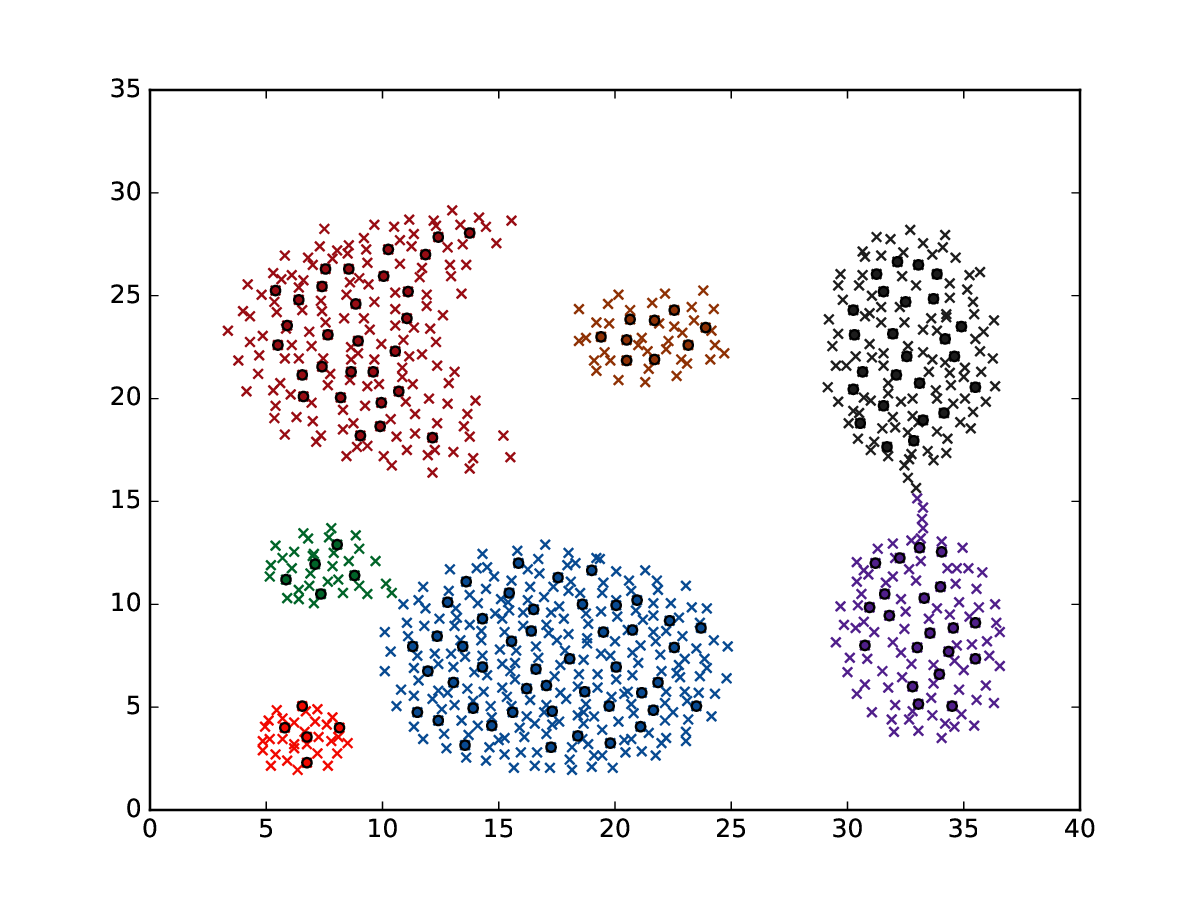}\label{fig:EAPsynaggregation1}} \hfill
	\subfigure[LEC histogram for $\Delta >1$]{\includegraphics[width=0.23\textwidth]{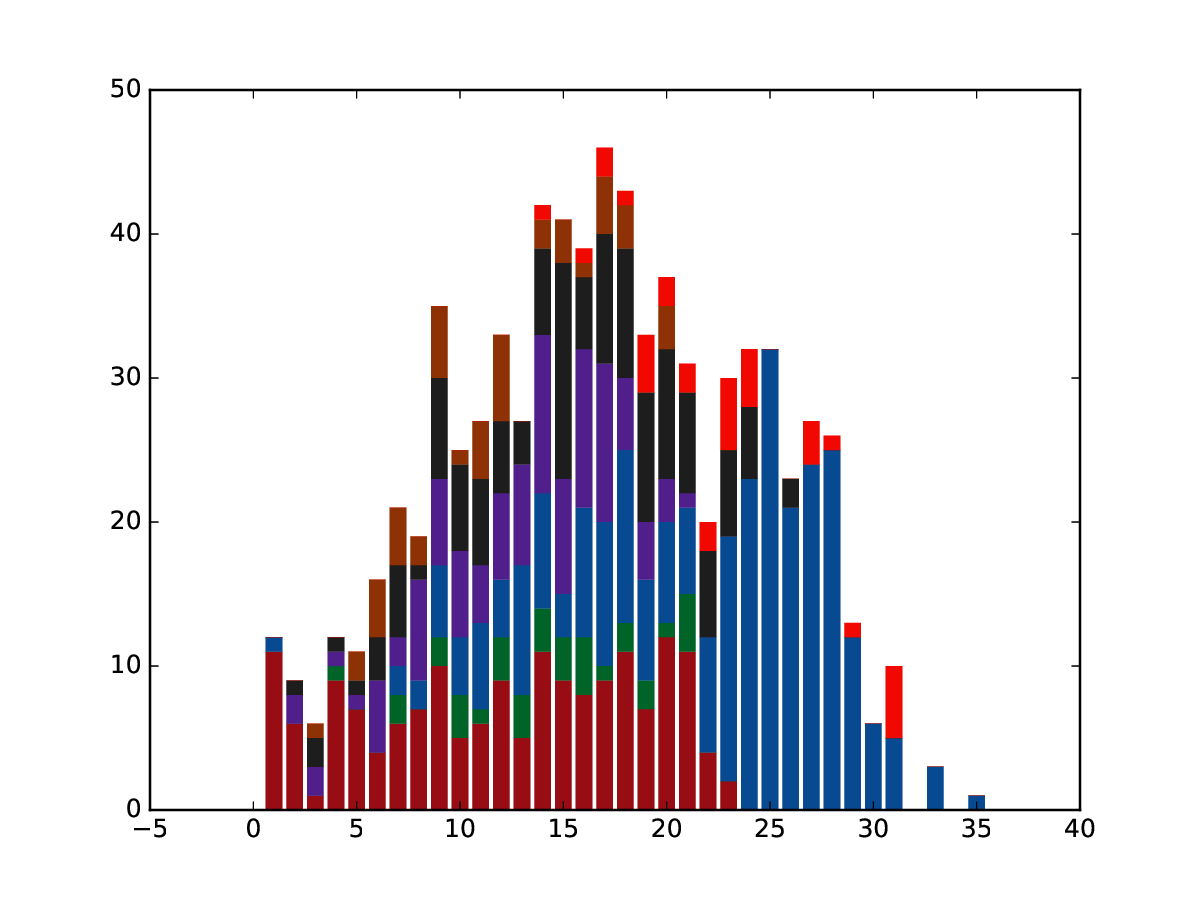}\label{fig:EAPsynaggregation6}} 
	\subfigure[Confidence values for $\Delta=0.995$]{
		\begin{tikzpicture}[spy using outlines={rectangle,magnification=3,size=3cm, connect spies}]
			\node[anchor=south west] at (0,0) {\includegraphics[width=0.5\textwidth]{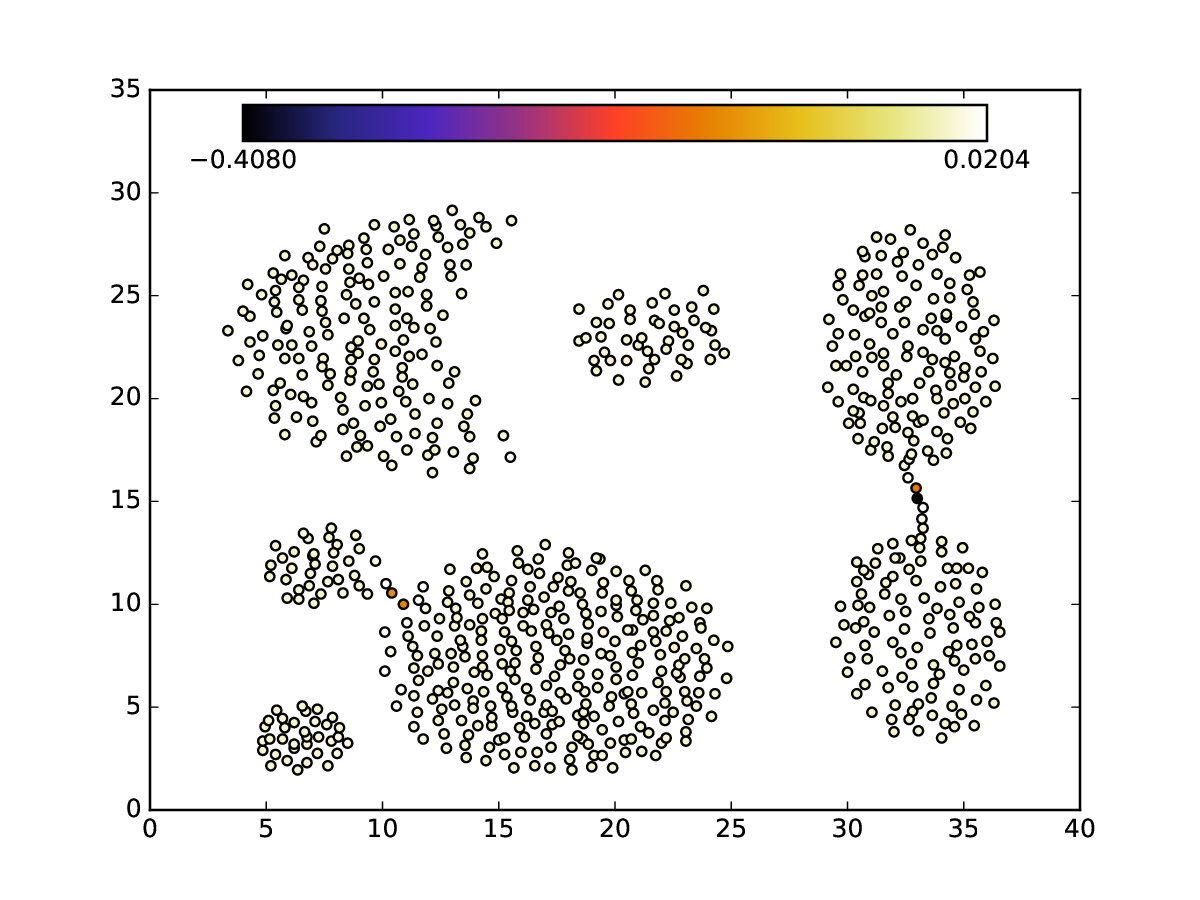}};
			\spy[red, every spy on node/.append style={thick}] on (3.2,2.5) in node [left] at (4.3, -1.5);
			\spy[blue, every spy on node/.append style={thick}] on (7.2,3.3) in node [left] at (8.3, -1.5);
		\end{tikzpicture}
		\label{fig:EAPsynaggregation3}
	} 
	\caption{EAP results for Aggregation dataset where $p=0.5$ and $q=0.97$. In (c) the noisy bridges between the clusters have been zoomed in to.}
	\label{fig:EAPsyntheticaggregation} 
\end{figure}

The important aspects of this dataset include the ground truth clusters of varying sizes as well different densities. Some of the ground truth clusters are also linked by small \enquote{noise bridges}. EAP is able to handle all these factors while determining cluster assignment. The data points forming the noisy bridges between clusters are assigned low confidence values as shown in Fig.~\ref{fig:EAPsynaggregation3} with boundary points zoomed in to outline confidence differences. Furthermore, if we apply EAP with $p$ and $q$ as specified in Table~\ref{tab:APEAPsynthetic} with $\Delta>1$, the LEC histogram so obtained, shown in Fig.~\ref{fig:EAPsynaggregation6}, clearly shows that the blue cluster has higher density  when compared to other discovered clusters as the data points corresponding to the blue clusters dominate the R.H.S. of the LEC histogram.

\subsection{Flame \cite{flame}}
\begin{figure}[tbh]
	\centering 
	\subfigure[Clustering results with exemplars for $\Delta =0.99$]{
		\begin{tikzpicture}
			\node[anchor=south west] at (0,0) {\includegraphics[width=0.5\textwidth]{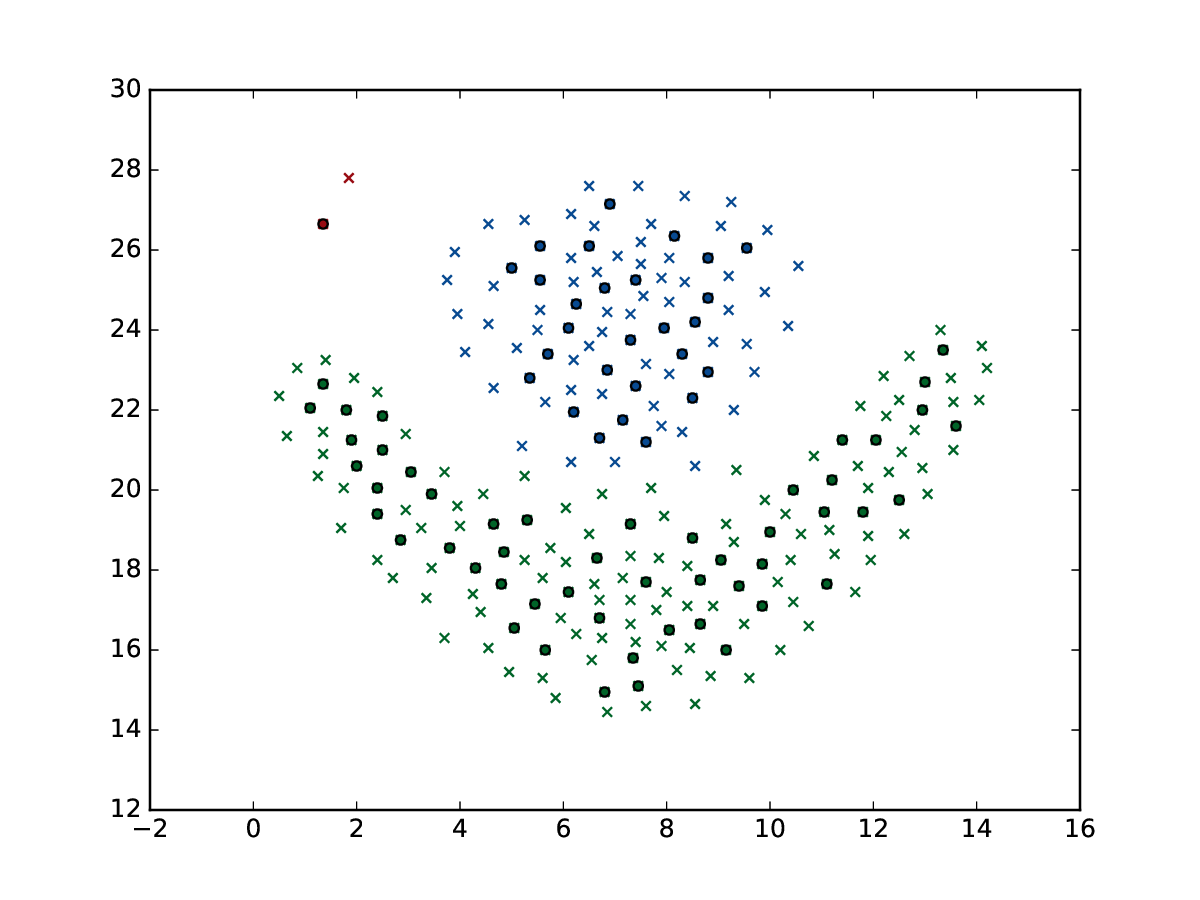}\label{fig:EAPsynflame1}};
		\end{tikzpicture}
	}
	\subfigure[Confidence values for $\Delta=0.99$]{
		\begin{tikzpicture}[spy using outlines={rectangle,magnification=3,width=6cm, height=2cm, connect spies}]
			\node[anchor=south west] at (0,0) {\includegraphics[width=0.5\textwidth]{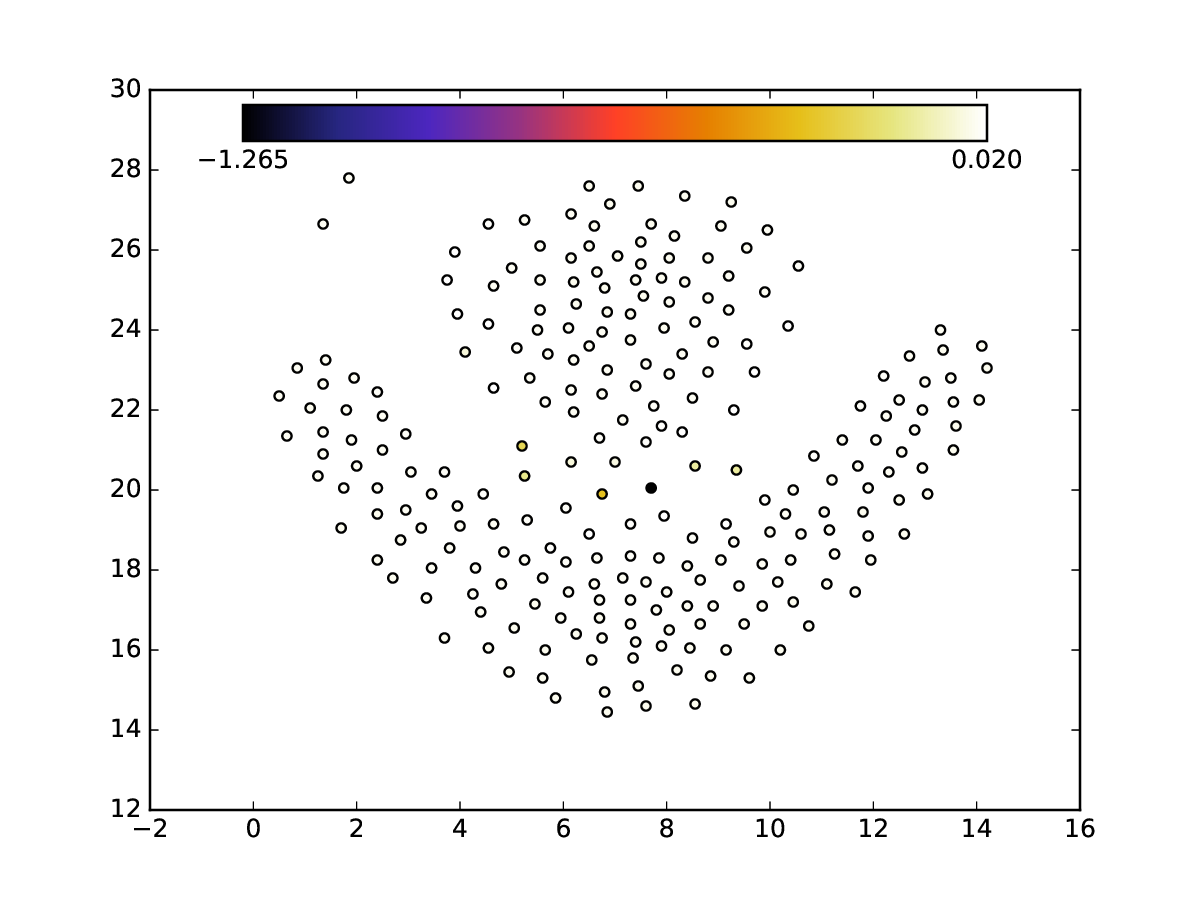}};
			\spy[red, every spy on node/.append style={thick}] on (4.8,3.3) in node [left] at (7.8, -1.5);
		\end{tikzpicture}
		\label{fig:EAPsynflame3}
	}
	\caption{EAP results for Flame dataset where $p=0.6$ and $q=-0.95$. In (b) the blurry boundary between the two clusters has been zoomed in.}
	\label{fig:EAPsyntheticflame} 
\end{figure}[t]
The dataset has a relatively long noisy boundary between the two non linearly separable unequal sized clusters. EAP not only discovers the clusters correctly but also highlights the noisy boundary between the two clusters as potential inconsistencies in the confidence plot Fig.~\ref{fig:EAPsynflame3}. These boundary points have been zoomed in to illustrate their low confidence values. The dataset also contains two outliers (visible in the upper left corner of  Fig.~\ref{fig:EAPsynflame1}). EAP recognizes the outliers and creates a separate cluster for these two nearby outliers (indicated by red colour in Fig.~\ref{fig:EAPsynflame1}). Note that the these two data points are considered a part of the upper cluster in the ground truth, reflecting on the subjective nature of clustering problem, as from a visual perspective they appear to be outliers. 
\subsection{R15  \cite{r15}}
\begin{figure}[t]
	\centering 
	\subfigure[Clustering results with exemplars for $\Delta =0.99$]{
		\begin{tikzpicture}
			\node[anchor=south west] at (0,0) {\includegraphics[width=0.5\textwidth]{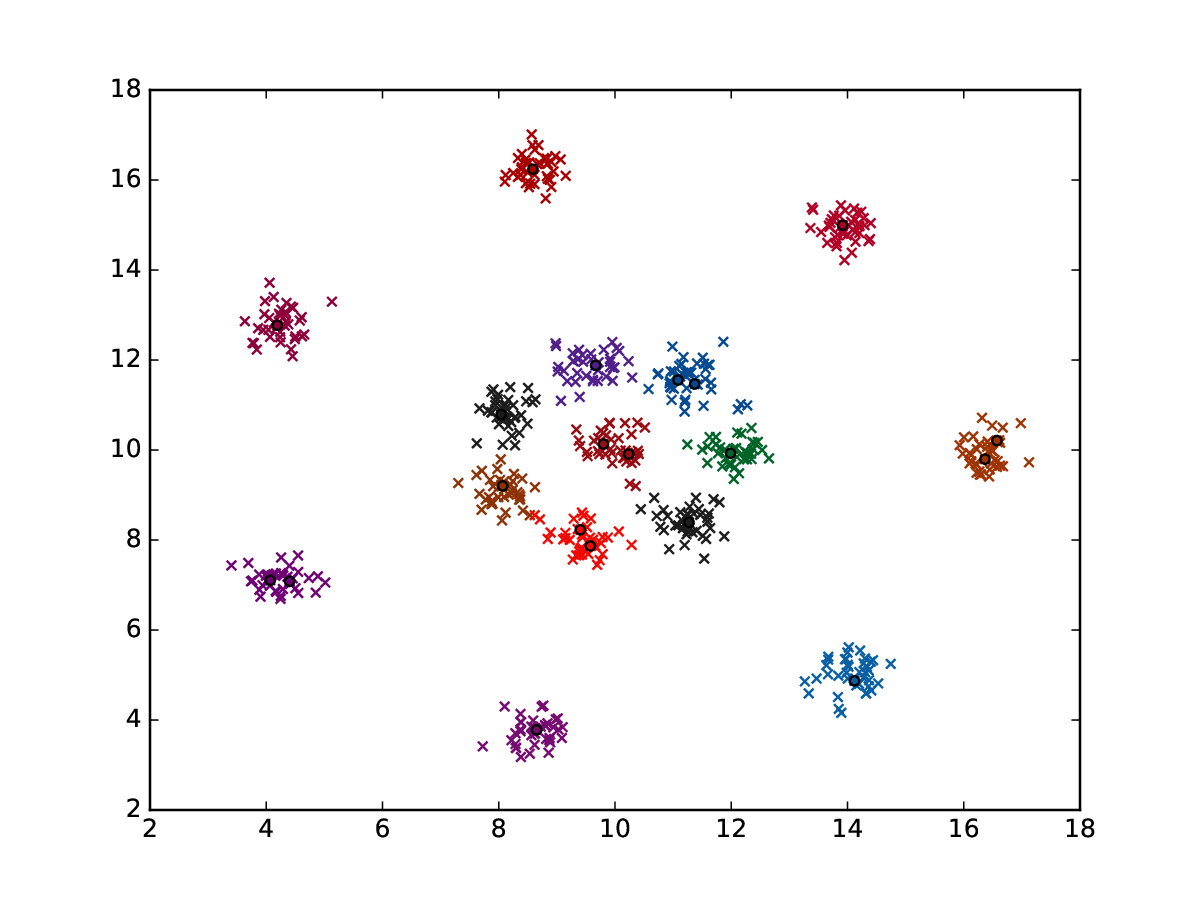}};
		\end{tikzpicture}
		\label{fig:EAPsynR151}
	}
	\subfigure[Confidence values for $\Delta=0.99$]{
		\begin{tikzpicture}[spy using outlines={circle,magnification=2.5,size=1cm, connect spies}]
			\node[anchor=south west] at (0,0) {\includegraphics[width=0.5\textwidth]{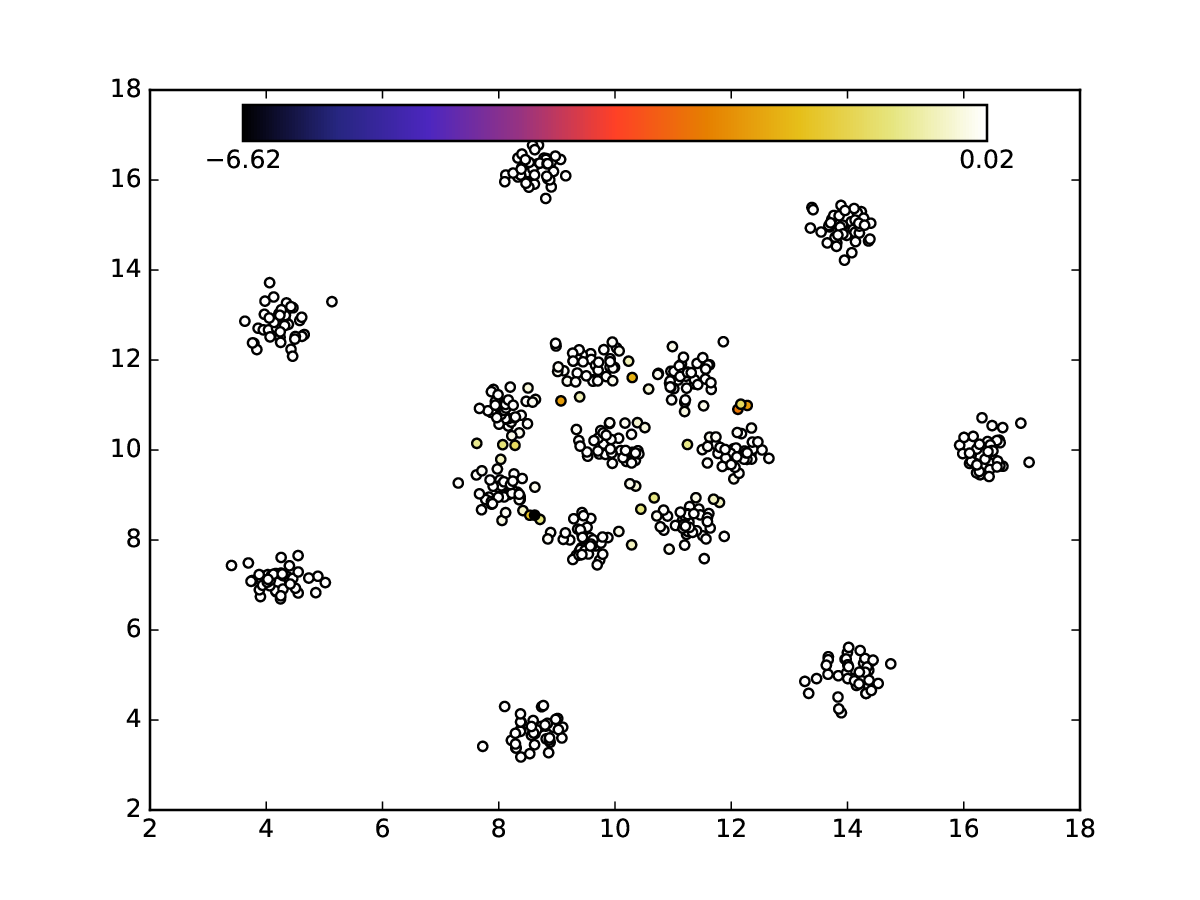}};
			\spy[red, every spy on node/.append style={thick}] on (5.7,3.8) in node [left] at (8, 5);
			\spy[blue, every spy on node/.append style={thick}] on (4.2,3) in node [left] at (3, 1.5);
			\spy[green, every spy on node/.append style={thick}] on (4.3,3.9) in node [left] at (3.8, 5);
			\spy[purple, every spy on node/.append style={thick}] on (3.9,3.6) in node [left] at (3, 3.5);
			\spy[black, every spy on node/.append style={thick}] on (5,4.1) in node [left] at (5.5, 5);
			\spy[brown, every spy on node/.append style={thick}] on (5.1,3.1) in node [left] at (5.5, 1.7);
		\end{tikzpicture}
		\label{fig:EAPsynR153}
	}
	\subfigure[LEC histogram for $\Delta =0.99$]{\includegraphics[width=0.23\textwidth]{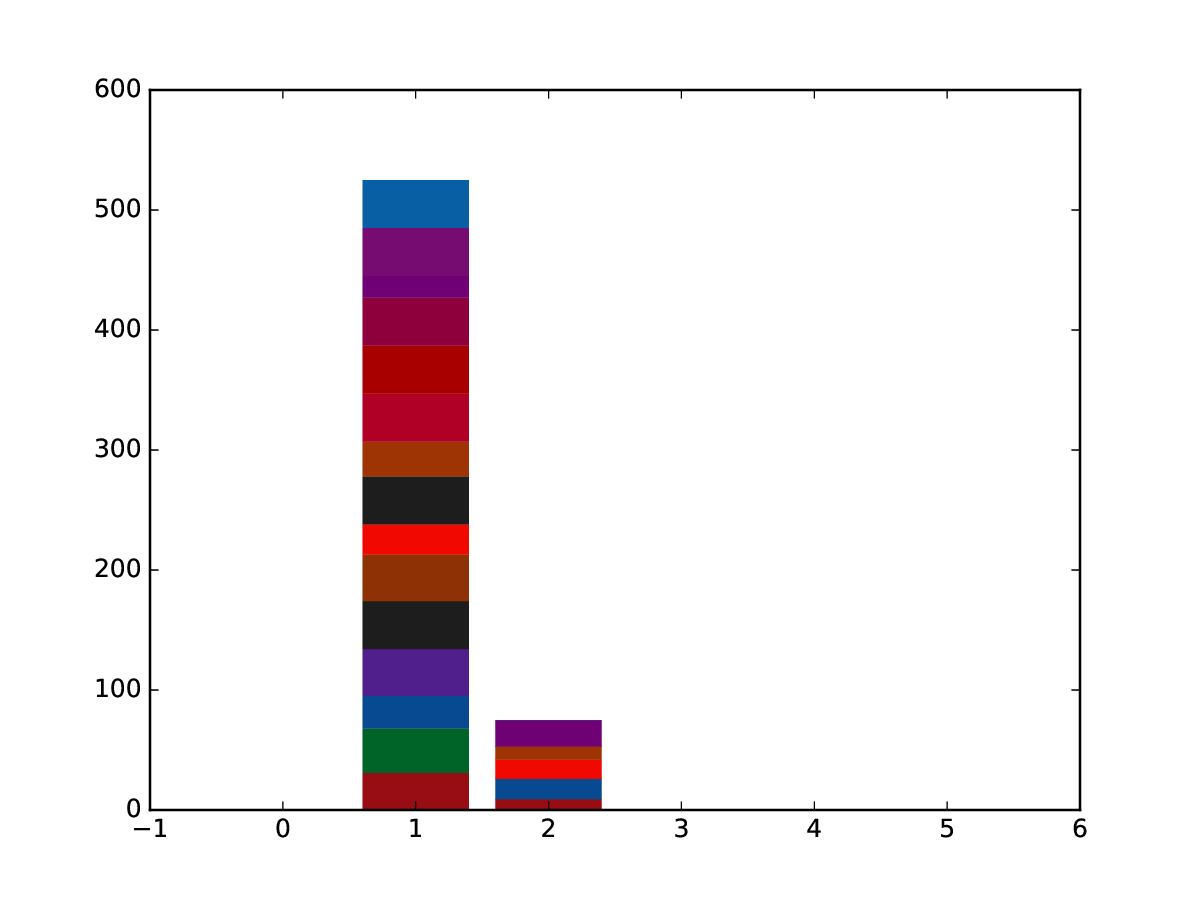}\label{fig:EAPsynR155}}  \hfill
	\subfigure[IES histogram for $\Delta=0.99$]{\includegraphics[width=0.23\textwidth]{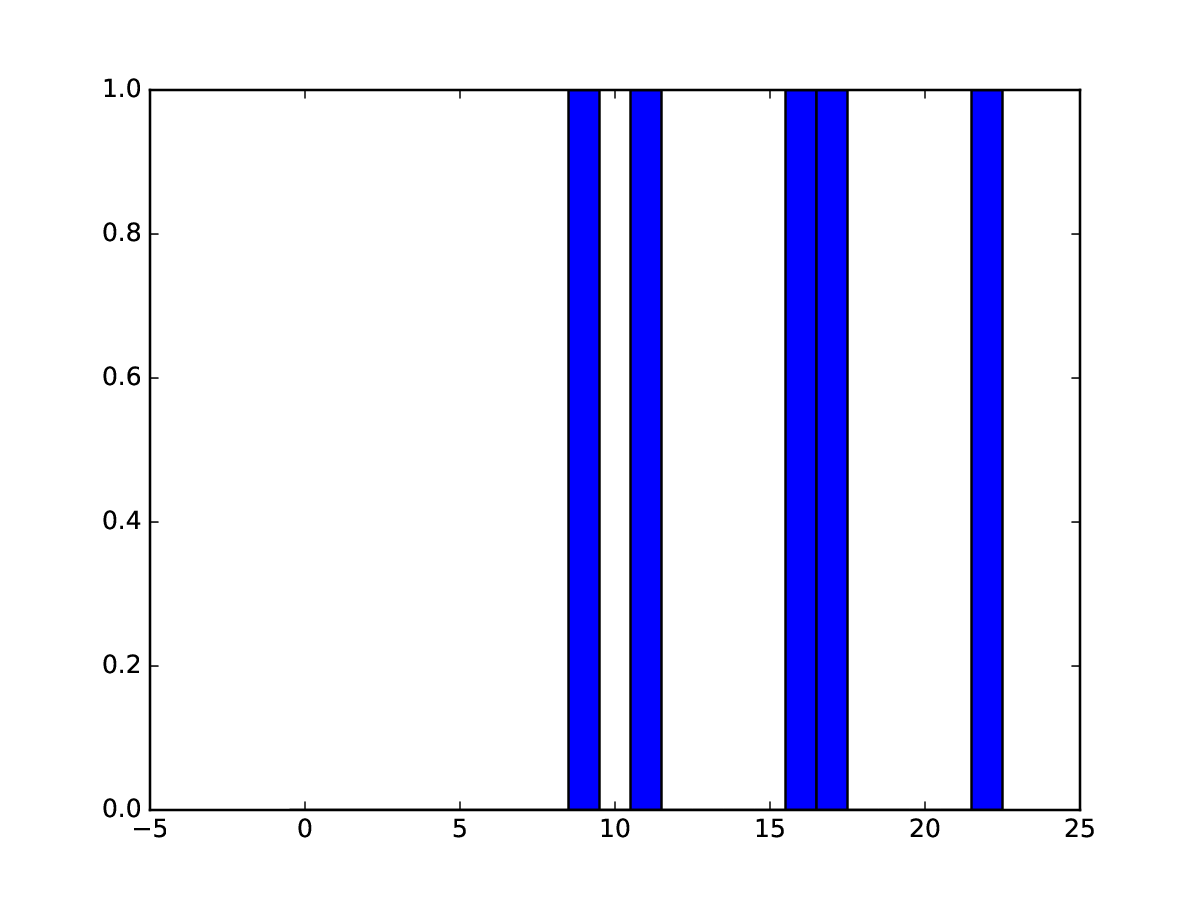}\label{fig:EAPsynR154}} 
	\caption{EAP results for R15 dataset where $p=0.6$ and $q=-0.97$. In (b) the boundary regions are zoomed into.}
	\label{fig:EAPsyntheticR15}
\end{figure}
The clusters in R15 are approximately spherical. Hence most of them are adequately representable by a single local exemplar. This is clearly visible in LEC histogram, shown in Fig.~\ref{fig:EAPsynR155}, where most of the points are connected with only one local exemplar and also in IES, shown in Fig.~\ref{fig:EAPsynR154}, where only there are only $5$ local exemplar pairs that share boundary connections. Hence LEC histogram in Fig.~\ref{fig:EAPsynR155} and IES histogram in Fig.~\ref{fig:EAPsynR154} indicate the presence of mainly relatively small spherical clusters in the dataset. Looking at the confidence values in Fig.~\ref{fig:EAPsynR153}, we can also infer that for some clusters (the ones in the outer radius), the neighbouring clusters are far enough as the confidence values for all data points in these clusters are high, whereas for some other clusters (the ones in the inner radii), the neighbouring clusters are much closer, indicated by the lower confidence values of some of the points in these clusters. 
\begin{figure}
	\centering 
	\subfigure[Clustering results with exemplars for $\Delta =0.99$]{\includegraphics[width=0.23\textwidth]{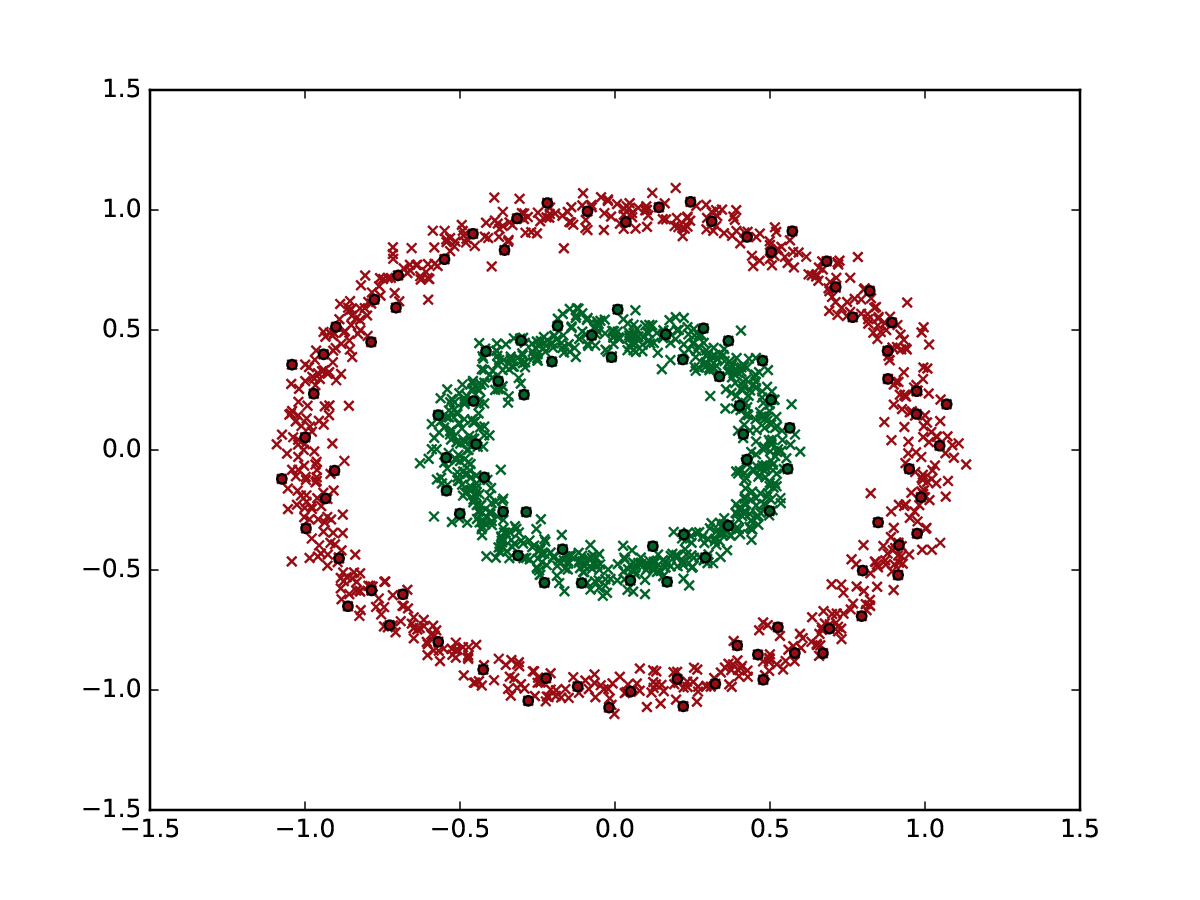}\label{fig:EAPsyncircles1}} \hfill
	\subfigure[LEC histogram for $\Delta >1$]{\includegraphics[width=0.23\textwidth]{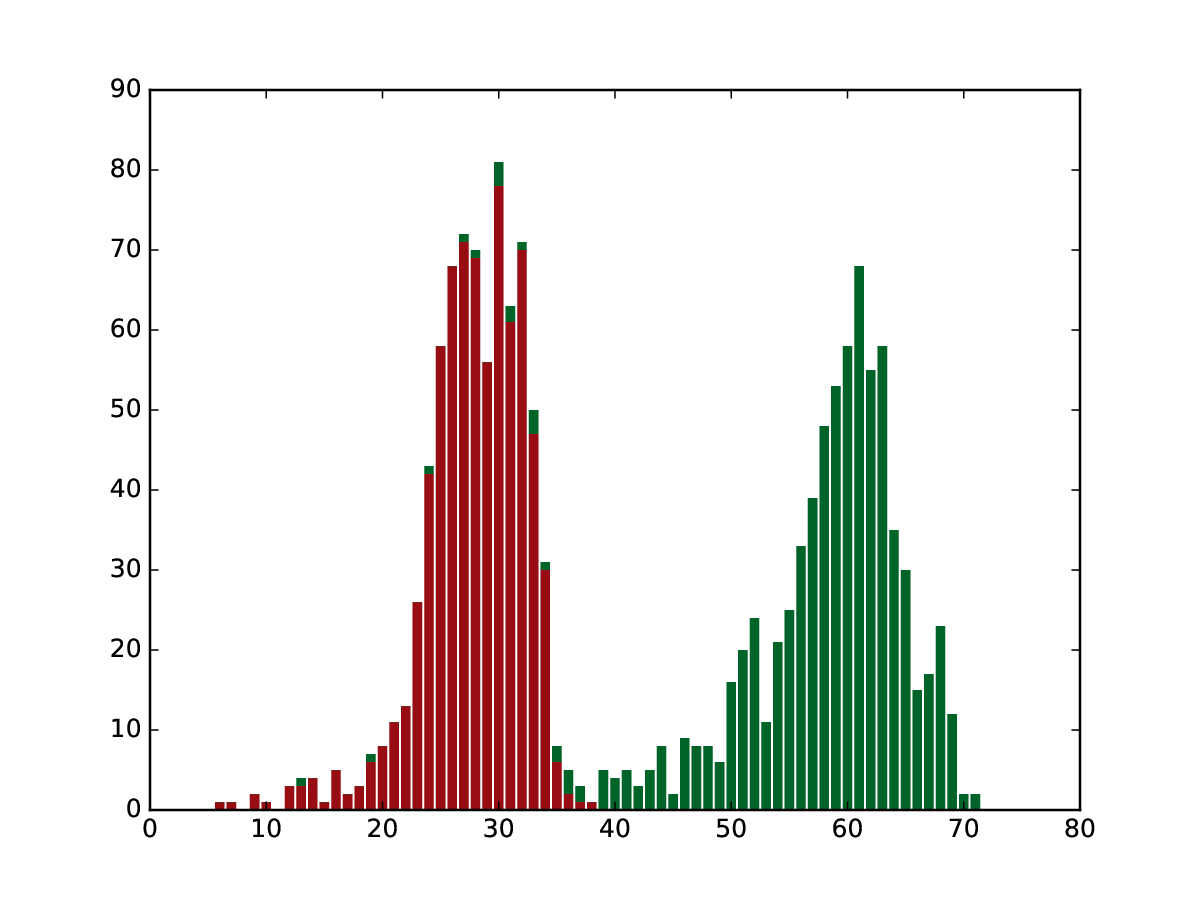}\label{fig:EAPsyncircles6}} 
	\caption{EAP results for concentric circles dataset where $p=0.6$ and $q=-0.97$}
	\label{fig:EAPsyntheticcircles} 
\end{figure}
\subsection{Concentric Circles}
The dataset consists of two concentric ring shaped equal sized clusters that have different densities. The LEC histogram in Fig.~\ref{fig:EAPsyncircles6}, obtained for $\Delta>1$, clearly highlights that the green cluster has a higher density. 
\subsection{Spirals \cite{spiral}}
\begin{figure}[t]
	\centering 
	\subfigure[Clustering results with exemplars for $\Delta =0.99$]{\includegraphics[width=0.23\textwidth]{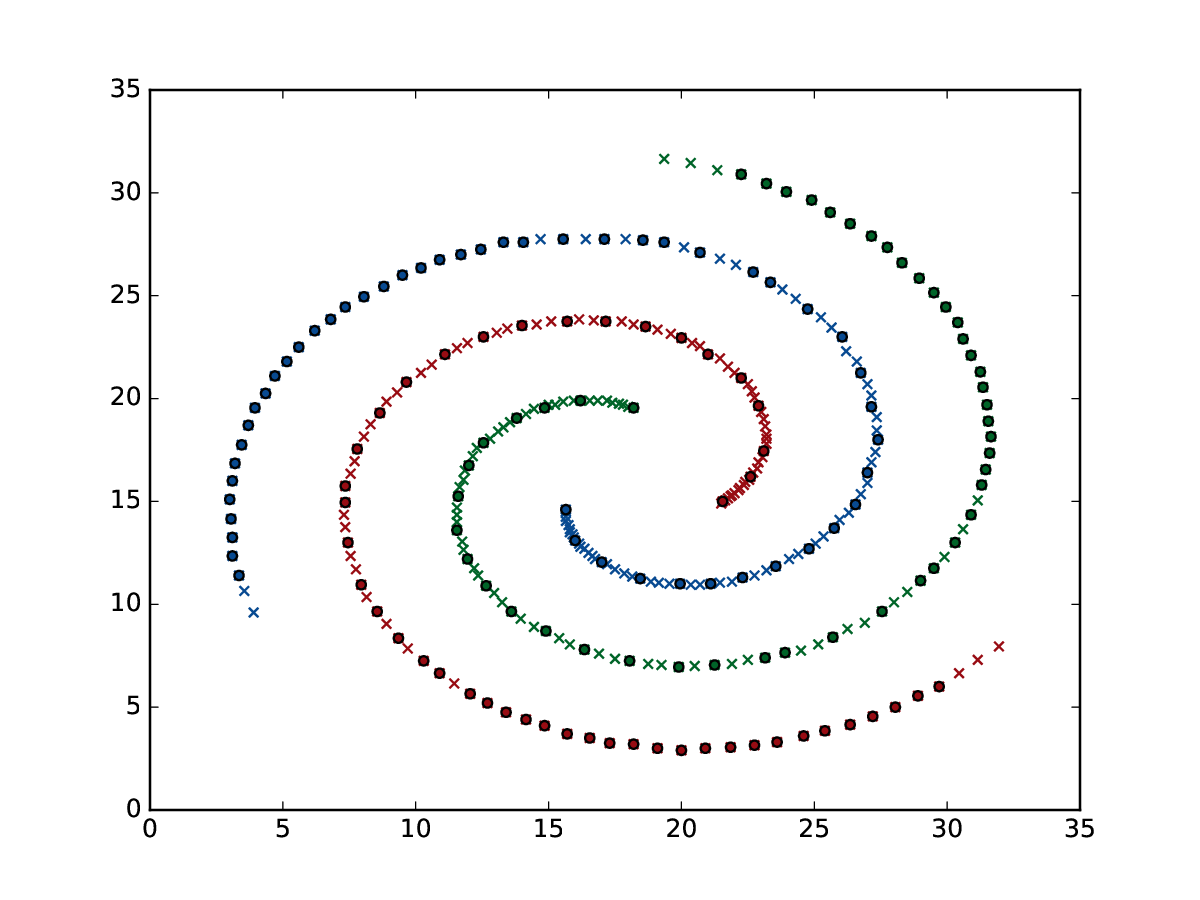}\label{fig:EAPsynspiral1}} \hfill
	\subfigure[IES histogram for $\Delta=0.99$]{\includegraphics[width=0.23\textwidth]{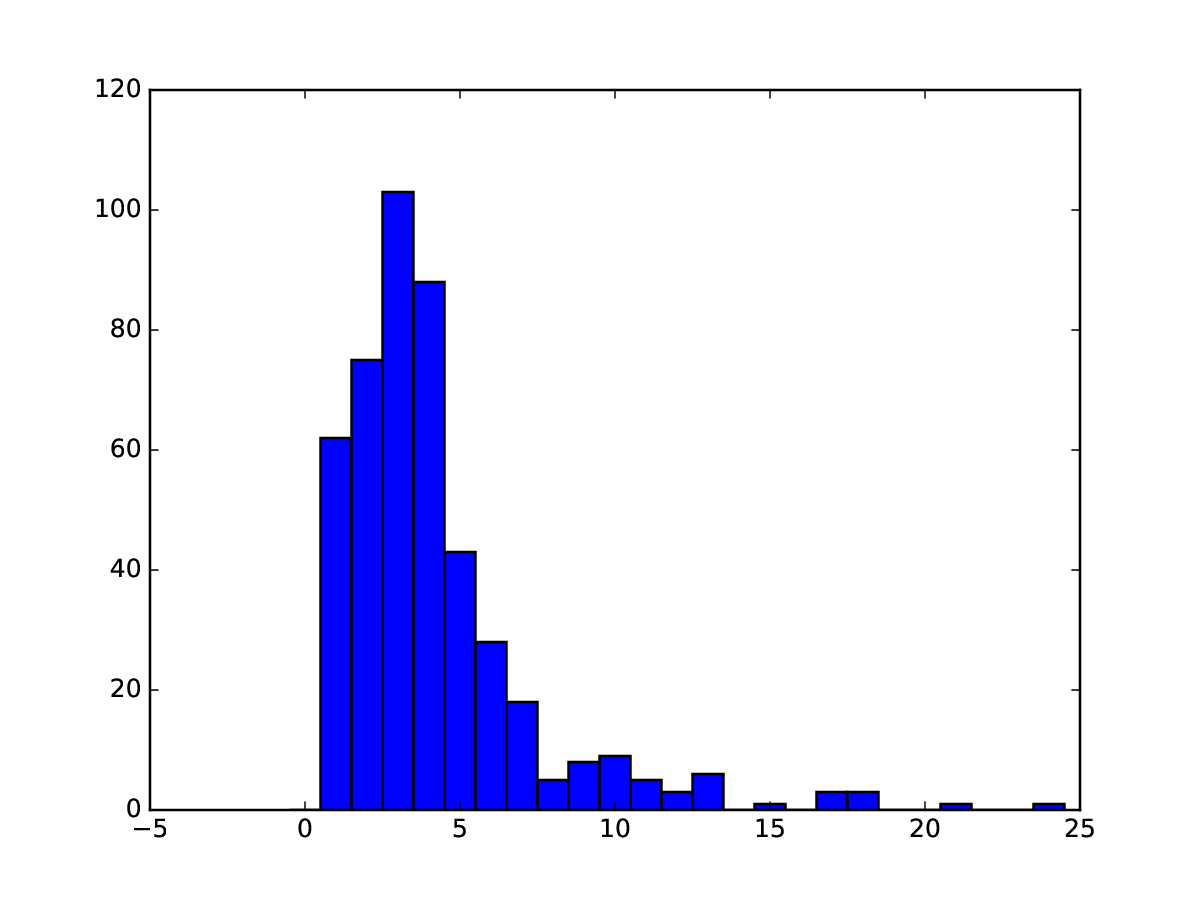}\label{fig:EAPsynspiral4}}
	\caption{EAP results for Spirals dataset where $p=0.6$ and $q=-0.96$}
	\label{fig:EAPsyntheticspiral} 
\end{figure}

The dataset consists of three intertwined spiraling clusters of almost equal size. Fig.~\ref{fig:EAPsyntheticspiral} shows how EAP neatly used the distributed local exemplars and the boundary connections between them to form chains and discovers the global spiral structures. Note that IES histogram in Fig.~\ref{fig:EAPsynspiral4} shows that some of the local exemplar pairs are strongly connected. These are the local exemplar pairs near the inner edge of the spirals where the density is significantly higher than the outer edge. This shows how IES histogram can also be used to indicate the varying densities inside a cluster.
\section{Real World Datasets}\label{sec:APrealworld}
In this section, we discuss the application of EAP to four datasets, Optdigits \cite{uci}, MNIST \cite{mnist}, Skin-segmentation \cite{uci} and protein interactions dataset \cite{complexes_annotated}. We will present both the performance obtained via successive tuning as well as the ceiling performance obtained by sweeping all 3 hyperparameters over a broad enough range. The comparison between the two depicts the performance gap one may experience in practice w.r.t. a plausible ground truth and it also signifies the effectiveness of the successive tuning procedure. We also compare the global performance of EAP with it's parent algorithm AP. We provide a comparison for both the heuristic performance and the ceiling performance of the two algorithms. The heuristic performance of AP is obtained by setting $p=0.5$  \cite{affinitypropagation}. We do not provide comparisons for other algorithms such as MCL and DBSCAN as they do not provide local information, hence they do not serve similar purpose. Interested readers can, however, compare our results for MNIST dataset to the ones discussed in \cite{meap}. Similarly, for protein interactions dataset, interested readers can compare our results to \cite{mcl_vs_ap_proteins} which provides a detailed comparison between AP and MCL. The performance comparison between AP and EAP is presented in Table~\ref{tab:APEAPrealresults} for all four datasets. We have only presented Acc values in the table to present the results in a compact way for comparison. NMI and ARI paint a very similar picture in terms of comparison. We have also included the number of clusters discovered by AP and EAP for the heuristic results. Finally, for Optdigits, MNIST and Skin-segmentation, we also show examples of what local exemplars may represent in real datasets and what does it imply for the data points that are connected to more than one well separated local exemplars.
\begin{table*}[t]
	\centering
	\begin{tabular}{|p{1.8cm}||p{1.2cm}|p{2cm}|p{1.2cm}||p{2.4cm}|p{1.2cm}|p{2cm}|p{1.2cm}|}
		\toprule
		& \multicolumn{3}{c||}{\textbf{AP}} & \multicolumn{4}{c|}{\textbf{EAP}} \\ \hline
		\textbf{Dataset}   & H: Acc & H: Clusters & C: Acc & H: Params          & H: Acc & H: Clusters & C: Acc \\			
		\midrule
		\textbf{Optdigits} & 0.354  & 137         & 0.695  & $p=0.6$, $q =-0.98$ & 0.86   & 26          & 0.871  \\
		\hline
		\textbf{MNIST}     & 0.339  & 113         & 0.456  & $p=0.5$, $q=-0.98$  & 0.562  & 53          & 0.583  \\
		\hline
		\textbf{Skin}      & 0.265  & 212         & 0.304  & $p=0.6$, $q=-0.97$  & 0.856  & 37          & 0.952  \\	
		\hline
		\color{black}
		\textbf{Proteins}  & 0.824  & 405         & 0.875  & $p=0.5$, $q=-0.97$  & 0.864  & 473         & 0.909  \\	\hline
	\end{tabular}
	\caption{Comparison of both heuristic (denoted by H:) and ceiling (denoted by C:) performance, in terms of Acc, of AP and EAP on real datasets. For all four datasets successive tuning results into $\Delta=0.99$. For Proteins dataset the hyperparameter values are mentioned in terms of the percentile of non-zero values in the pairwise similarity matrix since $\Svec$ is sparse. In case of sparse matrices, the parameter tuning should be done in terms of the non-zero pairwise similarities.}
	\label{tab:APEAPrealresults}
\end{table*}
\subsection{Optdigits}\label{subsec:EAPoptdigits}
Optdigits consists of grey scale $8\times 8$ images of handwritten digits \cite{uci}. We only use the test set, consisting of $1797$ images, for clustering. We use negative of Euclidean distance  between data points as pairwise similarities. We do recognize that for optical character recognition applications one can use a better adapted pairwise similarity metric, such as one based on SIFT features \cite{sift}, but we want to show what EAP is able to extract using this crude measure and not rely on the power of a strong metric, which may require expert knowledge, that already simplifies the clustering problem significantly for the algorithm. Besides, one can use the training set to extract the right scaling or other factors for the pairwise similarity metric but as our focus is not on how to design pairwise similarity metrics, we do not do employ any such techniques. For the ground truth we separate the dataset into $10$ clusters each corresponding to a different digit. Table \ref{tab:APEAPrealresults} shows the global performance of AP and EAP as well the corresponding hyperparameters and number of clusters discovered.  We can see that EAP outperforms AP in both ceiling as well as heuristic performance. Furthermore, EAP is able to achieve good accuracy and the gap between heuristic and ceiling performance is not big. Fig.~\ref{fig:EAPrealopt} shows an example of the discussion in Sec.~\ref{subsec:APdensehist}. Each image represents a data point and the local exemplars are marked with the red boundaries. We can see that the different local exemplars correspond to different handwriting styles and that the data points connected to two local exemplars exhibit a mix of the two handwriting styles. 
\subsection{MNIST}\label{subsec:EAPMNIST}
\begin{figure}[t]
	\centering 
	\subfigure[Optdigits]{\includegraphics[width=0.2\textwidth, height=0.3\textwidth]{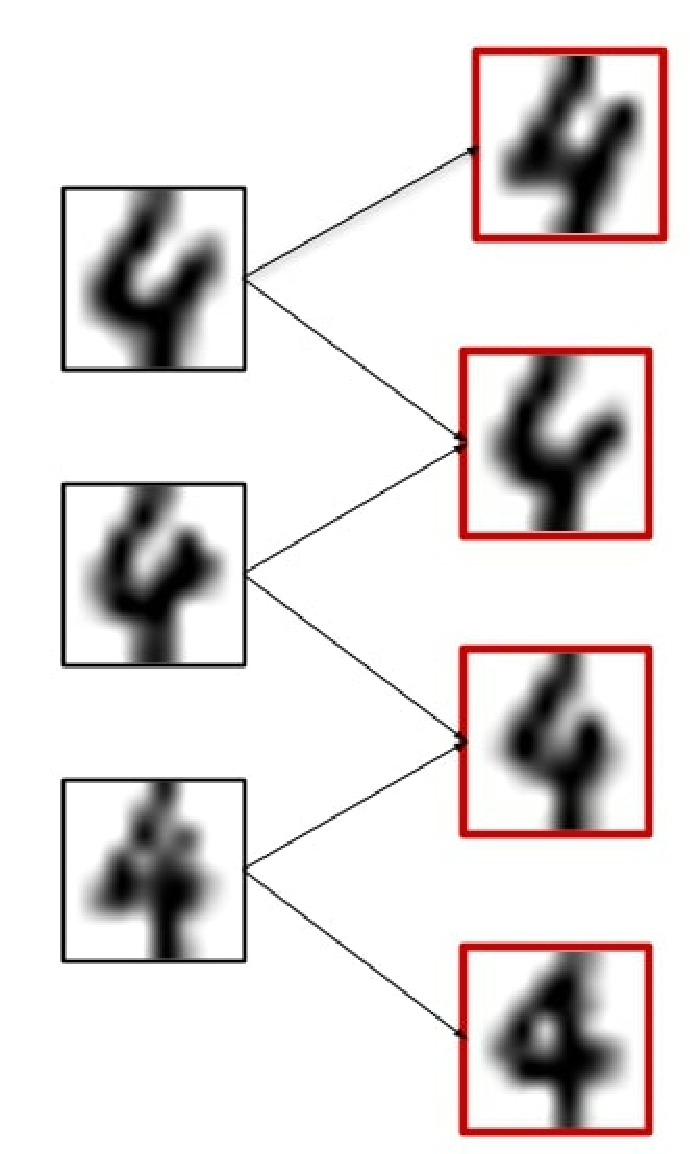}\label{fig:EAPrealopt}}\hfill
	\subfigure[MNIST]{\includegraphics[width=0.2\textwidth, height=0.3\textwidth]{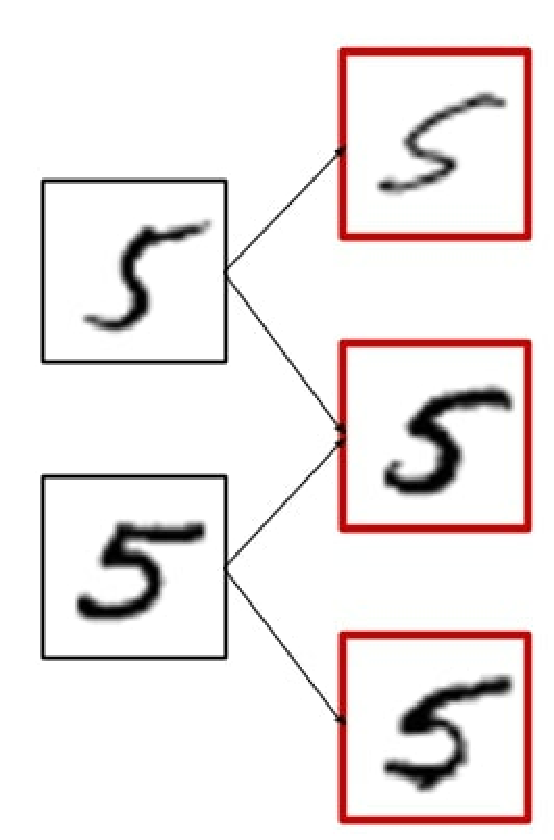}\label{fig:EAPrealmnist}}
	\caption{Examples of some local exemplars, corresponding to different hand writing styles for the same digit. The figure also shows some data points connected to two local exemplars, exhibiting a mix of the two hand writing styles.}
	\label{fig:APEAPreallocal}	
\end{figure}

MNIST is also a similar dataset of $28\times28$  gray scale images of handwritten digits. We choose $1000$ images from the dataset at random to define $\mc{X}$. We use negative of Euclidean distance  between data points as pairwise similarities  and the same discussion is valid for MNIST regarding the computation for better pairwise similarity metrics as for Optdigits. We consider all the examples of a specific digit as belonging to one ground truth cluster. Table \ref{tab:APEAPrealresults} shows the performance. We can again see that EAP provides a significantly better heuristic performance as compared to AP. The absolute performance itself is not very good but when one compares it to the performance of other algorithms reported in \cite{meap}, it is at par or better than the algorithms reported  in \cite{meap}. We believe that all these algorithms suffer in terms of absolute performance due to the crude pairwise similarity metric used. As the images get bigger in size, there is a higher chance of pixels corresponding to the digits not being aligned in different images and this leads to a lower pairwise similarity between different images of same digit as well. This again signifies the importance of finding a problem dependent suitable method to calculate pairwise similarities. We again notice that the gap between the heuristic and the ceiling performance of EAP is not significant.  Fig.~\ref{fig:EAPrealmnist} shows another example of the discussion in Sec.~\ref{subsec:APdensehist} where different local exemplars in the same cluster correspond to different handwriting styles and the data points connected to two local exemplars show a mix of the properties of both local exemplars.

\subsection{Skin Segmentation Dataset}
The Skin Segmentation dataset is also taken from UCI repository \cite{uci}. It is constructed over (B, G, R) color space, hence each data point is a three dimensional real vector. It has two major clusters, one for non-skin tones and the other for skin tones. The skin tones are obtained using skin textures from face images of people of different age, gender and race. Since different data points refer to different colors, this dataset is suitable for demonstration purposes. The negative of euclidean distance serves as a good similarity metric here since the data points are only $3$ dimensional vectors, with each dimension having a distinct meaning. The dataset consists of 245057 points. We randomly choose a subset of 5000 points that are equally distributed between the two classes.

\begin{figure}[t]
	\centering 
	\includegraphics[width=0.4\textwidth, height=0.3\textwidth]{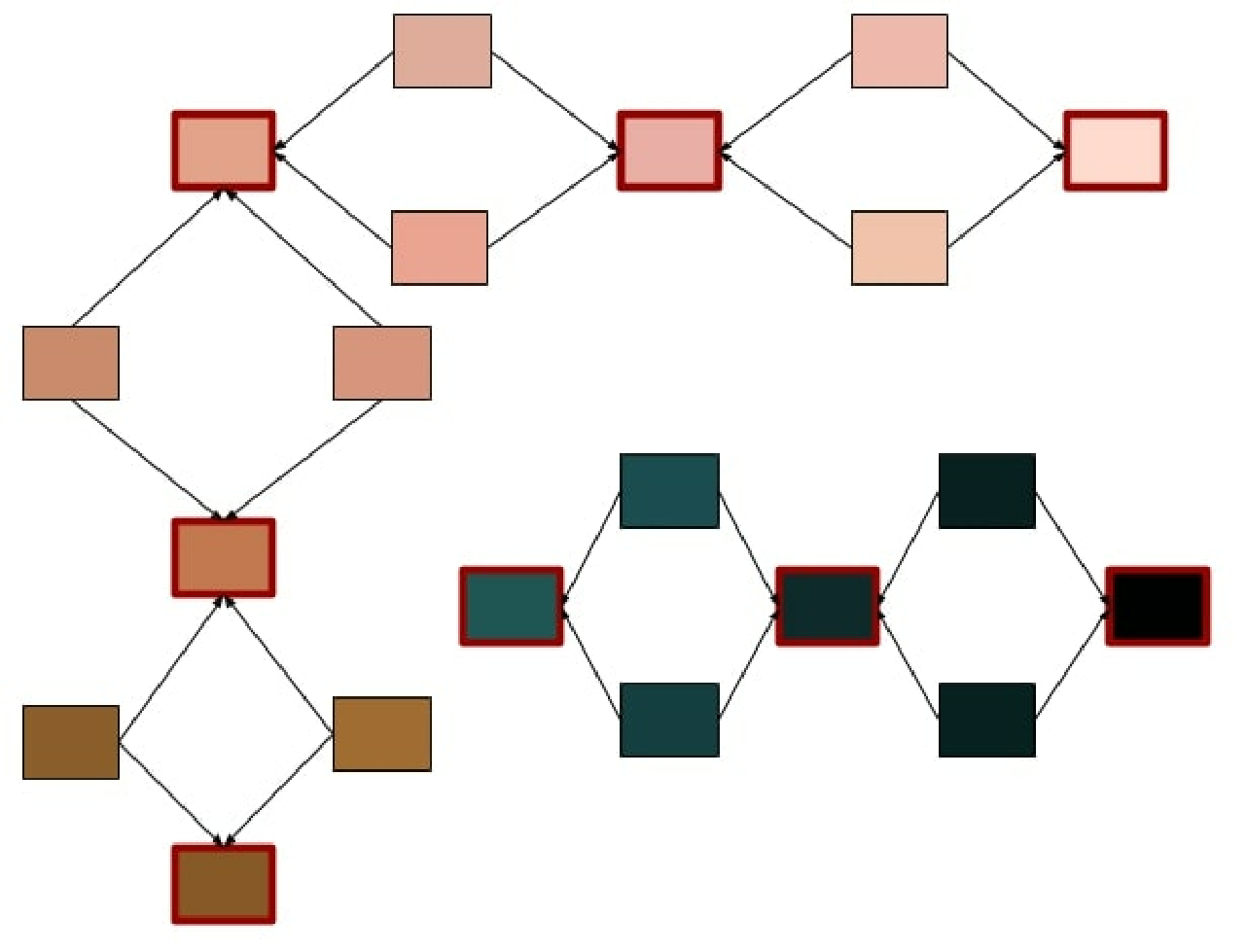}
	\caption{An illustration of some local exemplars and some data points connected to these local exemplars,  showing the chain structure formed by EAP using the boundary connections to discover the global structure.}
	\label{fig:EAPrealskin}
\end{figure}

The results in Table~\ref{tab:APEAPrealresults} show a stark difference between the results of AP and EAP. The cause of this can be understood by analyzing the dataset. We can see from Fig.~\ref{fig:EAPrealskin} that both skin and non-skin classes consist of a range of colors, hence the two clusters do not correspond to spherical structures. AP fails to find this structure and results in a larger number of clusters. Even the ceiling performance of AP is significantly worse than the heuristic performance of EAP. We can see the gradual shift from lighter to darker skin tones in  Fig.~\ref{fig:EAPrealskin} which is discovered by EAP. Apart from the two major clusters, most of the clusters identified by EAP consist of outliers. This is why we have a high accuracy despite the number of discovered clusters being $37$.
	
\subsection{Proteins Interactions Dataset}
Finally, we apply EAP to the protein interactions dataset \cite{complexes_annotated}.  The pairwise similarity values for this dataset represent protein-protein interactions. \cite{protein_interactions} proposes a probabilistic measure, which takes value between $0$ and $1$, to specify the pairwise interactions. We use this to define our pairwise similarity matrix. This dataset is different from the previously discussed synthetic and real datasets in the sense that the available ground truth does not necessarily assign each data point to only one ground truth cluster. Furthermore, some ground truth clusters are a subset of other ground truth clusters. In order to reduce such overlap in the discovered clusters, we threshold the protein-protein interactions such that the interactions below $0.4$ are set to $0$\footnote{This approach has been adopted in previous studies too e.g. \cite{mcl_vs_ap_proteins} and \cite{conf_0.38}, although different thresholds have been used there.}. Furthermore, we only consider proteins that are common with CYC2008. This leaves us with $1171$ proteins (data points) assigned to $276$ ground truth clusters. Table~\ref{tab:APEAPrealresults} shows the global performance. EAP outperforms AP but only by a small margin, both in terms of heuristic as well as ceiling performance, since AP already performs well on this dataset. Besides, the heuristic  and ceiling performance of EAP are again close. For this dataset we cannot provide any insights into the meaning of the obtained local information since we don't have the domain knowledge to interpret the results and since most of the discovered clusters consist of only a few proteins. 
%%%%%%%%%%%%%%%%%%%%%%%%%%%%%

\ifCLASSOPTIONcompsoc
\section*{Acknowledgments}
\else
\section*{Acknowledgment}
\fi
The work of Rana Ali Amjad has has been funded by the German Ministry of Education and Research in the framework of an Alexander von Humboldt Professorship.

\ifCLASSOPTIONcaptionsoff
\newpage
\fi

\bibliography{references}
\bibliographystyle{IEEEtran}

\begin{IEEEbiography}[{\includegraphics[width=1in,height=1.25in,clip,keepaspectratio]{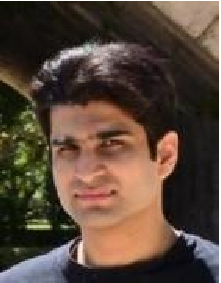}}]{Rana Ali Amjad}
	(S'13) was born in Sahiwal, Pakistan, in 1989. He received the Bachelors degree in Electrical Engineering  (with highest distinction) from University of Engineering and Technology, Lahore, Pakistan, in 2011. He completed his Masters degree in Communication Engineering (with highest distinction) from Technical University of Munich, Germany, in 2013.

	Since 2014 he is pursuing his PhD at the Institute for Communication Engineering at Technical University of Munich. He has received various awards in his academic career including the faculty award for best Master thesis, award for outstanding performance in Master's degree and Gold medal for best performance in Communications major during his Bachelors degree. His research interests cover information theory, machine learning, communication theory, channel coding and information-theoretic security.
\end{IEEEbiography}

\begin{IEEEbiography}[{\includegraphics[width=1in,height=1.25in,clip,keepaspectratio]{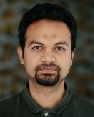}}]{Rayyan Ahmad Khan} was born in Pakistan. He received his Masters degree in Communication Engineering from Technical University of Munich, Germany, in 2018. After working for one year as a JAVA developer, he joined the research group at Mercateo, Munich in 2019 as a doctoral researcher. His interests lie primarily in the field of machine learning, deep learning, data mining, statistical signal processing and information theory.
\end{IEEEbiography}

\begin{IEEEbiography}[{\includegraphics[width=1in,height=1.25in,clip,keepaspectratio]{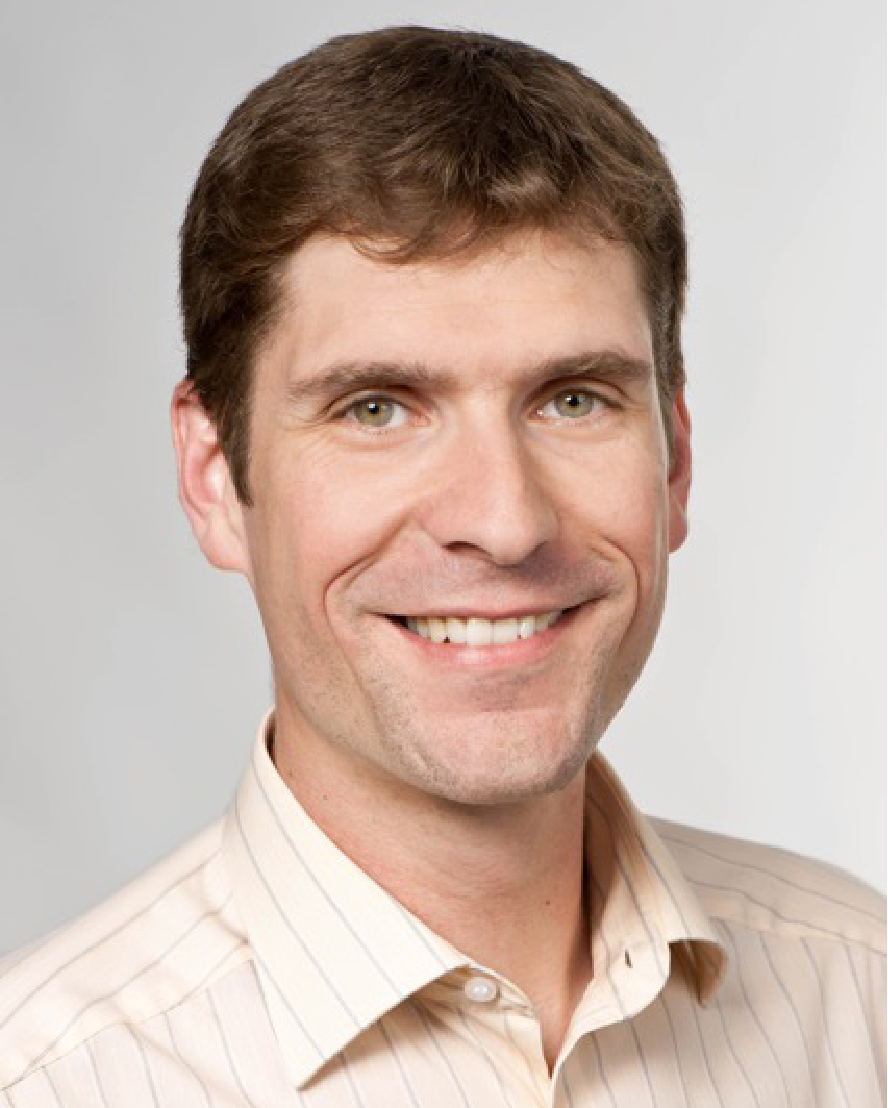}}]{Martin Kleinsteuber}
	received his Ph.D. in Mathematics from the University of Würzburg, Germany, in 2006. After post-doc positions at National ICT Australia Ltd., the Australian National University, Canberra, Australia, and the University of Würzburg, he has been appointed assistant professor for geometric optimization and machine learning at the Department of Electrical and Computer Engineering, TU München, Germany, in 2009. He won the SIAM student paper prize in 2004 and the Robert-Sauer-Award of the Bavarian Academy of Science in 2008 for his works on Jacobi-type methods on Lie algebras. Since 2016, he is Chief Information Officer and Lead Data Scientist for the Mercateo Group, Munich.
\end{IEEEbiography}

\clearpage

\appendices
\section{Derivation of \eqref{eq:APEAPmsgi}} \label{app:EAPeta}
Using max-sum algorithm, we have
\begin{equation}\label{eq:etaderiv1}
	\nu_{\overline{g}_i \rightarrow b_{ij}}(b_{ij}) = \max \limits_{\sim b_{ij}} \left( \overline{g}_i(\Bvec(i,:)) + \sum \limits_{l \neq j}  \mu_{b_{il} \rightarrow g_i}(b_{il}) \right)
\end{equation}
We will evaluate \eqref{eq:etaderiv1} for $b_{ij}=0$ and $b_{ij}=1$. Then we will obtain $\eta_{ij}$ using \eqref{eq:etadiffdef}. For $b_{ij}=1$, we satisfy the requirement $\sum\limits_{j} b_{ij} \geq 1$ regardless of the values of other variables involved. Hence $\overline{g}_i(\Bvec(i,:)) = \sum \limits_{k} b_{ik}q$
\begin{align}
	\nu_{\overline{g}_i \rightarrow b_{ij}}(1) & = \max \limits_{\sim b_{ij}} \left[ \sum \limits_{m} b_{im}q + \sum \limits_{l \neq j} \mu_{b_{il} \rightarrow \overline{g}_i}(b_{il}) \right]                               \\
	                                           & \overset{(a)}{=} q + \max \limits_{\sim b_{ij}} \left[ \sum \limits_{l \neq j} \Big( b_{il}q +\mu_{b_{il} \rightarrow \overline{g}_i}(b_{il}) \Big) \right]                  \\
	                                           & \overset{(b)}{=} q + \sum\limits_{l \neq j} \max \limits_{b_{il} \in \{0,1\}}  \Big( b_{il}q +\mu_{b_{il} \rightarrow \overline{g}_i}(b_{il}) \Big)                          \\
	                                           & = q + \sum \limits_{l \neq j} \max \left\{ q + \mu_{b_{il} \rightarrow \overline{g}_i}(1),\mu_{b_{il} \rightarrow \overline{g}_i}(0) \right\}                                \\
	                                           & \overset{(c)}{=} q+  \sum \limits_{k \neq j}  \mu_{b_{ik} \rightarrow \overline{g}_i}(0) + \sum\limits_{l \neq j} \max \left\{q + \beta_{il}, 0\right\} \label{eq:etaderiv2} 
\end{align}
where (a) follows by taking the penalty for $b_{ij}=1$ outside the maximum. (b) follows from the observation that each component of the sum depends on only one variable and can be optimized independently. (c) follows from the definition of $\beta_{il}$ in \eqref{eq:betadiffdef}. 

For $b_{ij}=0$, we need to have at least one $b_{il}=1$ in order to avoid $ \overline{g}_i(\Bvec(i,:)) = -\infty$. The rest of the variables involved can be chosen freely. 
\begin{align}
	\nu_{\overline{g}_i \rightarrow b_{ij}}(0) & =\max \limits_{m \neq j}\left[q + \mu_{b_{im} \rightarrow \overline{g}_i}(1) + \right. \nonumber                                                                               \\ &\left. \max \limits_{\{b_{il}\}\backslash\{b_{ij},b_{im}\}} \left(\sum \limits_{l \neq \{m, j\}} \left( b_{il}q +   \mu_{b_{il} \rightarrow \overline{g}_i}(b_{il}) \right) \right)\right] \nonumber \\
	                                           & = \max \limits_{m \neq j} \left[ q + \mu_{b_{im} \rightarrow \overline{g}_i}(1)  + \sum \limits_{k \neq \{j,m\}}  \mu_{b_{ik} \rightarrow \overline{g}_i}(0) \right. \nonumber \\ &\left. + \sum\limits_{l \neq \{m,j\}} \max \left\{q + \beta_{il}, 0\right\}  \right] \label{eq:etaderiv3}
\end{align}
where (d) follows by using same techniques as in (b) and (c). Combining \eqref{eq:etaderiv2} and \eqref{eq:etaderiv3} we get
\begin{align}
	\eta_{ij} & =  - \max \limits_{m \neq j} \left[ \mu_{b_{im} \rightarrow \overline{g}_i}(1)  -  \mu_{b_{im} \rightarrow \overline{g}_i}(0) + \max \left\{q + \beta_{im}, 0\right\} \right] \\
	          & = - \max \limits_{m \neq j} \left[ \beta_{im} + \min \left\{-q - \beta_{im}, 0\right\} \right]                                                                                \\
	          & = - \max \limits_{m \neq j} \left[ \min \left\{\beta_{im}, -q\right\} \right]                                                                                                 \\
	          & = \max \left[ -\max \limits_{m \neq j} \beta_{im} \ ,\  q \right]\label{eq:eta_eap}                                                                                           
\end{align}
\section{Derivation of \eqref{eq:eap_psi_main}} \label{app:EAPpsi}
For $i \in \mc{N}_j$ and $b_{ii}=1$, we need all the other involved variables to be $0$ in order to avoid $r_j(\mc{N}_j) = -\infty$. Hence 
\begin{equation} \label{eq:psideriv1}
	\nu_{r_j \rightarrow b_{ii}}(1) = \sum\limits_{\substack{k\in \mc{N}_j \\ k\neq i}}  \mu_{b_{kk} \rightarrow r_j}(0)
\end{equation}
For $b_{ii}=0$, in order to avoid $r_j(\mc{N}_j) = -\infty$ we can either choose all other involved variables to be also $0$ or we can allow one of them to be equal to $1$. Hence
\begin{align}
	\nu_{r_j \rightarrow b_{ii}}(0) & = \max \left\{\sum\limits_{\substack{k\in \mc{N}_j \\ k\neq i}}  \mu_{b_{kk} \rightarrow r_j}(0), \right. \nonumber \\ &\left. \max\limits_{\substack{l\in \mc{N}_j \\ l\neq i}} \left[  \mu_{b_{ll} \rightarrow r_j}(1)  + \sum\limits_{\substack{k\in \mc{N}_j \\ k\neq \{i,l\}}}  \mu_{b_{kk} \rightarrow r_j}(0)\right]\right\} \\
	&= \max \left\{0, \max\limits_{\substack{l\in \mc{N}_j \\ l \neq i}}  \phi_{lj} \right\} +  \sum\limits_{\substack{k\in \mc{N}_j \\ k\neq i}}  \mu_{b_{kk} \rightarrow r_j}(0) \label{eq:psideriv2}
\end{align}
Combining \eqref{eq:psideriv1} and \eqref{eq:psideriv2} we obtain
\begin{equation}
	\psi_{ij} = - \max\left\{0, \max \limits_{\substack{l \in \mc{N}_j \\ l \neq i}} \phi_{lj}\right\}
\end{equation}

\section{Additional Features of EAP}\label{app:addfeatures}
Besides the features that we already mentioned in Sec.~\ref{sec:APeapprop}, we will elaborate on two additional features of EAP in this appendix. 
\subsection{Clustering New Data Points}\label{app:newpoints}
We mentioned in Sec.~\ref{sec:APeapprop} that the local exemplars can be used to efficiently adapt the clustering results for evolving datasets. Now we will elaborate more on this. For example, they can be used to efficiently cluster a new data point $x_{new}$ after you have already clustered the original dataset $\mc{X}$ using EAP. We can compare  $x_{new}$ to the already found local exemplars ($\mc{E}$) and assign it to the cluster which contains the closest local exemplar. This usually lowers the complexity by orders of magnitude when compared to assigning a cluster by finding the closest neighbour in the already clustered dataset, as would be the case for algorithms that do not provide any local information, and provides further information as follows: If there are two or more local exemplars belonging to the same cluster which are almost equally close to the data point $x_{new}$ we can connect $x_{new}$ to all of them. If there are two exemplars belonging to different clusters that are almost equally close to $x_{new}$, then this new data point can be passed on to a human analyst as a potential inconsistency that needs to resolved by expert opinion. If the new data point has low enough similarity to all of the local exemplars, it can be treated as an outlier. We can apply this procedure to more than one new data points as well. In that case we should just be more careful that if we notice a sufficiently large number of new data points being declared as outliers this may be an indication of a new cluster being formed which was not present in the clustering results for $\mc{X}$ using EAP. Also if we notice enough data points having close enough similarities to two exemplars from different clusters, they may form a strong enough bridge between the two clusters to merge them to one. In these cases it is a good idea to either recluster the whole dataset or a partial subset of the dataset. 
\subsection{Outliers and Noisy Data}
\begin{figure}[t]
	\centering
	\includegraphics[width=0.5\textwidth]{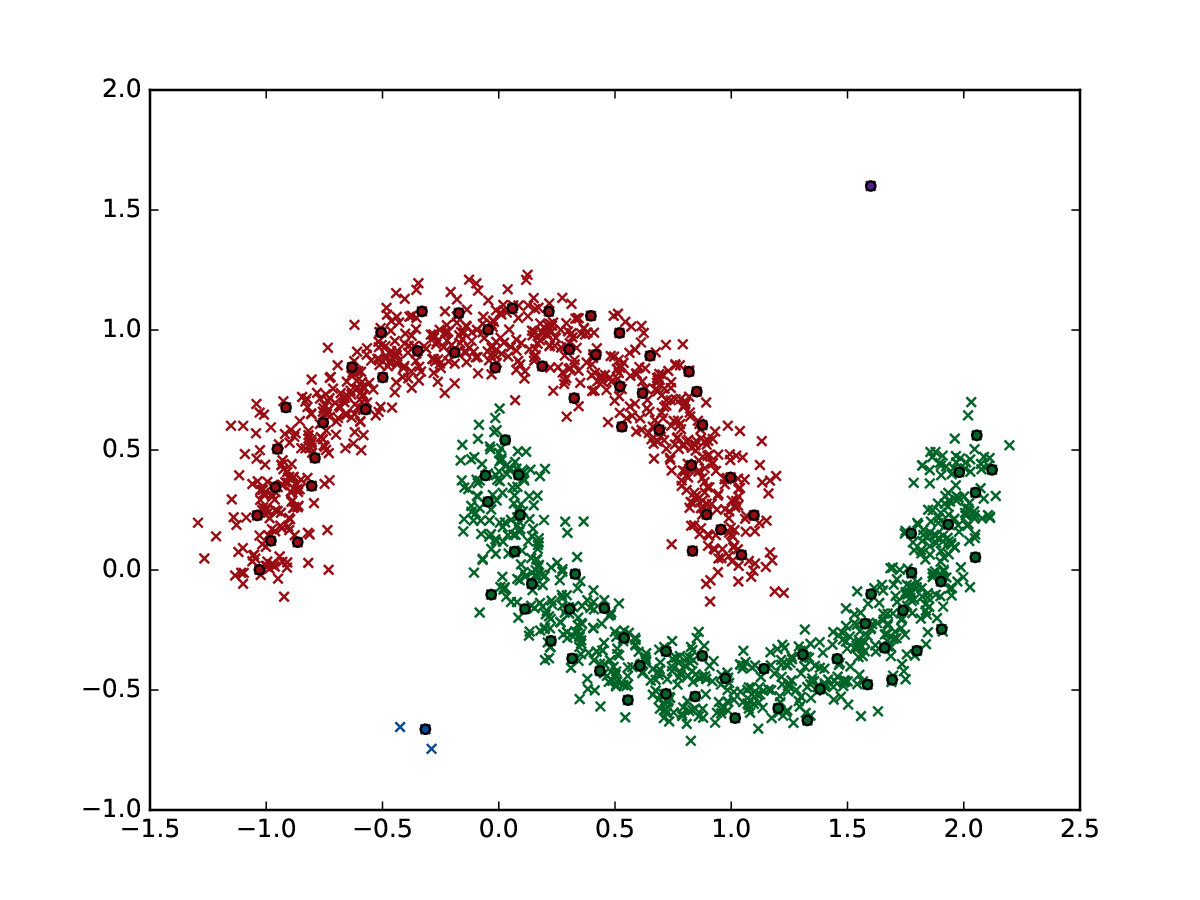}
	\caption{Half-moons  dataset with outliers. These results are obtained using the same parameters as Fig.~\ref{fig:APEAPwithneigh}, i.e., $p=0.6$, $q=-0.97$ and $\Delta=0.99$.}
	\label{fig:APEAPoutlier}
\end{figure}
We briefly mentioned for in Sec.~\ref{sec:APsynthetic} how EAP handles noisy boundaries between clusters for different synthetic datasets. We will now explore how EAP handles outliers. It starts by considering all data points as potential local exemplars and then explores the neighbourhood of each data point to later decide if a data point should become a local exemplar or not. This provides EAP the ability to discover outliers as they do not tend to connect to other local exemplars that belong to well formed bigger far away clusters. This is depicted in Fig.~\ref{fig:APEAPoutlier}, where we have the half-moons dataset with $2$ clusters and we have now added some outliers to it. Specifically, we have a lone outlier on the top right side of the figure and a small cluster of three outliers on the bottom left of the figure. EAP recognizes the lone outlier as a single point cluster, represented by purple colour. Similarly, EAP  also puts together the three outliers on the lower left corner as one small cluster, designated by blue colour. It is also important to note that the presence of these outliers does not impact the clustering of the two well formed bigger clusters. This can be observed by comparing Fig.~\ref{fig:APEAPoutlier} to Fig.~\ref{fig:APEAPwithneigh}, we we have the same dataset but without the outliers. 
\section{Successive Tuning Example}\label{sec:tuningexp}
\begin{figure}[t]
	\centering 
	\subfigure[LEC histogram]{\includegraphics[width=0.24\textwidth]{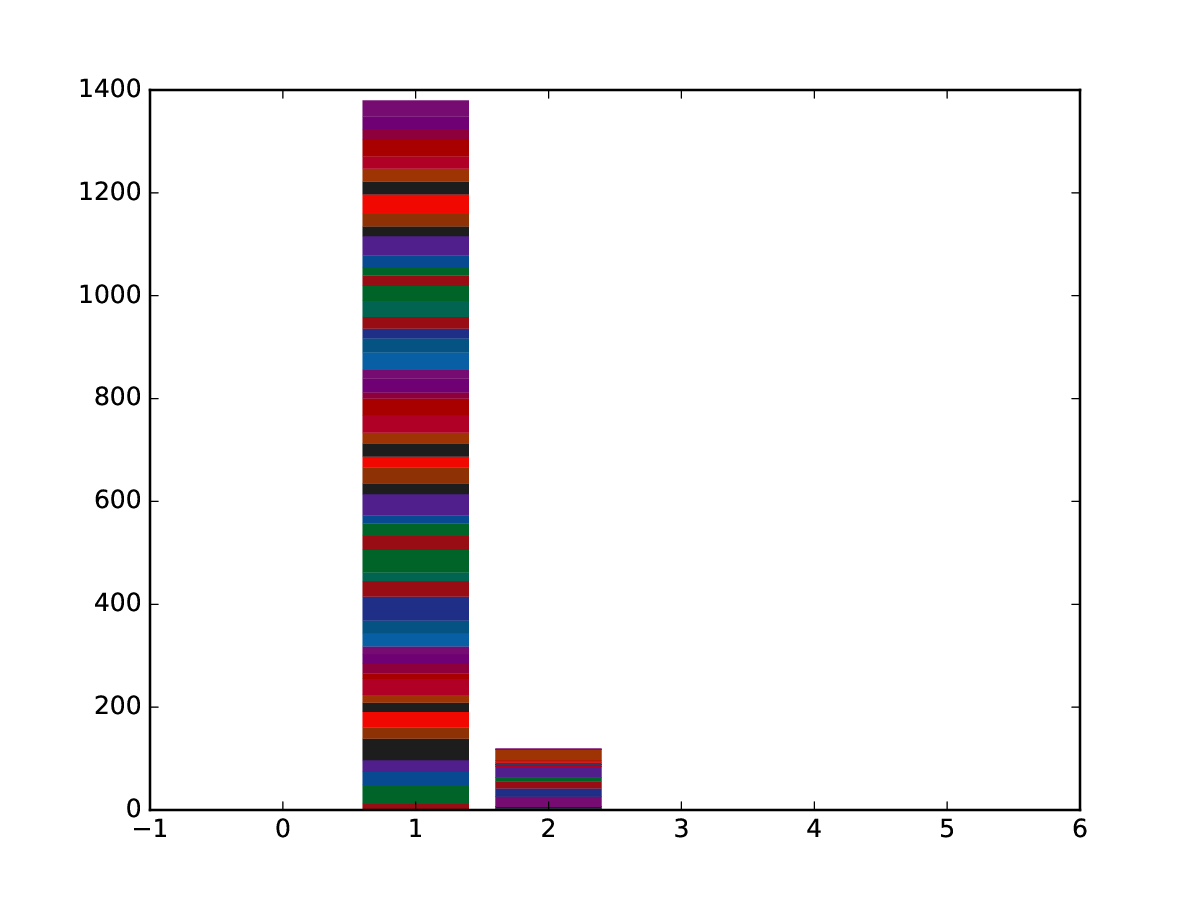}\label{fig:EAPsuccessivetuningstep1LEC}} \hfill
	\subfigure[IES histogram]{\includegraphics[width=0.24\textwidth]{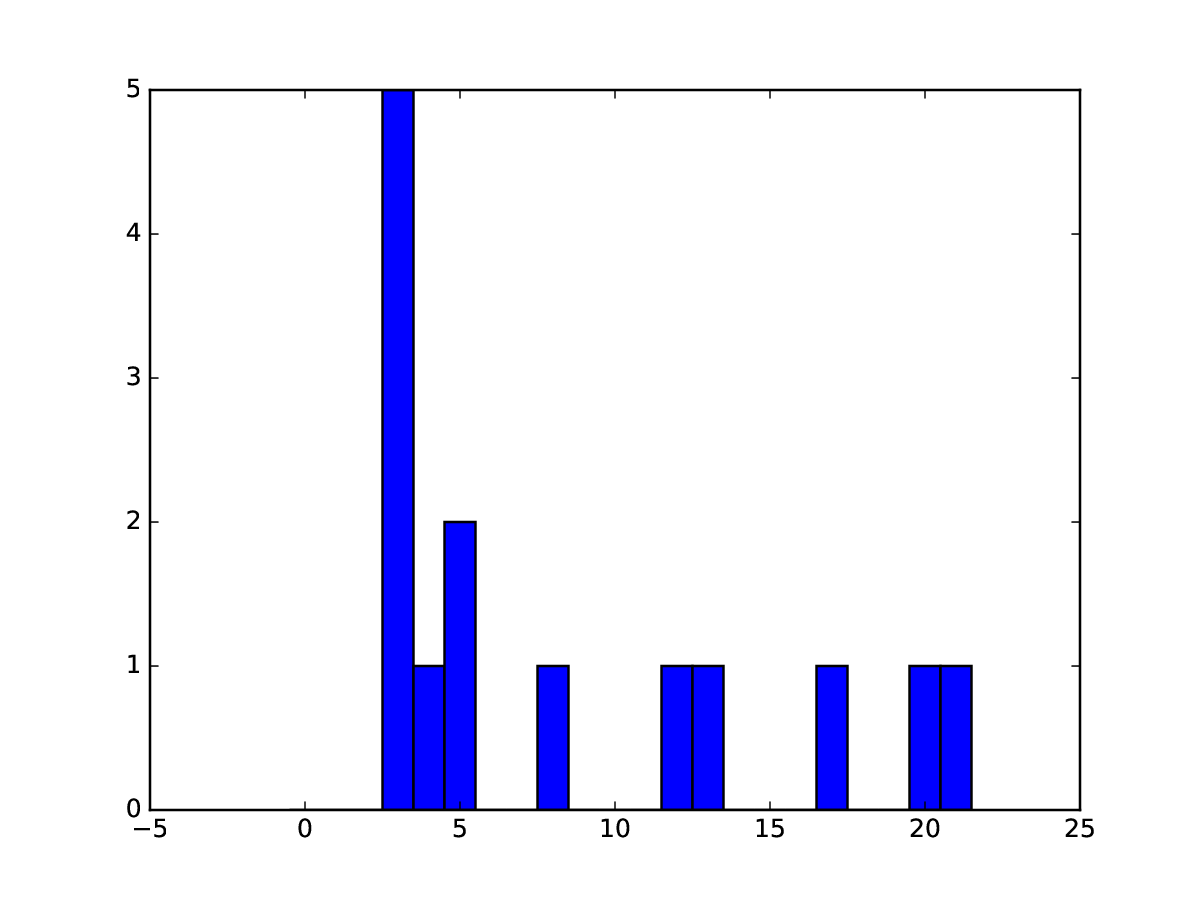}\label{fig:EAPsuccessivetuningstep1IES}}\hfill
	\subfigure[Clusters with Exemplars]{\includegraphics[width=0.4\textwidth]{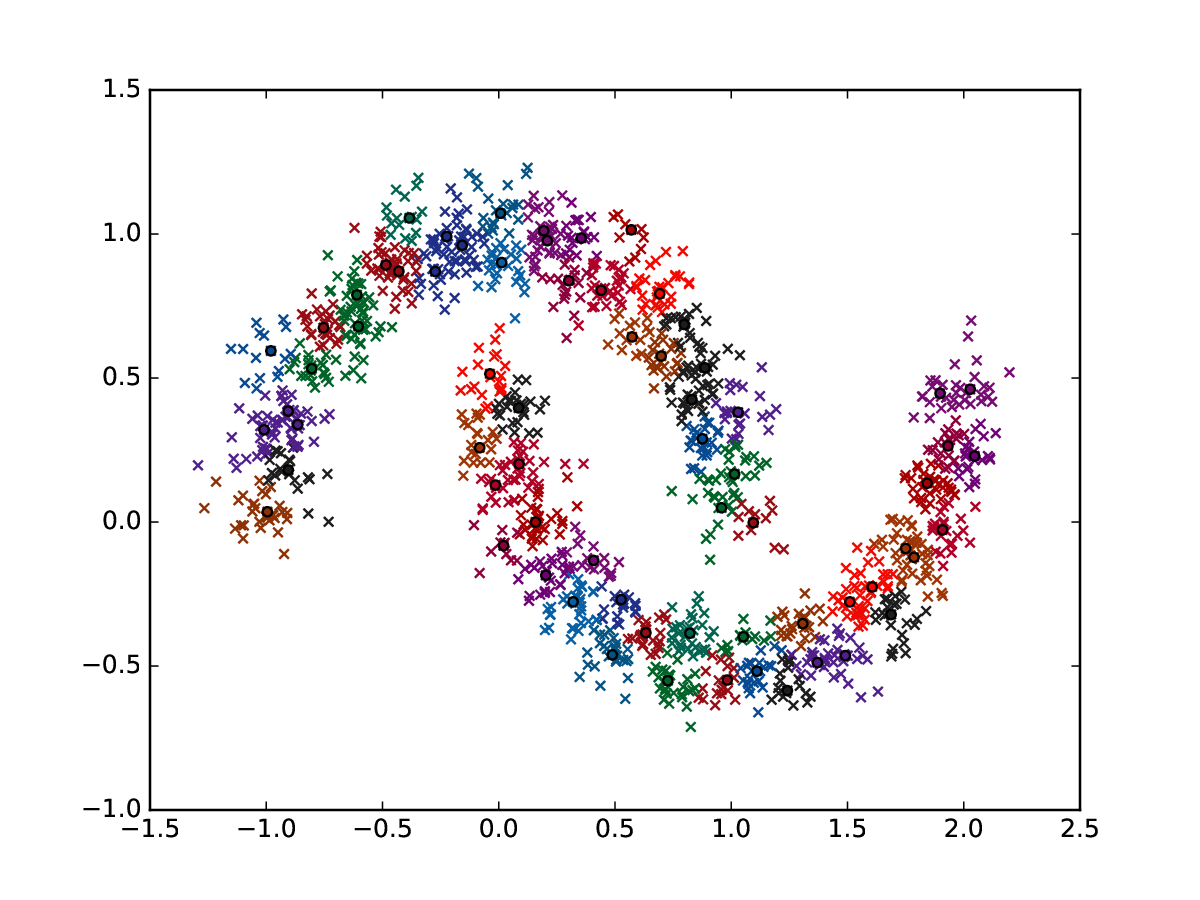}\label{fig:EAPsuccessivetuningstep1clusters}}\hfill
	\caption{The figure shows the results obtained in the first step of successive tuning corresponding to $p=0.6$, $q=-0.99$, $\Delta >1$.}
	\label{fig:EAPsuccessivetuningstep1} 
\end{figure}
We will now present a step by step example of using the insights from Sec.~\ref{sec:APtuning} to do hyperparameter tuning for the half-moons dataset earlier used in Fig.~\ref{fig:APEAPwithneigh}. As the first step, we apply EAP to the dataset for $p=0.6$, $q=-0.99$, $\Delta >1$. The results obtained contain $47$ clusters. None of these clusters correspond to outliers (i.e., none of them are clusters consisting of only a couple of points far away from the rest of the dataset), hence all of them represent regular clusters. The number of detected regular clusters are significantly higher than expected. This is also evident by looking at the LEC histogram in Fig.~\ref{fig:EAPsuccessivetuningstep1LEC}, where almost all the data points are connected to only one local exemplar. It is also clearly visible in the IES histogram in Fig.~\ref{fig:EAPsuccessivetuningstep1IES}, where there are very few local exemplars which share any boundary connections. These results are similar to what one can expect from AP (for the same $p$, AP will lead to even more clusters as it allows no boundary connections). The resulting clusters along with the local exemplars, many of which form separate clusters around them due to the lack of boundary connections, are shown in Fig.~\ref{fig:EAPsuccessivetuningstep1clusters} (some colours are reused to represent different clusters due to a limited number of clearly distinguishable colours in a small figure but in most cases it is clearly from the spatial locations of different clusters with same colour that they do not belong together). Note that Fig.~\ref{fig:EAPsuccessivetuningstep1clusters} is shown here just for illustration purposes and is not needed to decide the next step in successive tuning. One can easily infer from the number of obtained clusters, the LEC histogram and the IES histogram that we need to reduce the linkage penalty.
\begin{figure}[t]
	\centering 
	\subfigure[LEC histogram]{\includegraphics[width=0.24\textwidth]{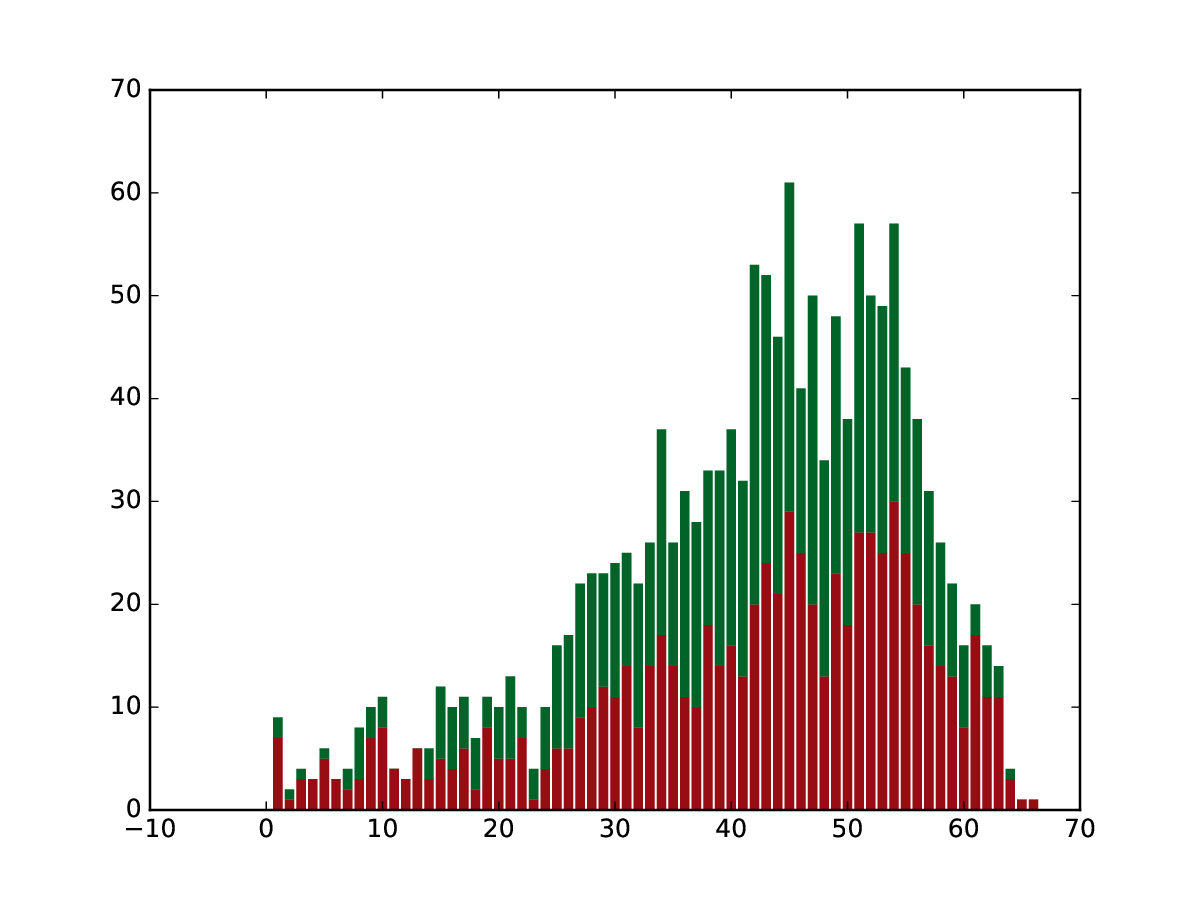}\label{fig:EAPsuccessivetuningstep2LEC}} \hfill
	\subfigure[IES histogram]{\includegraphics[width=0.24\textwidth]{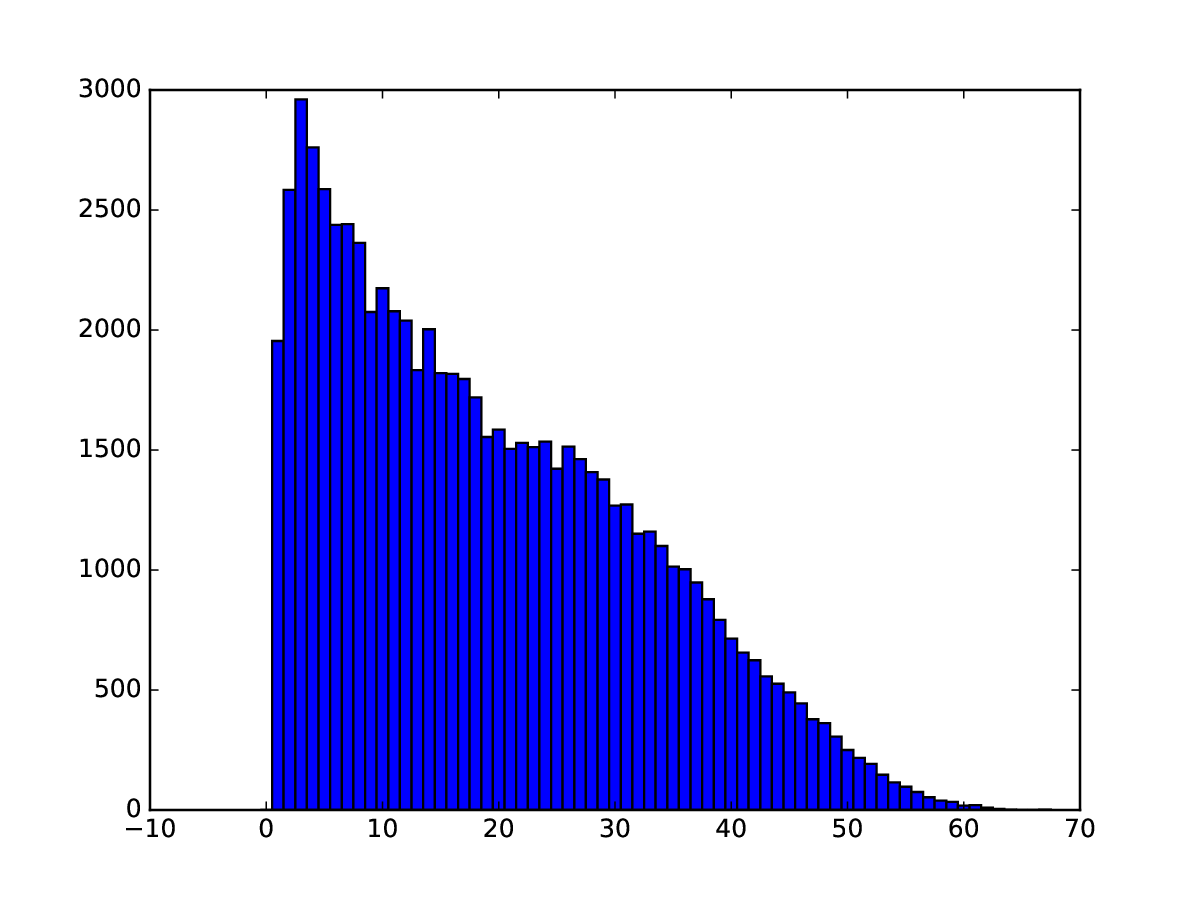}\label{fig:EAPsuccessivetuningstep2IES}}\hfill
	\subfigure[Clusters with Exemplars]{\includegraphics[width=0.4\textwidth]{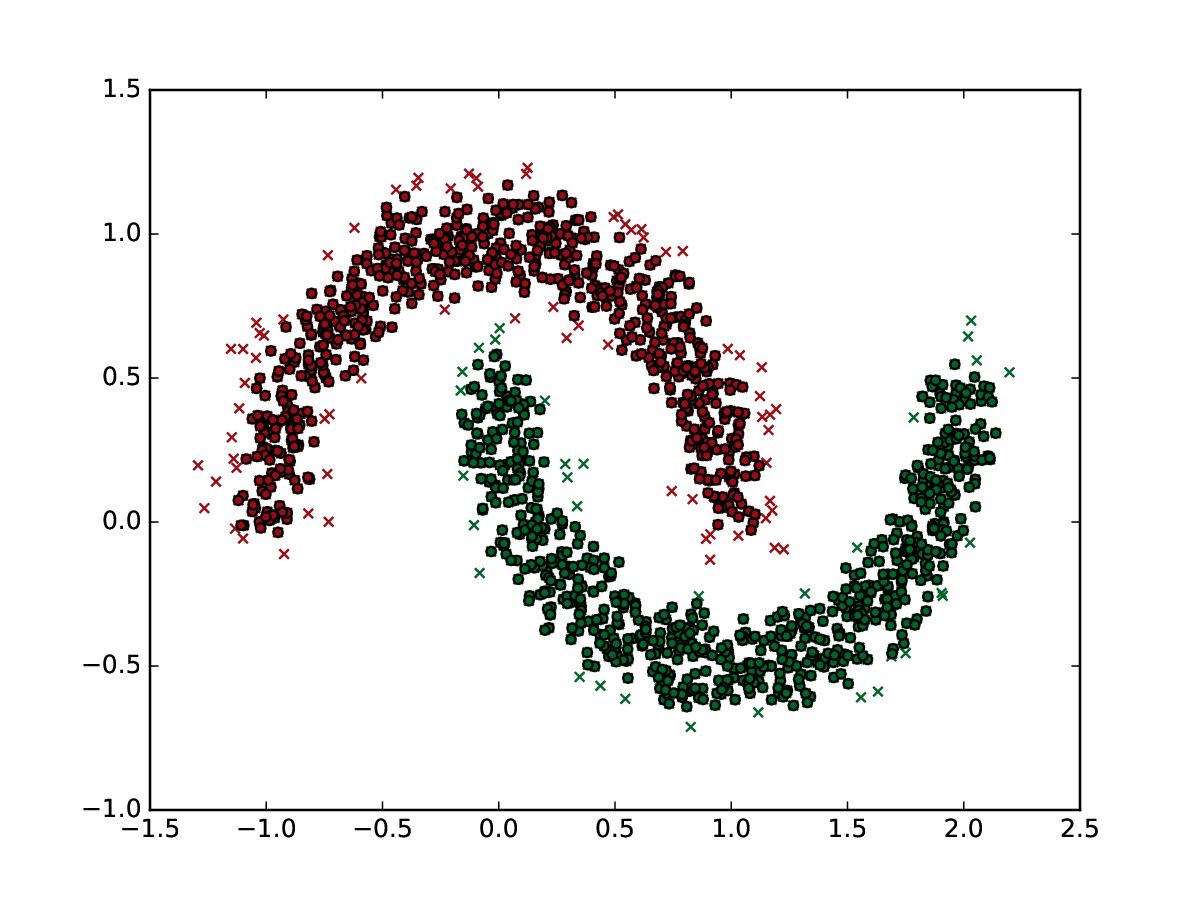}\label{fig:EAPsuccessivetuningstep2clusters}}\hfill
	\caption{The figure shows the results obtained in the second step of successive tuning corresponding to $p=0.6$, $q=-0.97$, $\Delta >1$.}
	\label{fig:EAPsuccessivetuningstep2} 
\end{figure}

For the second step we reduce the linkage penalty while keeping the other hyperparameters same. The new values of the parameters are $p=0.6$, $q=-0.97$ and $\Delta >1$. In this case we discover two clusters and the LEC histogram, shown in Fig.~\ref{fig:EAPsuccessivetuningstep2LEC}, as well as the IES histogram, shown in Fig.~\ref{fig:EAPsuccessivetuningstep2IES}, is no longer shifted either too much towards left or right (taking into account that we still have $\Delta >1$, hence there are a lot more boundary connections formed than needed). The resulting clusters along with the local exemplars are shown in Fig.~\ref{fig:EAPsuccessivetuningstep2clusters}. All three factors, number of cluster, LEC histogram and IES histogram, indicate that we have discovered the global structure well enough and now we need to decrease $\Delta$ if we want to obtain local information corresponding to well separated local exemplars.

As the final step, we reduce $\Delta$ while keeping the other hyperparameters same as the previous step. The new values are $p=0.6$, $q=-0.97$ and $\Delta=0.99$. We again discover two clusters as the number of global clusters is not impacted by a reasonable change in $\Delta$ for appropriately chosen $p$ and $q$. Now we can also see clearly from the LEC and the number of local exemplars discovered that we no longer have very close by local exemplars since the data points otherwise would connect to all the close by exemplars whereas now the data points only connect to few local exemplars. Similarly, we also notice that whole IES histogram is now scaled down, signifying that now the local exemplars are not so close as to share almost all points they are connected to but fewer points that lie on the boundary between the two well separated local exemplars. The sole purpose of $\Delta$ is  to help in discovering cleaner local information, hence one should not try to manipulate global results using $\Delta$ and it should be varied in a very narrow range. 
\begin{figure}[t]
	\centering 
	\subfigure[LEC histogram]{\includegraphics[width=0.24\textwidth]{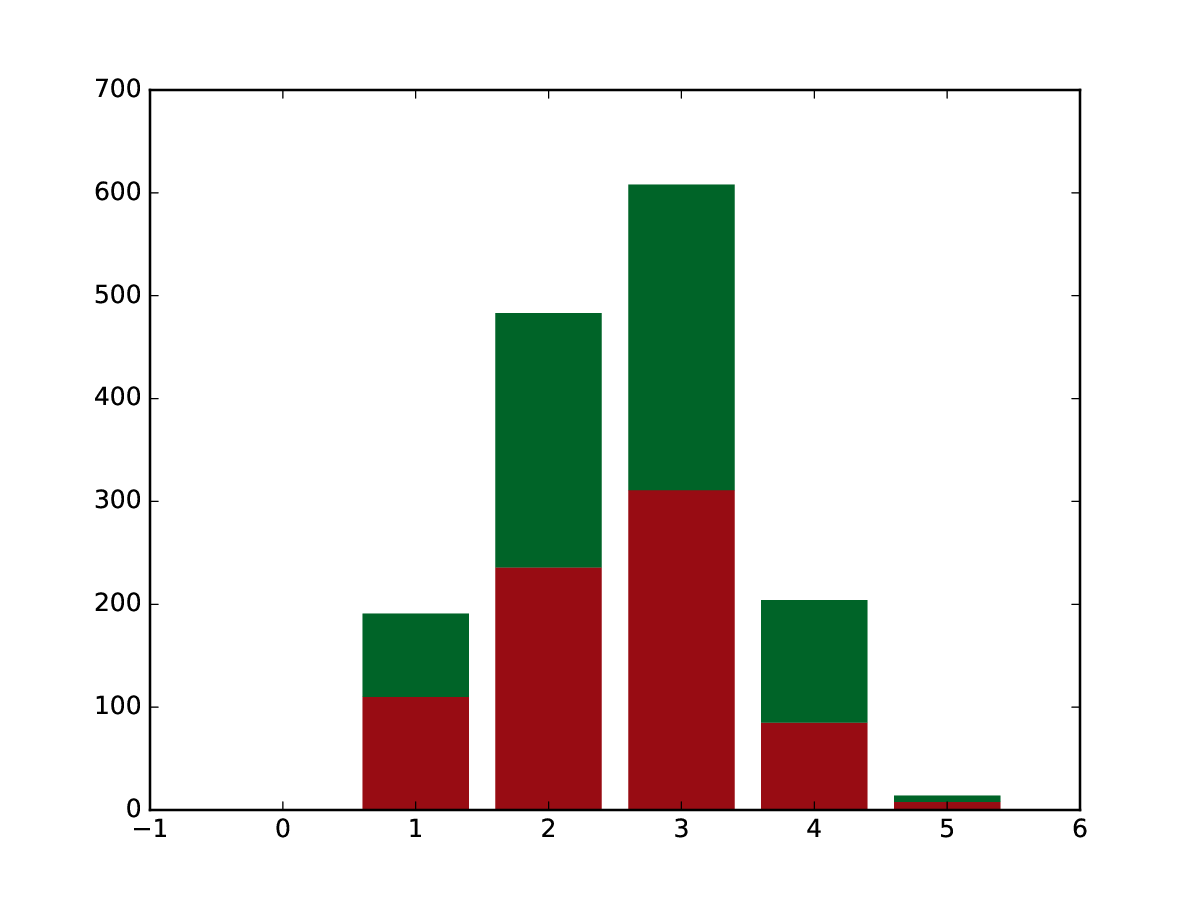}\label{fig:EAPsuccessivetuningstep3LEC}} \hfill
	\subfigure[IES histogram]{\includegraphics[width=0.24\textwidth]{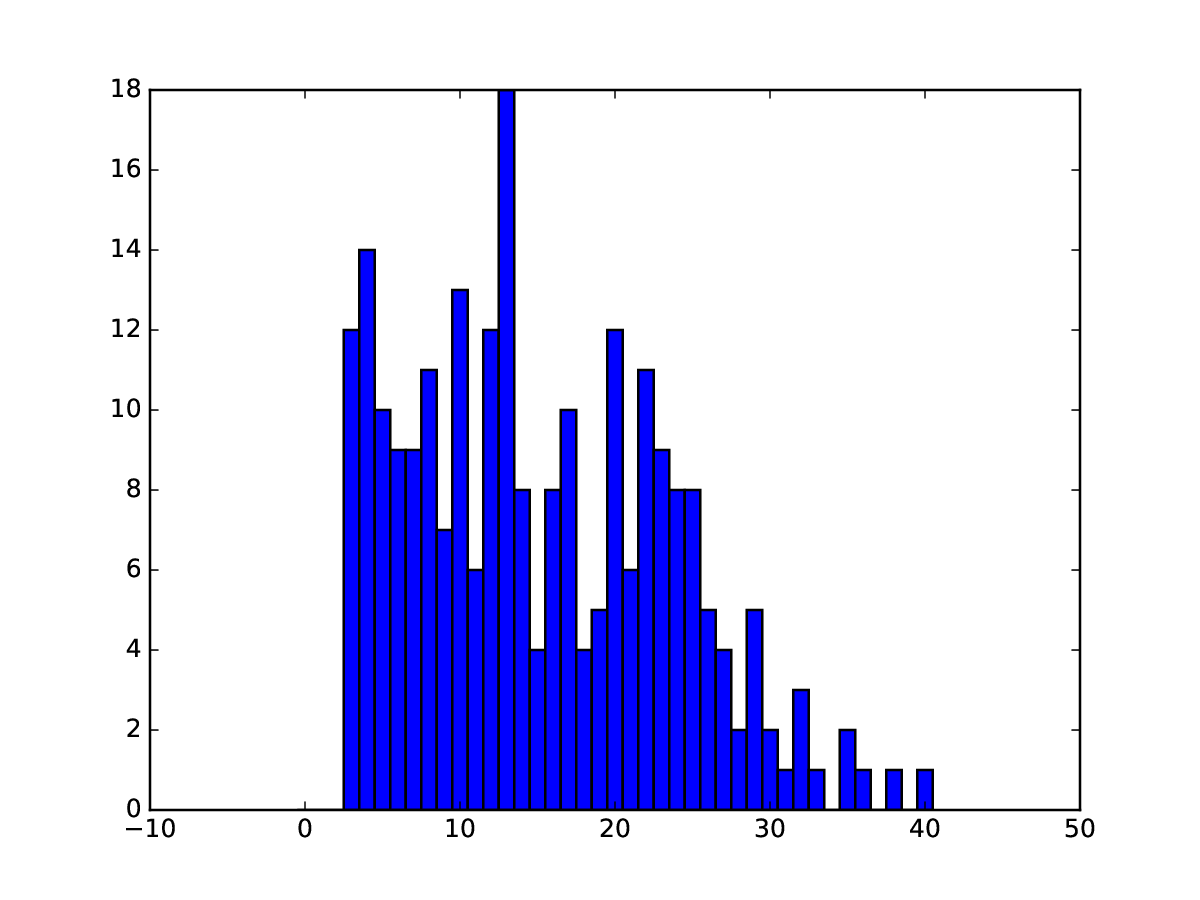}\label{fig:EAPsuccessivetuningstep3IES}}\hfill
	\subfigure[Clusters with Exemplars]{\includegraphics[width=0.4\textwidth]{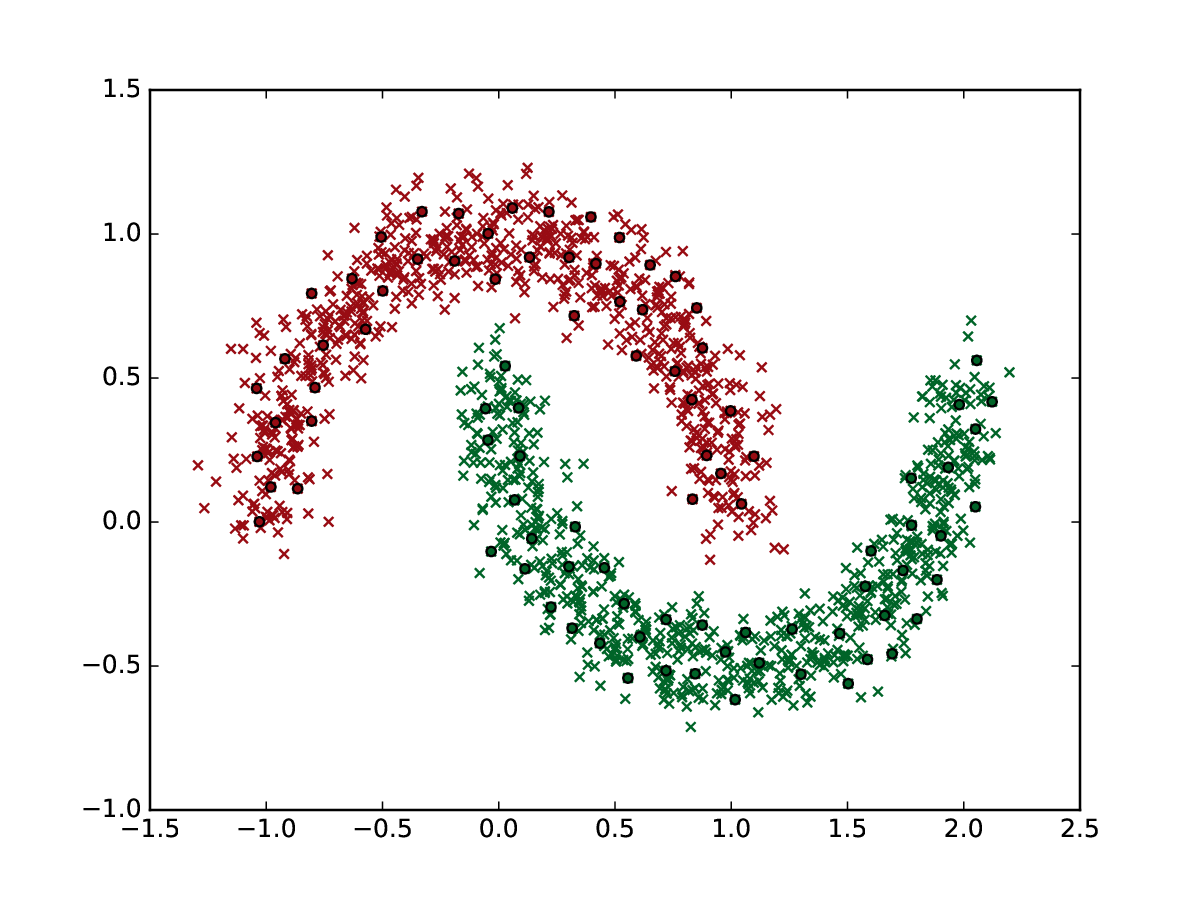}\label{fig:EAPsuccessivetuningstep3clusters}}\hfill
	\caption{The figure shows the results obtained in the final step of successive tuning corresponding to $p=0.6$, $q=-0.97$, $\Delta=0.99$.}
	\label{fig:EAPsuccessivetuningstep3} 
\end{figure}

Note that at every step we only used the available local insights to determine the next step for hyperparameter tuning. Hyperparameter tuning for EAP normally requires only a few steps before discovering the results with both suitable global clusters as well as appropriate local information. In the above example we started successive tuning with such high values of $q$ and $\Delta$ just to illustrate the use of local information for successive tuning. For normal application , one would start with more reasonable values $q$ and $\Delta$, for example $q=-0.97$ and $\Delta=0.99$ in case of Euclidean distance based or probability based pairwise similarities. Furthermore, in the example we did not use the confidence plots in each step but we can also use them to further help us in choosing the next step. 

\section{Sensitivity of Global Discovery to Hyperparameters}\label{sec:tuningsensitive}
We look at how global structure discovery is impacted by the variation of individual hyperparameters.
\begin{itemize}
	\item For a fixed $q$ and $\Delta$, we show how the global structure discovery is relatively robust to variations in $p$ in Fig.~\ref{fig:EAPparamrangep}. In general as long as $p$ is not too low the global cluster results don't get impacted. This is unlike AP where variations in $p$ have strong impact on the number of clusters obtained and the cluster assignments. In the case of EAP, as long as $p$ is varied in a reasonable range, this only changes the number of local exemplars appearing in any region but as long as the number is not too small and they can get connected via boundary connections, the number of globally discovered clusters does not change. Only when $p$ is too low, we are no longer able to form bridge connections between the appropriate exemplars and we lose the ability to discover the global structure. In Fig.~\ref{fig:EAPparamrangep} even for $p=0.1$, which is far outside the range in which we vary $p$ during successive tuning, EAP discovers the correct global structure. 
	      \begin{figure}[t]
	      	\centering 
	      	\subfigure[$p=0.1$]{\includegraphics[width=0.23\textwidth]{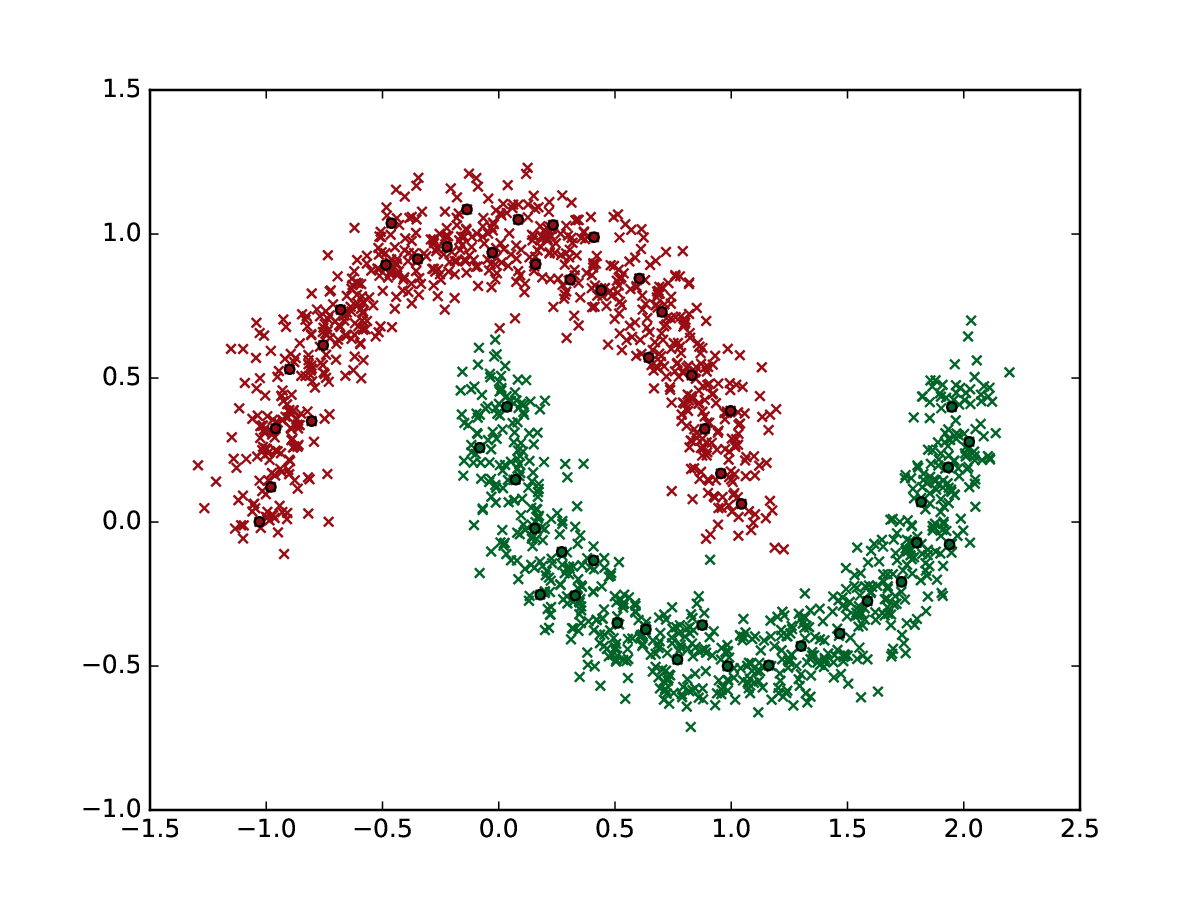}} \hfill
	      	\subfigure[$p=0.3$]{\includegraphics[width=0.23\textwidth]{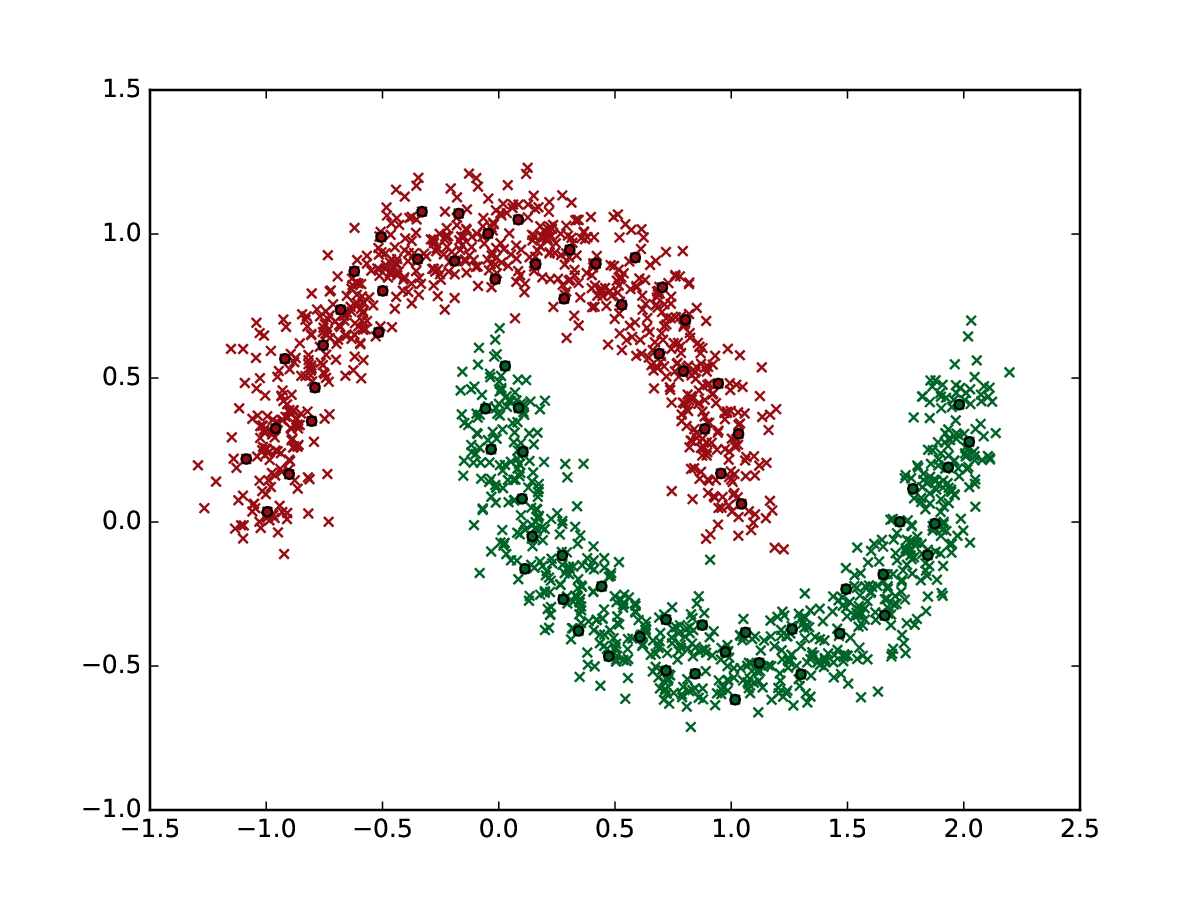}} \hfill
	      	\subfigure[$p=0.6$]{\includegraphics[width=0.23\textwidth]{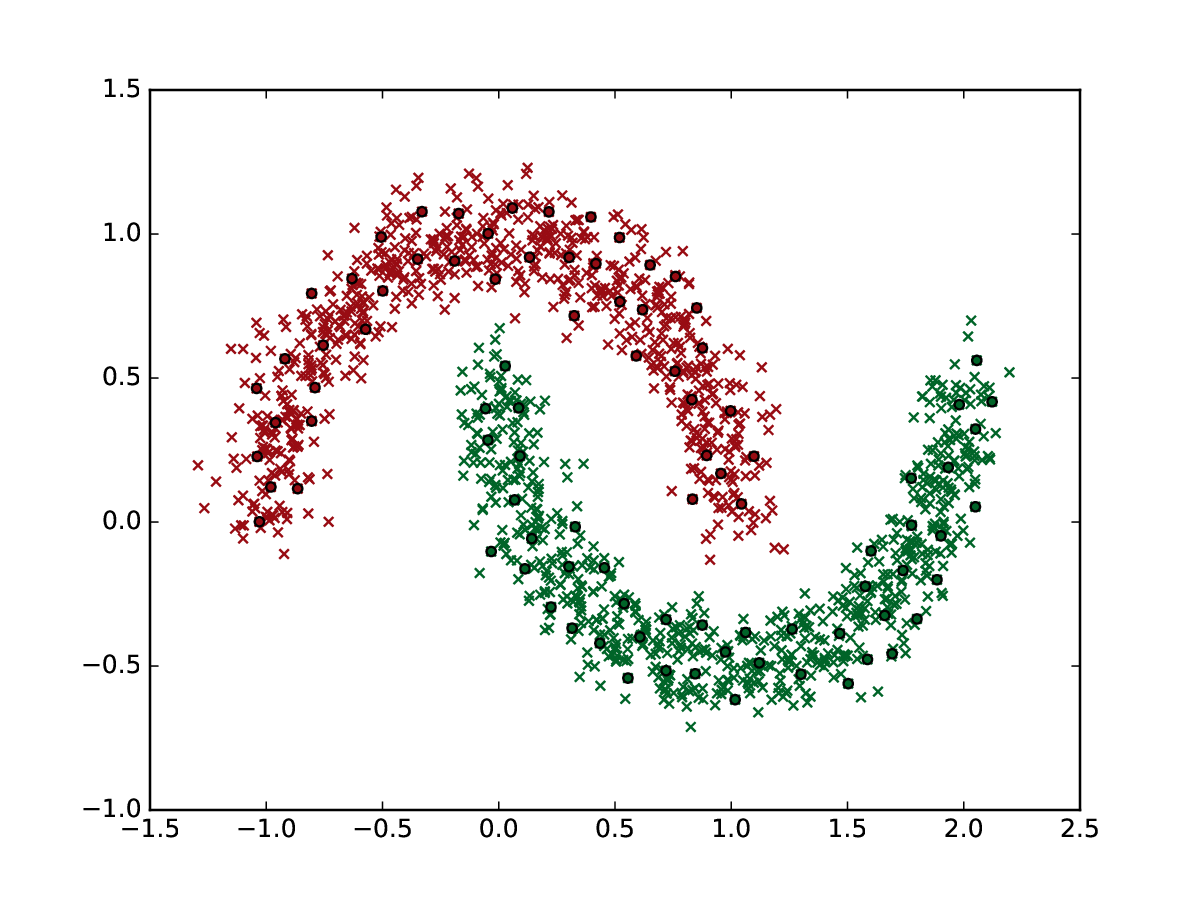}} \hfill
	      	\subfigure[$p=0.8$]{\includegraphics[width=0.23\textwidth]{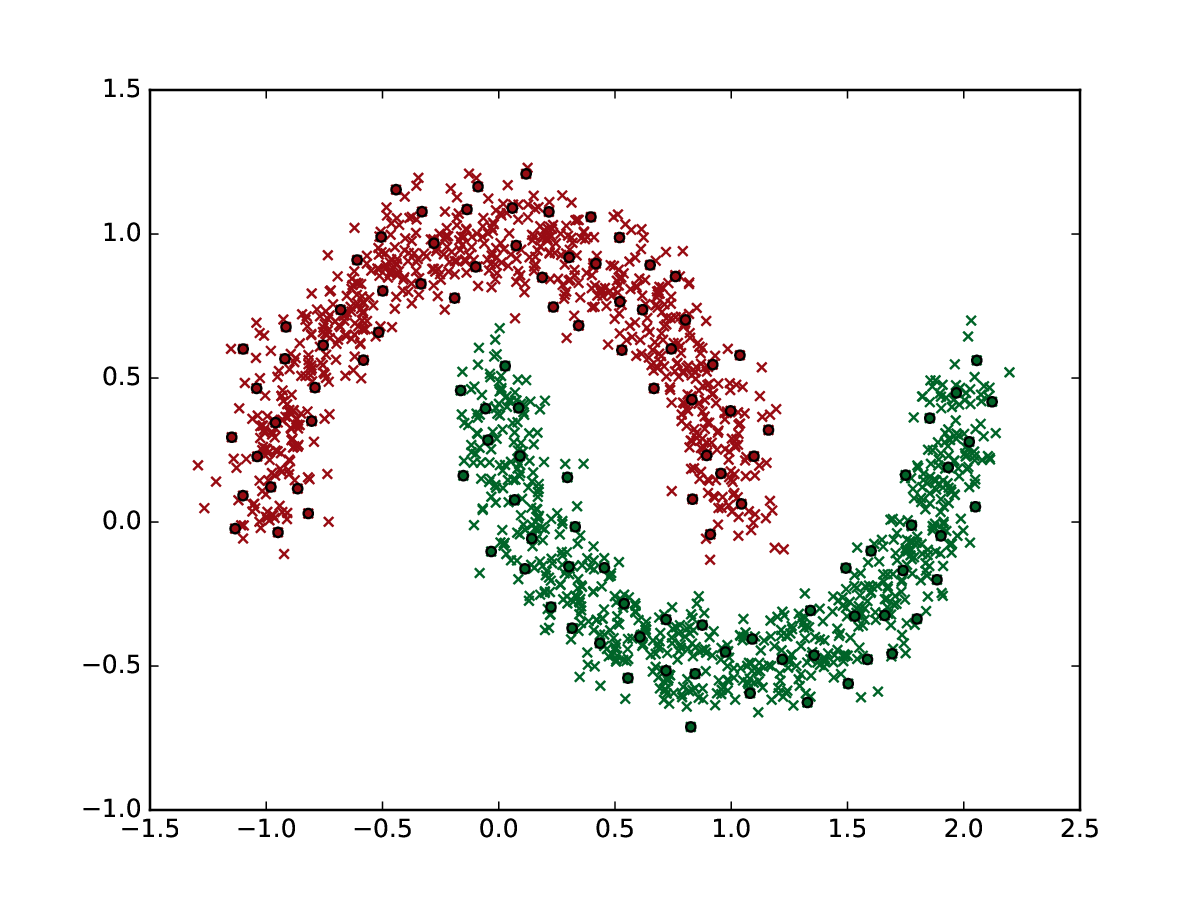}} \hfill
	      	\caption{The figure shows the robustness of the global structure discovery to variations in $p$ where  $q=-0.97$ and $\Delta =0.99$ for all the subfigures.}
	      	\label{fig:EAPparamrangep} 
	      \end{figure}
	\item For a fixed $p$ and $\Delta$, we show how the global structure discovery is impacted by the variation of $q$ in Fig.~\ref{fig:EAPparamrangeq}. $q$ is arguably the hyperparameter which needs most care but there is a reasonable range where the global clusters obtained are not affected and successive tuning is able to discover the a suitable value of $q$ in this range quickly. When we have a too high linkage penalty, clusters start to break at their weakly connected regions as seen in Fig.~\ref{fig:EAPqrangemax}. Note that still in Fig.~\ref{fig:EAPqrangemax} there have not been drastic changes in the results beyond the green cluster being broken where it was most weakly connected.  This can, however, be easily remedied by looking at the confidence plot and recognizing that there is a low confidence region existing between the blue and the green cluster. On the other hand when we have too low linkage penalty very far off local exemplars also get connected by boundary connections as shown in Fig.~\ref{fig:EAPqrangemin}.
	      \begin{figure}[t]
	      	\centering 
	      	\subfigure[$q=-0.955.$]{\includegraphics[width=0.23\textwidth]{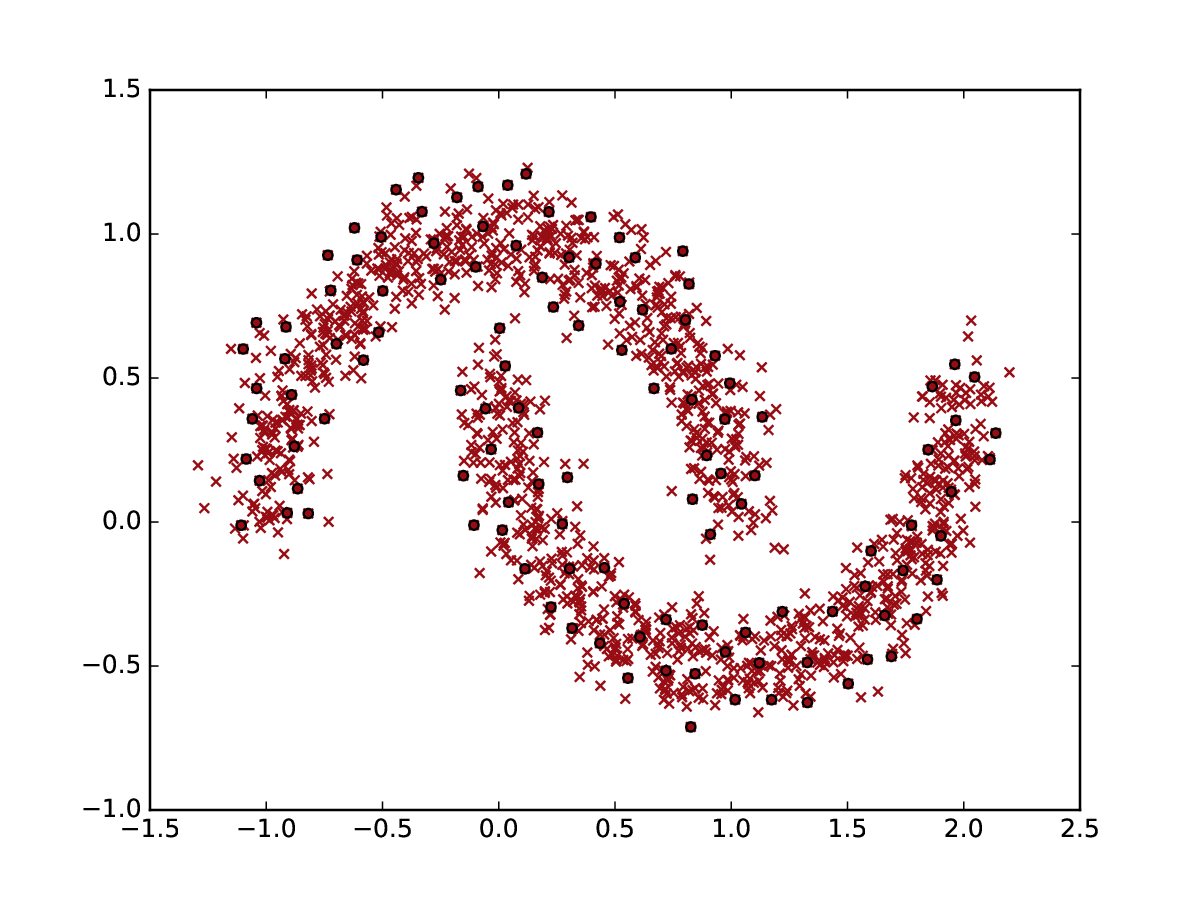}\label{fig:EAPqrangemin}} \hfill
	      	\subfigure[$q=-0.96$]{\includegraphics[width=0.23\textwidth]{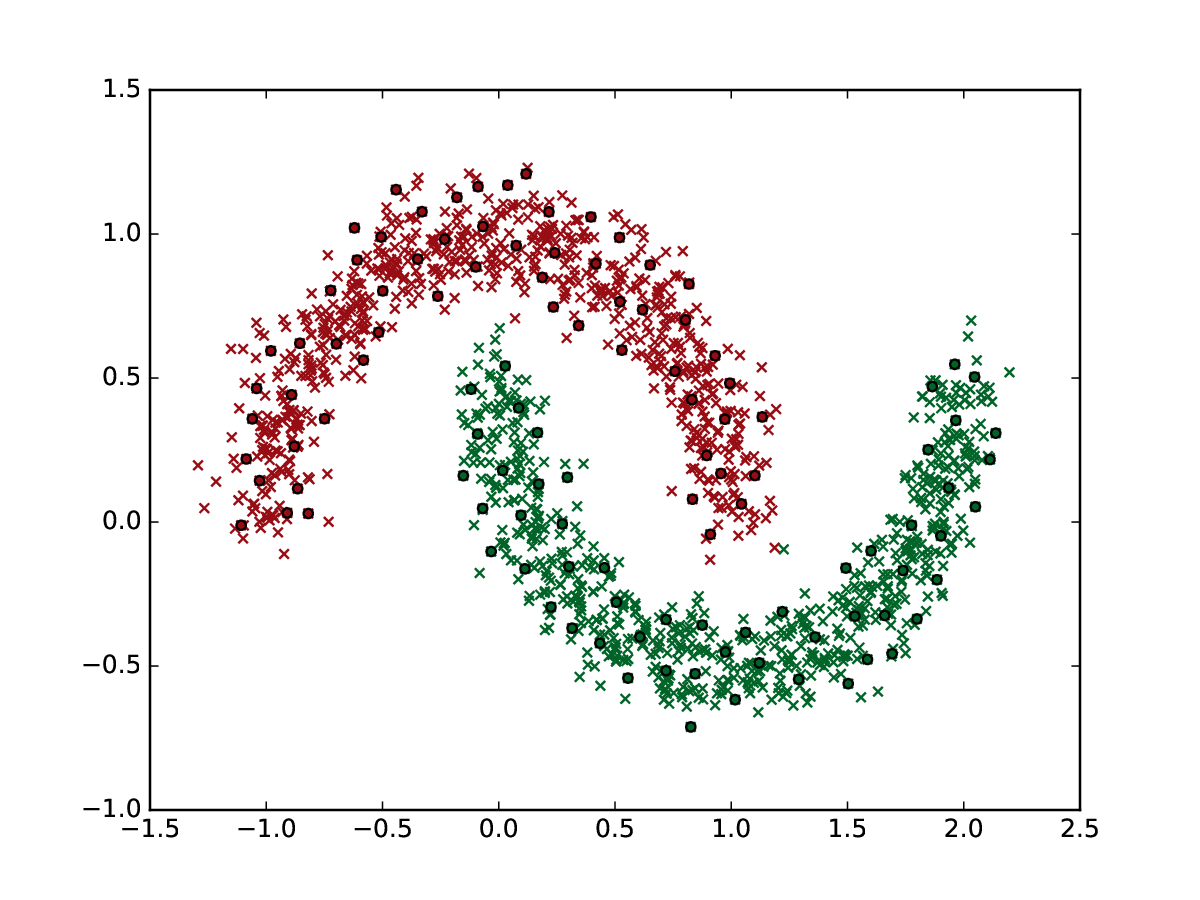}} \hfill
	      	\subfigure[$q=-0.97$]{\includegraphics[width=0.23\textwidth]{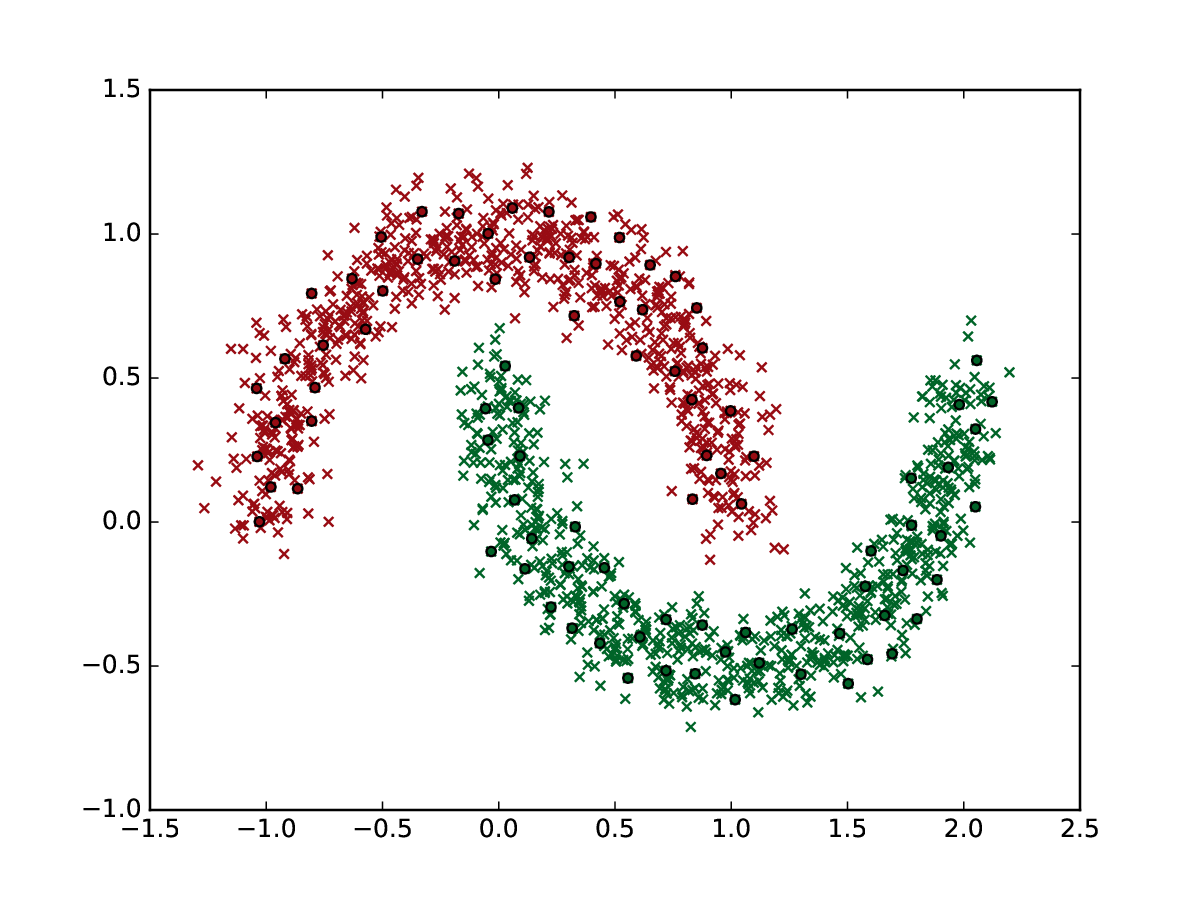}} \hfill
	      	\subfigure[$q=-0.975$]{\includegraphics[width=0.23\textwidth]{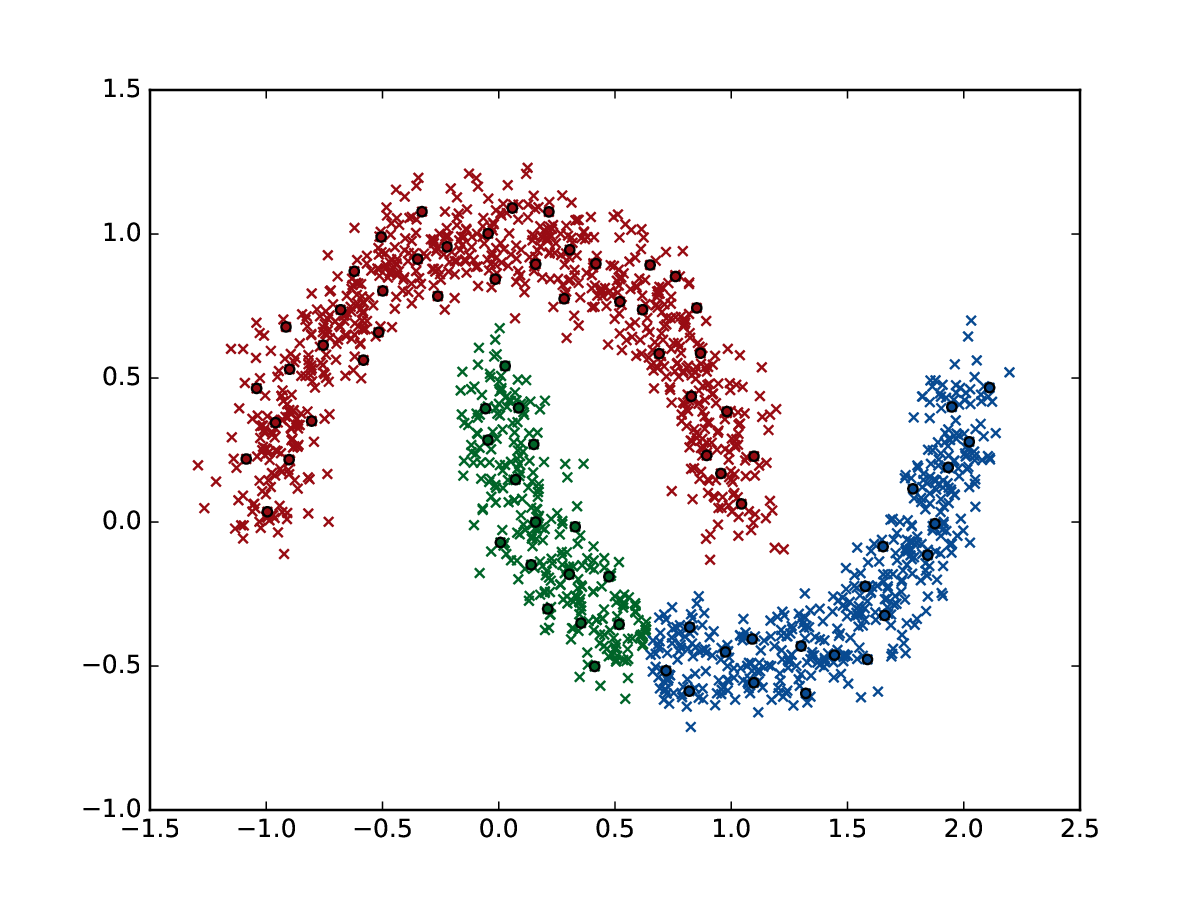}\label{fig:EAPqrangemax}} \hfill
	      	\caption{The figure shows the variation of the discovered clusters w.r.t. variations in $q$ where  $p=0.6$ and $\Delta =0.99$ for all the subfigures.}
	      	\label{fig:EAPparamrangeq} 
	      \end{figure}
	\item For a fixed $p$ and $q$, the global clusters do not change when we start lowering  $\Delta$ from $1$. Too small a value of $\Delta$ can have a similar impact as too small $p$ or too high $q$ where not enough local exemplars appear in the dataset such that they can be connected via boundary connections for global structure discovery. This can be seen in Fig.~\ref{fig:EAPepsrangemin} which leads to similar results as Fig.~\ref{fig:EAPqrangemax} but as we mentioned earlier, $\Delta$ should only be used to adapt the local information, hence one does not need to lower it too much as to change the  global results. It can be seen from experiments on synthetic and real-world datasets that it is quite safe to fix $\Delta$ at $99$ percentile of input similarity matrix.
	      \begin{figure}[b]
	      	\centering 
	      	\subfigure[$\Delta =0.985$]{\includegraphics[width=0.23\textwidth]{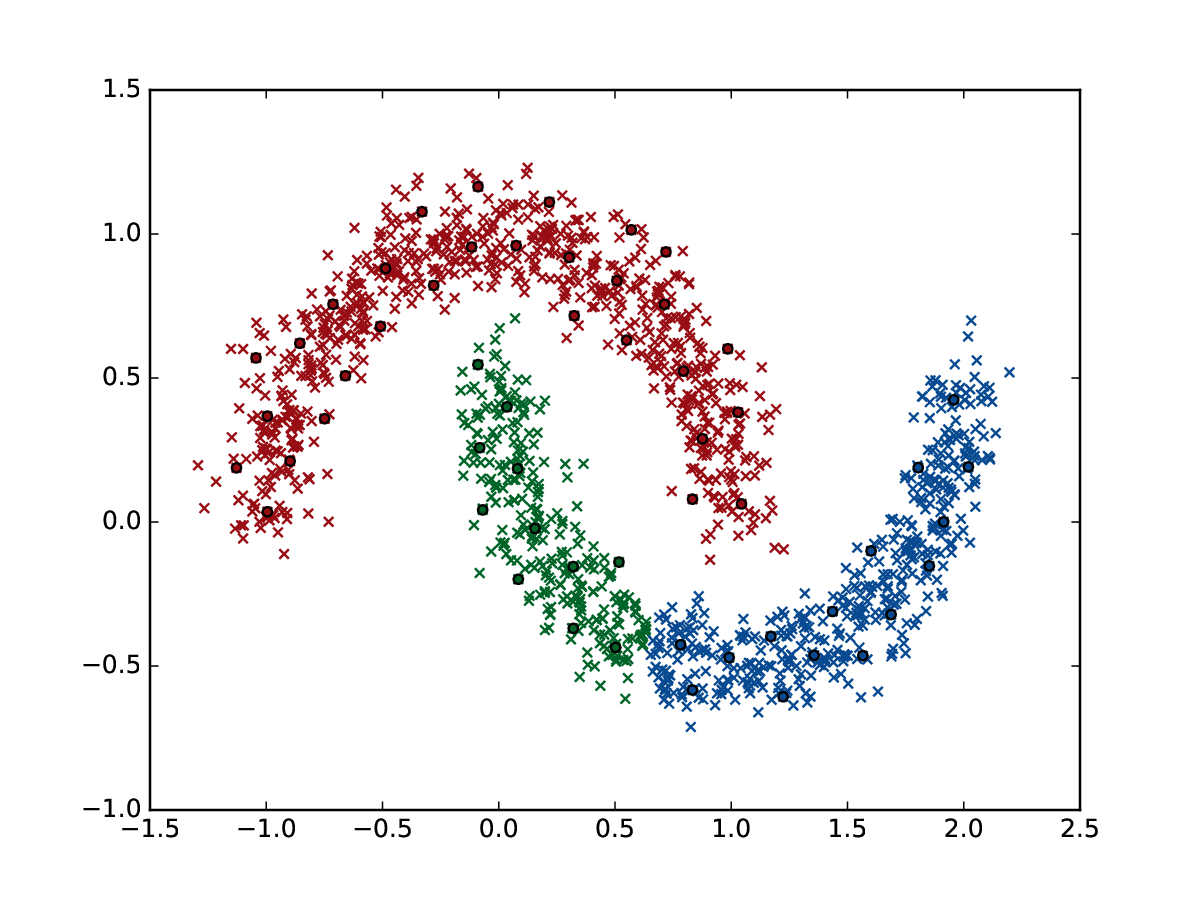}\label{fig:EAPepsrangemin}} \hfill
	      	\subfigure[$\Delta =0.986$]{\includegraphics[width=0.23\textwidth]{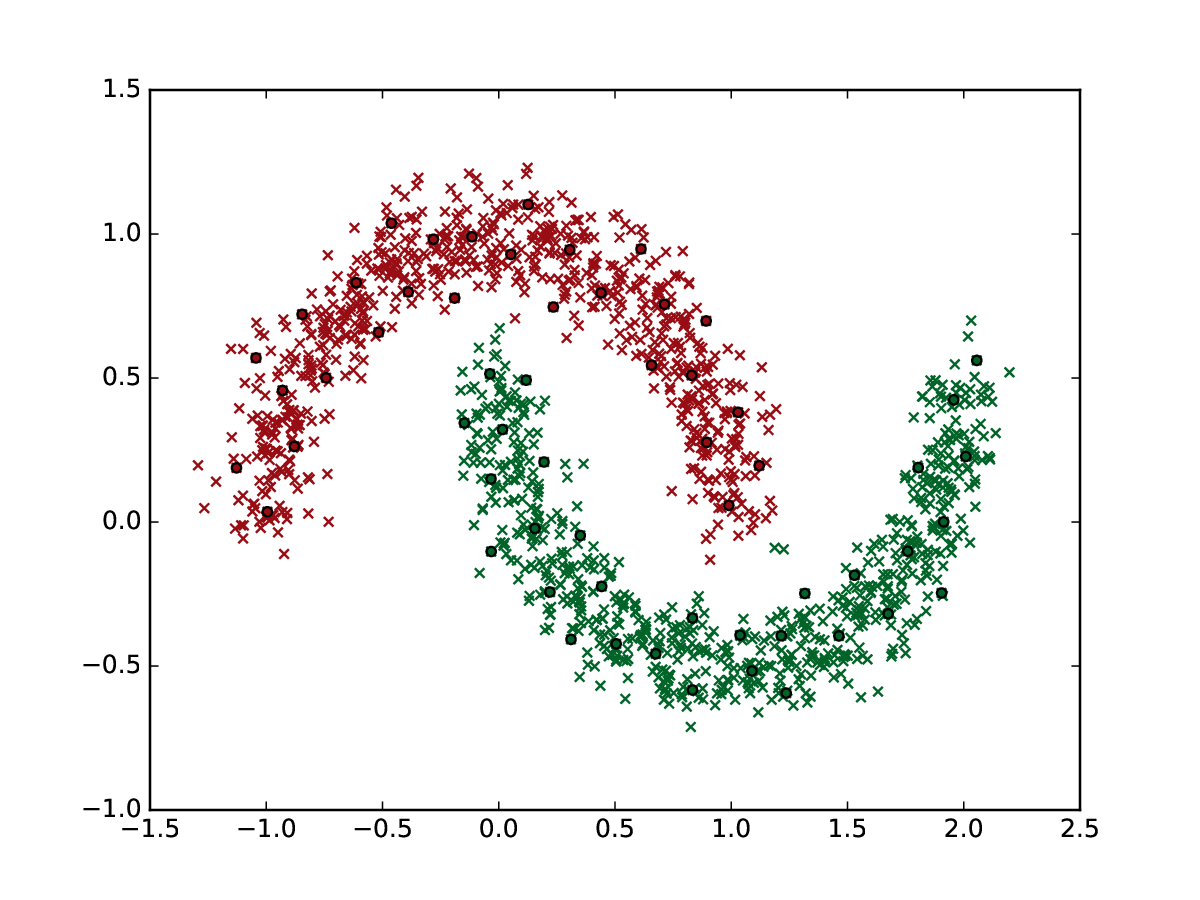}} \hfill
	      	\subfigure[$\Delta =0.99$]{\includegraphics[width=0.23\textwidth]{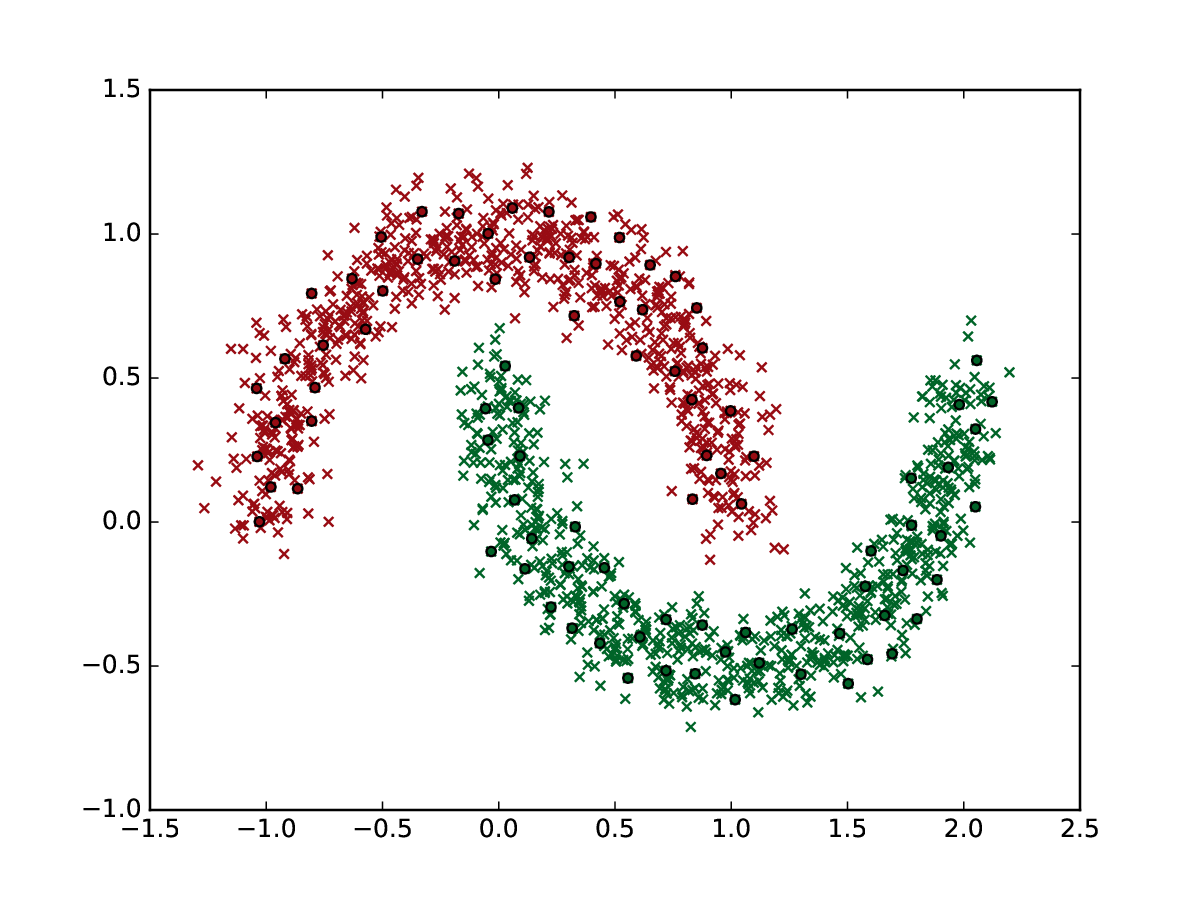}} \hfill
	      	\subfigure[$\Delta >1$]{\includegraphics[width=0.23\textwidth]{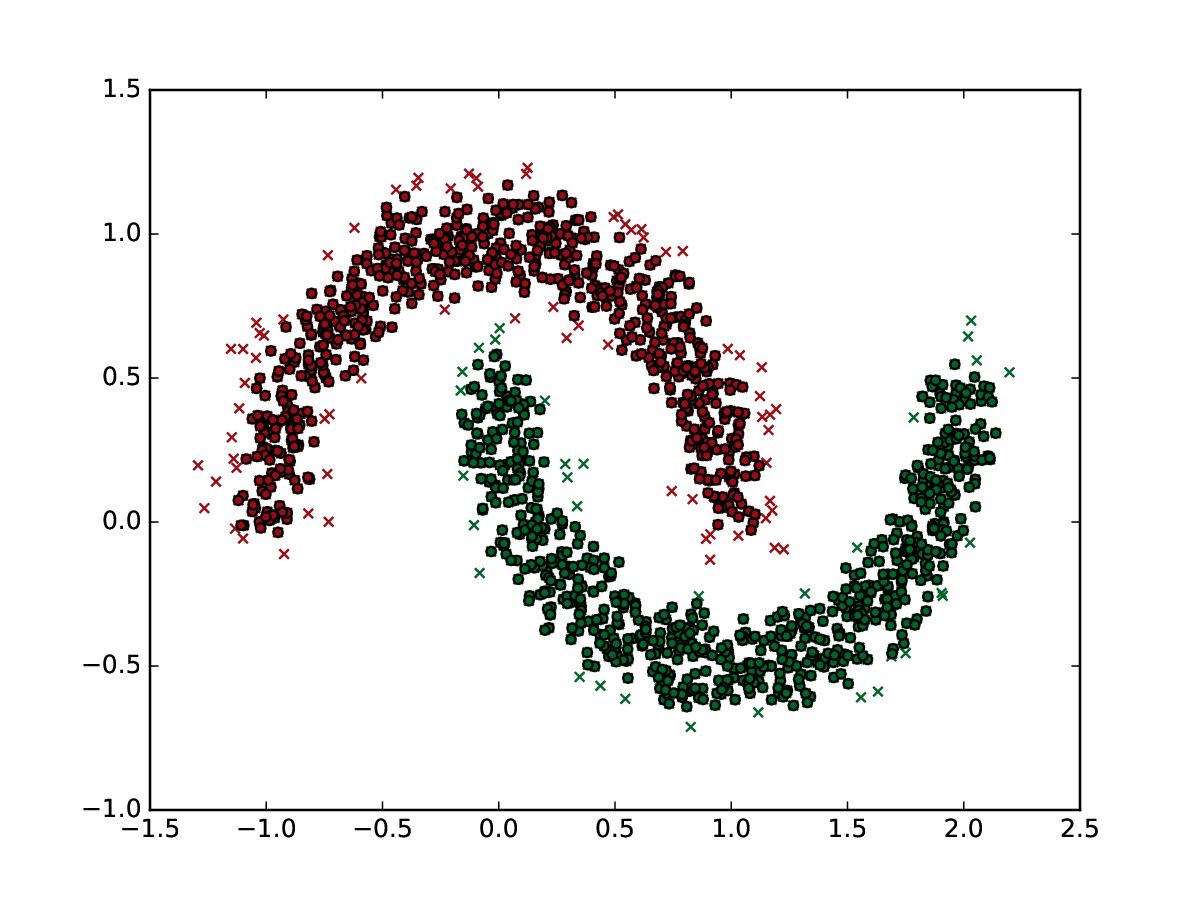}} \hfill
	      	\caption{The figure shows the variation of local and global clustering results due to changes in $\Delta$ where  $p=0.6$ and $q=-0.97$ for all the subfigures.}
	      	\label{fig:EAPparamrangeeps} 
	      \end{figure}
\end{itemize}
\end{document}

%% file: figures/AP_fac_graph.tex
\begin{tikzpicture}
[
>=stealth',shorten >=1pt,auto, node distance=3cm,
every loop/.style={},
var node/.style={circle,draw, thick, minimum size=1.4cm}, 
constraint node/.style={rectangle,draw, minimum width=1cm, minimum height=0.8cm, fill=gray!30}
]
\node[var node] (11) at (0,0) {$b_{11}$};
\node[var node] (i1) at (0,-3) { $b_{i,1}$};
\node[var node] (N1) at (0,-6) { $b_{n,1}$};
\node[constraint node] (S11) at (-1.2, -1.2) {$S_{11}$};
\node[constraint node] (Si1) at (-1.2, -4.2) {$S_{i,1}$};
\node[constraint node] (SN1) at (-1.2, -7.2) {$S_{n,1}$};

\node[var node] (1j) at (3,0) {$b_{1,j}$};
\node[var node] (ij) at (3,-3) {$b_{i,j}$};
\node[var node] (Nj) at (3,-6) {$b_{n,j}$};
\node[constraint node] (S1j) at (1.8, -1.2) {$S_{1,j}$};
\node[constraint node] (Sij) at (1.8, -4.2) {$S_{i,j}$};
\node[constraint node] (SNj) at (1.8, -7.2) {$S_{n, j}$};

\node[var node] (1N) at (6,0) { $b_{1,n}$};
\node[var node] (iN) at (6,-3) { $b_{i,n}$};
\node[var node] (NN) at (6,-6) { $b_{n,n}$};
\node[constraint node] (S1N) at (4.8, -1.2) { $S_{1,n}$};
\node[constraint node] (SiN) at (4.8, -4.2) {$S_{i,n}$};
\node[constraint node] (SNN) at (4.8, -7.2) {$S_{n, n}$};

\path (11) -- node[auto=false]{ \vdots} (i1);
\path (11) -- node[auto=false]{ \ldots} (1j);
\path (i1) -- node[auto=false]{ \vdots} (N1);
\path (i1) -- node[auto=false]{ \ldots} (ij);
\path (1j) -- node[auto=false]{ \vdots} (ij);
\path (1j) -- node[auto=false]{ \ldots} (1N);
\path (ij) -- node[auto=false]{ \vdots} (Nj);
\path (ij) -- node[auto=false]{ \ldots} (iN);
\path (1N) -- node[auto=false]{ \vdots} (iN);
\path (iN) -- node[auto=false]{ \vdots} (NN);
\path (N1) -- node[auto=false]{ \ldots} (Nj);
\path (Nj) -- node[auto=false]{ \ldots} (NN);

\node[constraint node] (E1) at (1.5, 2) {$h_1$};% $E_1$
\node[constraint node] (Ej) at (4.5, 2) {$h_j$};% $E_j$
\node[constraint node] (EN) at (7.5, 2) {$h_{n}$};% $E_N$

\node[constraint node] (I1) at (-4, 2) {$g_1$};% $I_1$
\node[constraint node] (Ii) at (-4, -1) { $g_i$};% $I_i$
\node[constraint node] (IN) at (-4, -4) {$g_{n}$};% $I_N$

\path[]
(11)	edge node  {} (I1)
edge node  {} (E1)
edge node  {} (S11)
(i1)	edge node  {} (Ii)
edge node  {} (E1)
edge node  {} (Si1)
(N1)	edge node  {} (IN)
edge node  {} (E1)
edge node  {} (SN1)

(1j)	edge node  {} (I1)
edge node  {} (Ej)
edge node  {} (S1j)
(ij)	edge node  {} (Ii)
edge node  {} (Ej)
edge node  {} (Sij)
(Nj)	edge node  {} (IN)
edge node  {} (Ej)
edge node  {} (SNj)

(1N)	edge node  {} (I1)
edge node  {} (EN)
edge node  {} (S1N)
(iN)	edge node  {} (Ii)
edge node  {} (EN)
edge node  {} (SiN)
(NN)	edge node  {} (IN)
edge node  {} (EN)
edge node  {} (SNN);

\end{tikzpicture}

%% file: figures/AP_messages.tex
\begin{tikzpicture}
[
>=stealth',shorten >=1pt,auto, node distance=2cm,
every loop/.style={},
var node/.style={circle,draw}, 
constraint node/.style={rectangle,draw, minimum width=0.8cm, minimum height=0.6cm, fill=gray!30}
]
\node[var node] (1) {$b_{ij}$};

\node[constraint node] (2) [above of=1]{$h_j$};%$E_j$
\node[constraint node] (3) [left of=1] {$g_i$};%$I_i$
\node[constraint node] (4) [below of=1] {$S_{ij}$};%$s_{ij}$

\draw[thick,->] (1) ++(0.2, 0.5) -- ++(0,1) node[midway ,right]{$\rho_{ij}$};
\draw[thick,->] (2) ++(-0.2, -0.5) -- ++(0,-1) node[midway ,left]{$\alpha_{ij}$};

\draw[thick,->] (1) ++(-0.5, 0.2) -- ++(-1,0) node[midway ,above]{$\beta_{ij}$};
\draw[thick,->] (3) ++(0.5, -0.2) -- ++(1,0) node[midway ,below]{$\eta_{ij}$};

\draw[thick,->] (4) ++(0.2, 0.5) -- ++(0,1) node[midway ,right]{$s_{ij}$};
\path[]
(1) edge node [right] {} (2)
edge node [left] {} (3)
edge node [below left] {} (4);

\end{tikzpicture}

%% file: figures/EAP_diagonalnode.tex
	\begin{tikzpicture}
	[
	>=stealth',shorten >=1pt,auto, node distance=2cm,
	every loop/.style={},
	var node/.style={circle,draw}, 
	constraint node/.style={rectangle,draw, minimum width=0.8cm, minimum height=0.6cm, fill=gray!30}
	]
	\node[var node] (hi) {$b_{ii}$};

	\node[constraint node] (Ei) [above of=hi]{$h_i$};%$E_j$
	\node[constraint node] (Ii) [left of=hi] {$\overline{g}_i$};%$I_i$
	\node[constraint node] (Si) [below of=hi] {$S_{ii}$};%$s_{ij}$
	
	\node[constraint node] (Fj) [right of=hi]{$r_j$};
	\node[constraint node] (Fi) [right of=Ei]{$r_i$};
	\node[constraint node] (Fk) [right of=Si]{$r_k$};

	\draw[thick,->] (hi) ++(0.2, 0.5) -- ++(0,1) node[midway ,right]{$\rho_{ii}$};
	\draw[thick,->] (Ei) ++(-0.2, -0.5) -- ++(0,-1) node[midway ,left]{$\alpha_{ii}$};
	
	\draw[thick,->] (hi) ++(-0.5, 0.2) -- ++(-1,0) node[midway ,above]{$\beta_{ii}$};
	\draw[thick,->] (Ii) ++(0.5, -0.2) -- ++(1,0) node[midway ,below]{$\eta_{ii}$};
	
	\draw[thick,->] (Fj) ++(-0.5, 0.2) -- ++(-1,0) node[midway ,above]{$\psi_{ij}$};
	\draw[thick,->] (hi) ++(0.5, -0.2) -- ++(1,0) node[midway ,below]{$\phi_{ij}$};
	
	\draw[thick,->] (Si) ++(0.2, 0.5) -- ++(0,1) node[midway ,right]{$s_{ii}$};
	
	\draw [decorate,decoration={brace,amplitude=16pt},xshift=12pt,yshift=0pt]
	(Fi)++(1, 0) -- ++(0, -4) node [black,midway,xshift=0.6cm] {\footnotesize $\mc{M}_i$};
	\path[]
	(hi)    edge node [right] {} (Ei)
	edge node [left] {} (Ii)
	edge node [below left] {} (Si)
	edge node [right] {} (Fj)
	edge node [right] {} (Fi)
	edge node [right] {} (Fk);
	\path[dotted, ultra thick]  		
	(Fi)++(0, -0.75) edge ++(0, -0.5)
	(Fj)++(0, -0.75) edge ++(0, -0.5);

	% 	(Fj)    edge ++(1, 0.5)
	% 	        edge ++(1, 0)
	% 	        edge ++(1, -0.5)
	% 	(Fk)    edge ++(1, 0.5)
	% 	        edge ++(1, 0)
	% 	        edge ++(1, -0.5);

	\end{tikzpicture}